\documentclass{article}

\usepackage{arxiv}
\usepackage[square,compress,numbers]{natbib}

\usepackage[utf8]{inputenc} % allow utf-8 input
\usepackage[T1]{fontenc}    % use 8-bit T1 fonts
\usepackage{hyperref}       % hyperlinks
\hypersetup{
    colorlinks=false,
    linkcolor=black,
    filecolor=black,      
    urlcolor=black,
}
\usepackage{url}            % simple URL typesetting
\usepackage{booktabs}       % professional-quality tables
\usepackage{amsfonts}       % blackboard math symbols
\usepackage{nicefrac}       % compact symbols for 1/2, etc.
\usepackage{microtype}      % microtypography
\usepackage{lipsum}
\usepackage{graphicx}
\usepackage{amsmath}
\usepackage{multirow}
\usepackage{xcolor} %--->remove me, for corrections!
\usepackage{xspace}
\newcommand{\hlsfml}{\texttt{hls4ml}\xspace}
\newcommand{\QKeras}{\textsc{QKeras}\xspace}
\newcommand{\Keras}{\textsc{Keras}\xspace}
\newcommand{\TensorFlow}{\textsc{TensorFlow}\xspace}
\newcommand{\PyTorch}{\textsc{PyTorch}\xspace}
\newcommand{\ONNX}{\textsc{ONNX}\xspace}
\newcommand{\AutoQKeras}{\textsc{AutoQKeras}\xspace}
\newcommand{\QTools}{\textsc{QTools}\xspace}
\newcommand{\unit}[1]{\ensuremath{\text{\,#1}}\xspace}
\usepackage{doi}
\newlength\cmsTabSkip

\DeclareMathOperator*{\argmax}{arg\,max}
\title{%Ultra-compressed, low-latency 
Fast convolutional neural networks on FPGAs with hls4ml}

\author{
  Thea Aarrestad, Vladimir Loncar\thanks{Also at Institute of Physics Belgrade, Serbia}, Nicol\`{o} Ghielmetti\thanks{Also at Politecnico di Milano, Italy}, Maurizio Pierini, Sioni Summers \\
  European Organization for Nuclear Research (CERN) \\
  CH-1211 Geneva 23, Switzerland
    \And
    Jennifer Ngadiuba\\
    California Institute of Technology\\
    Pasadena, CA 91125, USA
    
    \And
  Christoffer Petersson\thanks{Also at Chalmers University of Technology, Sweden}, Hampus Linander\\
  Zenseact \\
  Gothenburg, 41756, Sweden
  \And
  Yutaro Iiyama \\
  ICEPP, University of Tokyo \\
  Tokyo, Japan
  \And
  Giuseppe Di Guglielmo \\
  Columbia University, New York, \\
  NY 10027, USA
  \And
  Javier Duarte \\
  University of California San Diego\\
  La Jolla, CA 92093, USA \\
  \And 
  Philip Harris, Dylan Rankin\\
  Massachusetts Institute of Technology\\
  Cambridge, MA 02139, USA 
  \And
  Sergo Jindariani, Kevin Pedro, Nhan Tran \\
  Fermi National Accelerator Laboratory\\
  Batavia, IL 60510, USA\\
  \And
  Mia Liu \\
  Purdue University \\
  West Lafayette, IN 47907, USA
  \And
  Edward Kreinar\\
  HawkEye360\\
  Herndon, VA 20170, USA\\
  \And
  Zhenbin Wu\\
  University of Illinois at Chicago\\
  Chicago, IL 60607, USA
  \And
  Duc Hoang\\
  Rhodes College\\
  Memphis, TN 38112, USA

}

\begin{document}
%%%%%%%%%%%%%%%%%%%%%%%%%%%%%%%%%%%
% hls4ml logo
\begin{center}
\includegraphics[width=8cm]{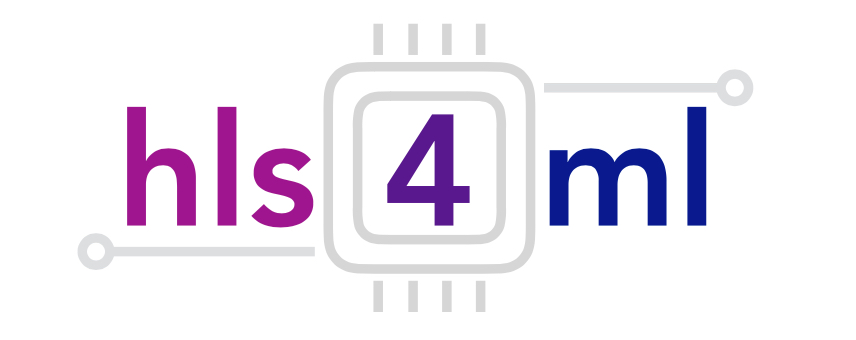}
\end{center}
%%%%%%%%%%%%%%%%%%%%%%%%%%%%%%%%%%%
\maketitle

\begin{abstract}
  We introduce an automated tool for deploying ultra low-latency, low-power deep neural networks with convolutional layers on FPGAs. 
  By extending the \texttt{hls4ml} library, we demonstrate an inference latency of 5\,$\mu$s using convolutional architectures, targeting microsecond latency applications like those at the CERN Large Hadron Collider. 
  Considering benchmark models trained on the Street View House Numbers Dataset, we demonstrate various methods for model compression in order to fit the computational constraints of a typical FPGA device used in trigger and data acquisition systems of particle detectors. 
  In particular, we discuss pruning and quantization-aware training, and demonstrate how resource utilization can be significantly reduced with little to no loss in model accuracy. We show that the FPGA critical resource consumption can be reduced by 97\% with zero loss in model accuracy, and by 99\% when tolerating a 6\% accuracy degradation.
\end{abstract}

% keywords can be removed

\section{Introduction}

The \hlsfml library~\cite{Duarte:2018ite,hls4ml_github} is an open source software designed to facilitate the deployment of machine learning (ML) models on field-programmable gate arrays (FPGAs), targeting low-latency and low-power edge applications.  
Taking as input a neural network model, \hlsfml generates \texttt{C/C++} code designed to be transpiled into FPGA firmware by processing it with a high-level synthesis (HLS) library. 
The development of \hlsfml was historically driven by the need to integrate ML algorithms in the first stage of the real-time data processing of particle physics experiments operating at the CERN Large Hadron Collider (LHC). 
The LHC produces high-energy proton collisions (or \textit{events}) every 25\unit{ns}, each consisting of about 1\unit{MB} of raw data. 
Since this throughput is overwhelming for the currently available processing and storage resources, the LHC experiments run a real-time event selection system, the so-called Level-1 trigger (L1T), to reduce the event rate from 40\unit{MHz} to 100\unit{kHz}~\cite{ATLASL1T,ATLASP2L1T,CMSL1T,CMSP2L1T}. 
Due to the size of the buffering system, the L1T system operates with a fixed latency of $\mathcal{O}$(1\unit{$\mu$s}). 
While \hlsfml excels as a tool to automatically generate low-latency ML firmware for L1T applications, it also offers interesting opportunities for edge-computing applications beyond particle physics whenever efficient, e.g. low power or low latency, on-sensor edge processing is required.

The \hlsfml software is structured with a set of different back-ends, each supporting a different HLS library and targeting different FPGA vendors. 
So far, new development has been focused on the Vivado HLS~\cite{vivadohls} back-end targeting Xilinx FPGAs.
We have demonstrated this workflow for fully-connected, or dense, neural networks (DNNs)~\cite{Duarte:2018ite}, binary and ternary networks~\cite{DiGuglielmo:2020eqx}, boosted decision trees~\cite{Summers:2020xiy}, and graph neural networks~\cite{Iiyama:2020wap,Heintz:2020soy}. 
The \hlsfml library accepts models from \TensorFlow~\cite{tensorflow2015-whitepaper}, \Keras~\cite{chollet2015keras}, \PyTorch~\cite{pytorch}, and via the \ONNX interface~\cite{onnx_2017}.
It has recently been interfaced to \QKeras~\cite{Coelho:2020zfu}, in order to support quantization-aware training (QAT) allowing the user to better balance resource utilization and accuracy. 

The \hlsfml design focuses on fully-on-chip deployment of neural network architectures. 
This avoids the latency overhead incurred by data transmission between the embedded processing elements and off-chip memory, reducing the overall inference latency. 
Conversely, this approach constrains the size and complexity of the models that the HLS conversion can easily support. 
Nevertheless, complex architectures can be supported, as discussed in~\cite{Iiyama:2020wap,Heintz:2020soy} in the case of graph neural networks. 

In this paper, we introduce support for convolutional neural networks (CNNs), through the implementation of streaming-based novel convolutional and pooling layers. 

Given the larger number of operations associated to each convolutional layer, a successful deployment on FPGA relies on model compression, through pruning and quantization. 
The \hlsfml library supports both these forms of compression through removal of all zero-multiplications during the firmware implementation (a feature of HLS we take advantage of when designing the layer implementation), and through its interface with \QKeras~\cite{Coelho:2020zfu}. 

We demonstrate the \textsc{QKeras}+\hlsfml workflow on a digit classifier trained on the Street View House Numbers (SVHN) dataset~\cite{Netzer2011}, with a depth and input size appropriate for the latency- and resource-restricted triggering systems at LHC.

This paper is organized as follows: Section~\ref{sec:related} describes related works. 
Section~\ref{sec:large_CNN} introduces the stream-based implementation of CNN layers; Section~\ref{sec:data} describes the SVHN dataset. 
The benchmark model is introduced in Section~\ref{sec:model}, while results obtained by pruning and quantization (after and during training) are presented in Sections~\ref{sec:pruning}~and~\ref{sec:quantization}, respectively. 
Section~\ref{sec:fpgaporting} discusses the model porting to FPGAs. 
Conclusions are given in Section~\ref{sec:conclusion}.

\section{Related work}
\label{sec:related}
An early attempt to deploy convolutional neural networks (CNNs) on FPGAs for particle physics was shown in~\cite{Calafiura-NIPS}, and surveys of other existing toolflows for mapping CNNs on FPGAs are given in~\cite{2018arXiv180305900V,10.1145/3289185,Shawahna_2019,abdelouahab2018accelerating}. 
The FINN~\cite{FINN,FINNR} framework from Xilinx Research Labs is designed to explore quantized CNN inference on FPGAs, with emphasis on generating dataflow-style architectures customized for each network. 
It includes tools for training quantized NNs such as \textsc{Brevitas}~\cite{brevitas}, the FINN compiler, and the finn-hlslib Vivado HLS library of FPGA components for QNNs. 
The fpgaConvNet library~\cite{venieris2017fpgaconvnet,venieris2017fpga,venieris2017fpl,venieris2016fccm} converts CNNs specified in Caffe~\cite{jia2014caffe} or Torch formats into generated Xilinx Vivado HLS code with a streaming architecture. 
FP-DNN~\cite{fpdnn} is a framework that takes \TensorFlow~\cite{tensorflow2015-whitepaper}-described CNNs as input, and generates the hardware implementations on FPGA boards with RTL-HLS hybrid templates. 
DNNWeaver~\cite{dnnweaver:micro16} is an open-source alternative, which also supports CNNs specified in Caffe format and automatically generates the accelerator Verilog code using hand-optimized Verilog templates with a high degree of portability.
Caffeine~\cite{caffeinatedFPGAs} is another CNN accelerator for Caffe-specified models targeting Xilinx devices that support a co-processing environment with a PCIe interface between the FPGA and a host. 
Snowflake~\cite{snowflake} is a scalable and efficient CNN accelerator with models specified in Torch~\cite{torch} and a single, sequential computation architecture designed to perform at near-peak hardware utilization targeting Xilinx system-on-chips (SoCs). In ~\cite{majumder2019flexible}, an FPGA-based accelerator design to execute CNNs is proposed, leveraging \TensorFlow for model description and exploiting reuse along all dimensions with a 1D systolic array of processing elements. The NullHop~\cite{nullhop} accelerator architecture takes advantage of sparse computation in convolutional layers to significantly speed up inference times.
%is highly flexible and avoids reconfiguration while allowing high utilization for arbitrary aspect ratio tiles of the larger layer dimensions.
A flexible, efficient 3D neuron array architecture for CNNs on FPGAs is presented in~\cite{7459526}, describing a technique to optimize its parameters including on-chip buffer sizes for a given set of resource constraint for modern FPGAs. 
Vitis AI~\cite{vitisai} is Xilinx's development platform for AI inference on Xilinx hardware platforms, consisting of optimized IP cores, tools, libraries, models, and example designs for both edge devices and Alveo cards.

%A comparison between the work presented here and NullHop is provided in Section~\ref{sec:fpgaporting}. 

Our approach is distinct from many of those above with its emphasis on being a completely open-source and multi-backend tool. In addition, a fully on-chip design is embraced in order to target the microsecond latency imposed in LHC physics experiments.

\section{Convolutional layers implementation in hls4ml}
\label{sec:large_CNN}

A direct implementation of a two-dimensional convolutional layer (Conv2D) requires six nested loops over image height $H$, width $W$, number of input channels $C$, number of output filters $N$, and filter height $J$ and width $K$~\cite{abdelouahab2018accelerating}.
In particular, calculating one element of the $V \times U \times N$ output tensor $\boldsymbol{Y}$ of a Conv2D layer from the $H \times W \times C$ input tensor $\boldsymbol{X}$, $J \times K \times C \times N$ weight tensor $\boldsymbol{W}$, and length-$N$ bias vector $\boldsymbol{\beta}$ requires three nested loops\footnote{Note that in practice $\boldsymbol{X}$ in equation \eqref{convLayer} is shifted by e.g. $\left(\frac{J+1}{2}, \frac{K+1}{2}\right)$ in order to be symmetric around $(v,u)$.},
\begin{align}
\label{convLayer}
\boldsymbol{Y}[v,u,n] &= \boldsymbol{\beta}[n] + \sum_{c=1}^{C} \sum_{j=1}^{J} \sum_{k=1}^{K} \boldsymbol{X}[v+j,u+k,c]\, \boldsymbol{W}[j,k,c,n]\,,
\end{align}
and repeating the calculation for all output elements $ \left\{ u, v, n \right\} \in \times \left[ 1,V \right] \times \left[ 1,U \right] \times \left[ 1,N \right]$ requires three additional nested loops.
For simplicity, we assume $J=K$ (square kernel) in the remainder of this paper.

Without additional optimizations, a plain implementation of these nested loops would result in high latency because, in the register transfer level (RTL) implementation, one clock cycle is required to move from an outer loop to an inner loop, and another one to move from an inner loop to an outer loop. 
This is usually addressed with loop pipelining. However, pipelining an outer loop requires completely parallelizing (or \emph{unrolling}) all nested inner loops, which significantly increases the size of the RTL implementation and the resources used. This approach is then feasible only for very small input sizes or model architectures. Utilizing this direct approach, the total number of unrolled loop iterations (the product $K^2VUN$) was limited to be less 4,096 to avoid the Vivado HLS partitioning limit.

While the direct implementation has the advantage of not requiring extra memory for temporary storage, on most modern computing architectures convolution is implemented using general matrix multiplication (GEMM), with algorithms like \texttt{im2col} and \texttt{kn2row}~\cite{gemmconv}. 
In im2col, each input window is flattened into a column vector and stacked together to form the input matrix, while the kernels are flattened into row vectors and concatenated to form the weight matrix. 
Matrix multiplication can then be performed using the accelerated library available on the platform, for example using the routines from basic linear algebra subprograms (BLAS). 
While this approach can be implemented on FPGAs using HLS, the design choices of \hlsfml, particularly the decision to store all tensors on the chip itself, mean this approach requires additional $\mathcal{O}(K^2HWC)$ units of memory, either as block random access memory (BRAM) or registers, to store the input matrix.
Additionally, to achieve the lowest latency, \hlsfml completely partitions the input arrays into individual registers as this allows access to each element within the same clock cycle. 
This strategy works well for fully connected layers but in case of convolutional layers the input tensor is usually much larger. Following this strategy, one would quickly reach the partitioning limit of Vivado HLS. 
Relaxing this constraint and using block or cyclic partitioning to create multiple array slices presents another challenge as the access pattern between consecutive layers of the model has to be consistent, otherwise scheduling issues arise and the design may fail to meet timing constraints.

To avoid these limitations, we have implemented convolutional layers using streams. 
Streams are synthesized in hardware as first in, first out (FIFO) buffers and as they do not require additional address management, they consume less resources than designs based on arrays. 
Since streams only allow sequential access, and have additional limitations on the reads and writes from different tasks (\texttt{C++} functions), this requires re-implementing most of the neural network layers in \hlsfml to support sequential processing. % which we will now describe.

Our implementation uses an approach similar to the im2col algorithm. However, it does not build the entire input matrix, and rather considers one column vector at a time. 
This allows us to reuse the existing matrix-vector multiplication functions of \hlsfml. 
In order to use streams for this implementation, a special \texttt{C++} class \texttt{hls::stream<>} provided in Vivado HLS is used. 
Given an $H \times W \times C$ input image tensor, we create a stream of $HW$ items where each item is an array containing the $C$ elements. 
This scheme allows us to efficiently read from the stream and construct column vectors. 
Because it usually takes one cycle to read or write one element of the stream, the latency of the layer will be at least $HW$ cycles.

\begin{figure}[th!]
    \centering
    \includegraphics[height=4.5cm,keepaspectratio]{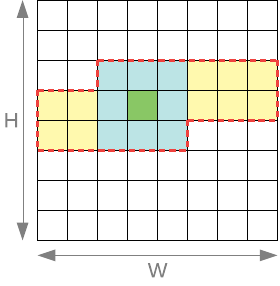}
    \hspace{2cm}
    \includegraphics[height=4.5cm,keepaspectratio]{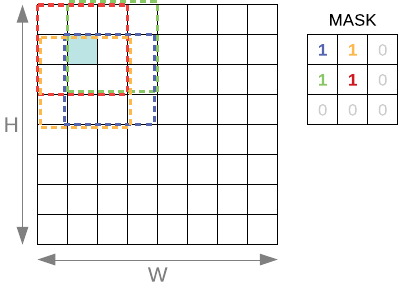}
    \caption{The image on the left shows an illustration of the sliding window buffer.
    All elements (yellow) of one kernel window (blue) are buffered to compute one output element (green). The right image shows the computation of the binary mask (an instruction) for one input element. The highlighted element (light blue) contributes four times to the sliding window in different positions, with the mask having bits set at the appropriate locations. 
    The bits of the mask are concatenated and stored in a 9-bit unsigned integer, in this example the number 27 (000011011 in binary).
    \label{fig:kernel_buffer}}
\end{figure}

Processing input sequentially through streams requires that we buffer all values that we wish to reuse at a later stage as an internal state. 
For a two-dimensional convolution, we need to buffer all values between the first and last element of the convolutional kernel, or \emph{sliding window}, as shown on the left in Fig.~\ref{fig:kernel_buffer}. 
The buffer can be defined as an array in \texttt{C++} and implemented by the HLS compiler as a shift register, however this approach requires keeping track of the position in the array, further complicating the implementation. 
We choose a simpler approach using streams. 
We create $K^2$ streams, corresponding to the size of the sliding window, and buffer values at the appropriate position in the window as they stream in. 
The depth of these streams is determined by the width of the output image and the square kernel size. 
Once we reach an element that is at the last position of a sliding window, we can compute one output by reading from the buffer. 
This is highly efficient as we can read the entire column vector in one clock cycle. 
With the column vector prepared, we can invoke the multiplication with the weight matrix and store the result in the output stream.

% \begin{figure}[t!]
%     \centering
%     \includegraphics[width=0.3\textwidth]{figures/cnn_buffer.pdf}
%     \caption{Illustration of sliding window buffer.
%     All elements of one kernel window (blue) are buffered to compute one output element (green).
%     \label{fig:kernel_buffer}}
% \end{figure}

\begin{figure}[th!]
    \centering
    \includegraphics[height=4.5cm,keepaspectratio]{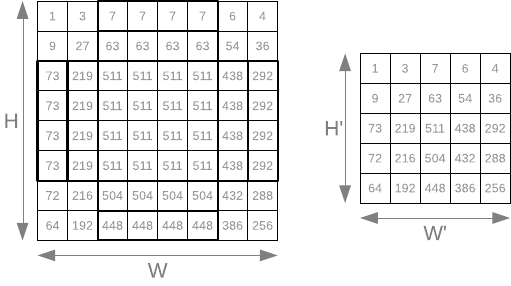}
    \caption{Example of compression of the instruction array. The left image shows the binary mask corresponding to each pixel, here represented as an integer rather than as a bit sequence. The shown values are specific of a $3 \times 3$ kernel with unit stride. Instruction duplicates are highlighted by the bold-line rectangles. On the right image, we reduce the instruction array by computing the instruction array of a $5 \times 5$ image, which has no duplicates, and translating the position of the element in the input array to the compressed array.\label{fig:compression}}
\end{figure}

While the algorithm described so far allows us to process larger inputs than a plain implementation would, significant resources are allocated for accounting, e.g. the position of the element in the sliding window or handling of the corners of the input image, and this prevents pipelining of loops at the desired latency. 
We address this issue by eliminating all branching code that handles these special cases, along with their associated state variables, leaving only the internal sliding window buffer. 
Instead, we pre-compute the positions in the sliding window where a given input element is used, and store this information as a binary mask, represented as a $K^2$-bit unsigned integer.
In the mask we set bits corresponding to every position in the sliding window where the input element is used, and leave the remaining bits unset (equal to $0$), as illustrated on the right in Fig.~\ref{fig:kernel_buffer}.
This mask can be used as an instruction on how to populate the sliding window buffer, eliminating the need for all branching code.
The procedure is applied to every element of the input image, and stored in the instruction array.
The instruction array can be significantly compressed by eliminating duplicates and translating the position of the element in the input array to the compressed array. 
As an example of the compression scheme, Fig.~\ref{fig:compression} illustrates how every convolution with a $3 \times 3$ kernel and unit stride can be represented with only $H'\times W'=5\times5$ instructions, regardless of the input image size ($H\times W$).

% \begin{figure}
%     \centering
%     \includegraphics[width=0.47\textwidth]{figures/cnn_mask.pdf}
%     \caption{Computing the instruction array. The highlighted element (light blue) contributes to four sliding windows, with the mask having bits set at the appropriate locations. 
%     The bits of the mask are concatenated and stored in a 9-bit unsigned integer, in this example the number 27 (000011011 in binary).\label{fig:mask}}
% \end{figure}

For the pooling layers, a similar technique is used to collect the data in sliding window buffers. 
As the most common form of pooling assumes a stride equal to the pooling region size (i.e. no overlaps between pooled regions), we create a simpler and more optimal instruction-encoding scheme for this case. 
Unlike convolution, in both max and average pooling operation, we do not need the position of elements in the sliding window, only which window they belong to. 
This allows us to create a simple lookup table of $H+W$ elements without the need for translation of the position of the input element.

\section{Dataset}
\label{sec:data}

\begin{figure}[h!]
    \centering
    \includegraphics[width=0.45\textwidth]{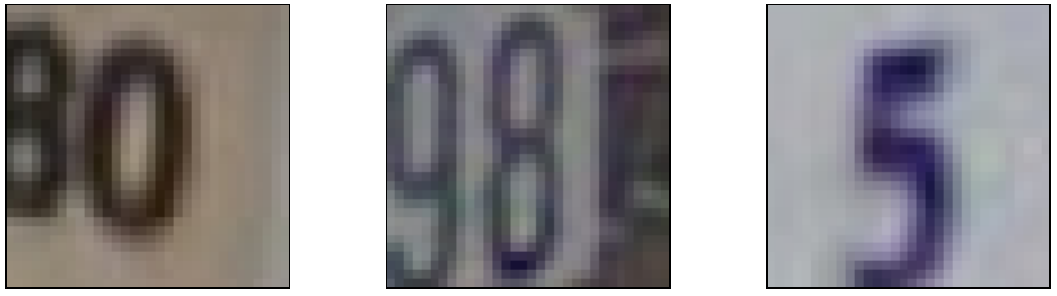}
        \hspace{1 cm}
    \includegraphics[width=0.45\textwidth]{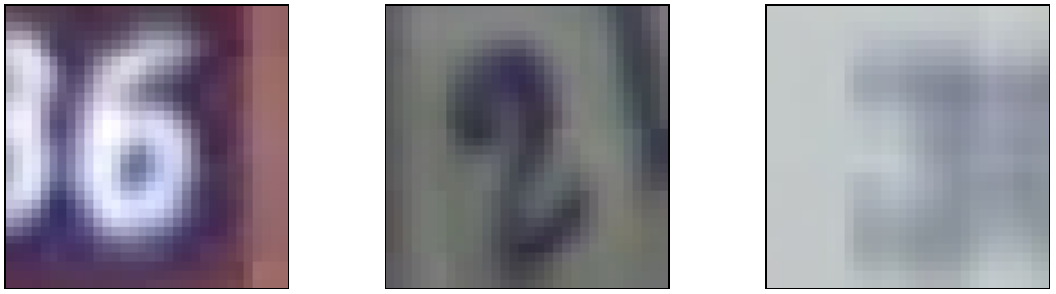}
    \caption{Examples of digit images extracted from the SVHN train (three leftmost images) and test (three rightmost images) datasets.\label{fig:SVHN_example}}
\end{figure}

To demonstrate the functionality of CNNs in \hlsfml, we consider as a benchmark example a digit classifier trained on the Street View House Numbers (SVHN) Dataset~\cite{Netzer2011}. The SVHN dataset consists of cropped real-world images of house numbers extracted from Google Street View images, in a format similar to that of the MNIST~\cite{MNISTdata} dataset. 
However, it presents a much more challenging real-world problem, as illustrated by the examples shown in Fig.~\ref{fig:SVHN_example}. 
The numbers are part of natural scene images, possibly with other digits appearing as a background on the two sides of the central one, different colors and focus, orientation, etc.

All the images are in RGB format and have been cropped to $32\times 32$ pixels. 
Unlike MNIST, more than one digit can be present in the same image. 
In these cases, the center digit is used to assign a label to the image, which is then used as ground truth when training the classifier. 
Each image can belong to one of 10 classes, corresponding to digits ``0'' through ``9.''
As a preprocessing step, we divide each pixel by the max RGB value of 255 in order to have numbers in the range between zero and one. We then standardize the input images to have a mean of zero and unit variance by applying a per-pixel scaling factor computed from the full training dataset. The same scaling is applied to the test set.

The SVHN dataset consists of 604,388 images for training (of which 531,131 are considered extra data that are slightly easier to classify) and 26,032 images for testing. 

Training is performed using a k-fold cross-validation procedure. The training dataset is split in 10 training and validation samples such that 10\% of the training set is used for validation and the remaining 90\% for training. 
Training and validation is then repeated $k$ times until each fold of the training set has been used to evaluate the accuracy of the model. 
Model-performance figures of merit (e.g., accuracy, true and false positive rates, etc.) are defined considering the mean across the 10 folds on the test set. 
The corresponding uncertainty is quantified through the standard deviation across the 10 folds.

\section{Baseline model}
\label{sec:model}

\begin{figure}[t!]
    \centering
    \includegraphics[width=\textwidth]{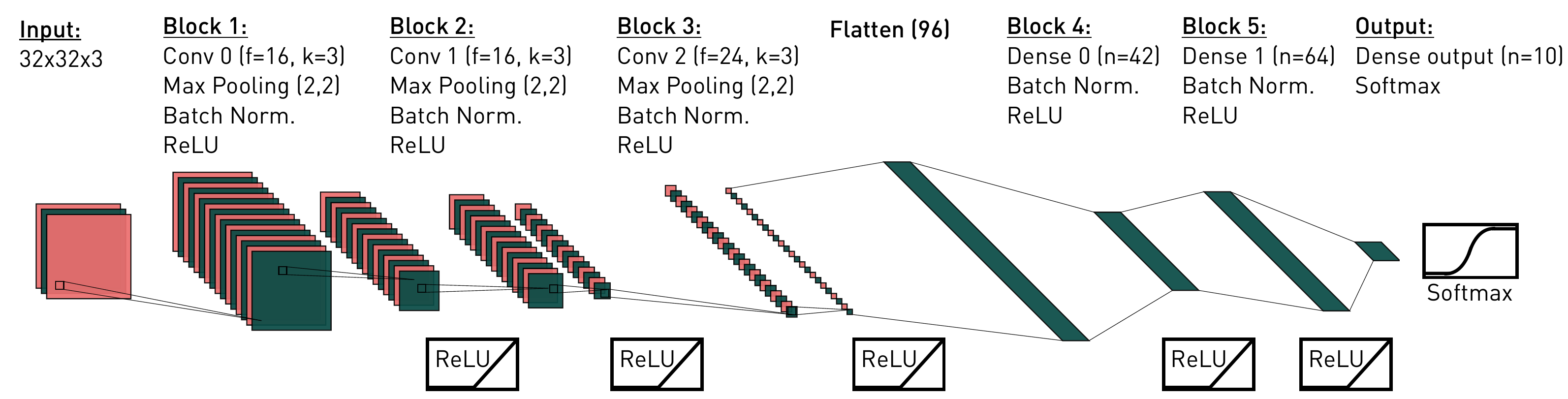}
    \caption{The neural network architecture, chosen through a Bayesian optimization over the hyperparameters, for classifying digits from the SVHN dataset.
    Each convolutional block consists of a convolutional layer, max pooling, batch normalization, and ReLU activation.
    The convolutional layers in the three convolutional blocks use 16, 16, and 24 filters, respectively, and each has a kernel size of $3\times 3$. 
    The pooling layers have a size of $2\times 2$. 
    The convolutional blocks are followed by two fully-connected layers consisting of 42 and 64 neurons, with batch normalization and ReLU activation. 
    The bias term is removed from all layers except the final output layer.
    \label{fig:architecture}}
\end{figure}
 \begin{table}[th!]
  \centering
 \caption{Number of trainable weights, floating-point operations, energy consumption and layer size in bits for each convolutional or dense layer (not including the activation layers). Batch normalization and pooling layers are not included as they are negligible in size and energy consumption in comparison. The energy is estimated assuming a 45 nm process using \QTools. The total energy and bit size includes all model layers.
 \label{tab:flopsPerLayer}}
\begin{tabular}{lcccccc}
Layer name & Layer type & Input shape & Weights & MFLOPs & Energy [nJ]& Bit size \\
\hline
\texttt{Conv 0}      & Conv2D & (32, 32, 3)  & 432  & 0.778    & 1,795 & 3,456\\
% \texttt{Max Pool 0}  & Conv2D & (30, 30, 16) & -    & 0.014    &  - \\
\texttt{Conv 1}      & Conv2D & (15, 15, 16) & 2,304 & 0.779  & 1,802 & 18,432\\
% \texttt{Max Pool 1}  & Conv2D & (13, 13, 16) & -    & 0.002  &  - \\
\texttt{Conv 2}      & Conv2D & (6, 6, 16)   & 3,456 & 0.110  & 262 & 27,648\\
% \texttt{Max Pool 2}  & Conv2D & (2, 2, 24)   & -    & 0.0003 &  - \\
\texttt{Dense 0} & Dense  & (96)             & 4,032 & 0.008  & 26  & 32,256\\
\texttt{Dense 1} & Dense  & (42)             & 2,688 & 0.005  & 17 & 21,504\\
\texttt{Output}   & Dense  & (64)      & 65   & 0.001  & 4 & 5,200\\ 
\hline
 \multicolumn{3}{l}{\textbf{Model total}}&\multicolumn{1}{c}{\textbf{12,858}}& \multicolumn{1}{c}{\textbf{1.71}} & \textbf{3,918} & \textbf{170,816} \\
\end{tabular}
\end{table} 

%For the lowest possible latency, each layer should have a maximum number of trainable parameters of 4096. 
%This is due to fixed limits in the Vivado compiler, beyond which maximally unrolled compilation will fail.
%Additionally, 
Keeping in mind that the model is designed for deployment on the resource limited environment of an FPGA, we limit the depth and complexity of the baseline model while preserving reasonable performance. As a target, we aimed at a test error close to 5\%, where state-of-the-art test error lies between 1--5\%~\cite{svhn1,svhn2,svhn3,svhn4,svhn5,svhn6}. 
To reduce the overall model latency as much as possible, models with fewer large layers (wider) are preferred over models with several smaller layers (deeper). This is due to the parallel nature of the FPGA, making it more resource-efficient to process one large layer in parallel over several small ones sequentially. The dependency of inference latency and resource consumption for increasing depth and width will be further discussed in Section~\ref{sec:fpgaporting}. 

A Bayesian optimization over the model hyperparameters is performed using \textsc{Keras Tuner}~\cite{kerastuner}. 
The first few layers are chosen to be 2D convolutional blocks. Each block consists of a convolutional layer followed by a max pooling layer, a batch normalization~\cite{bn} layer, and a rectified linear unit (ReLU)~\cite{relu1,relu2} activation function. The optimization range is set so that the maximum number of loop iterations per layer is below the unroll limit described in Section~\ref{sec:large_CNN} in order to achieve the lowest possible latency. Pooling layers are used to keep the size of the final dense layers small. 

The convolutional blocks are followed by a series of fully-connected layers, the amount of layers and their size again determined through the hyperparameter optimization. 
A final ten-node dense layer, activated by a softmax function, returns the probability for a given image to be assigned to each of the ten classes.
The result of the Bayesian optimization, shown in Fig.~\ref{fig:architecture}, consists of three convolutional blocks and two dense layers. 
The convolutional layers in the three blocks have 16, 16, and 24 filters, respectively, and each has a kernel size of $3\times 3$. 
The pooling layers have a size of $2\times 2$. 
The two hidden dense layers consist of 42 and 64 neurons, with batch normalization and ReLU activation. 
The model is implemented in \TensorFlow~\cite{tensorflow2015-whitepaper}, using the \Keras API~\cite{chollet2015keras}. 
To reduce the number of required operations, the bias term is removed from all layers, except for the final output layer, while keeping batch normalization on to prevent internal covariate shift~\cite{bn}. 

We refer to this model as the Baseline Floating-point (BF) model. 
The number of floating-point operations (FLOPs) and weights for each convolutional or dense layer is listed in Table~\ref{tab:flopsPerLayer}. 
In addition, an estimate of the per-layer energy consumption and the layer size in bits is quoted. These estimates are obtained using \QTools~\cite{Coelho:2020zfu}, a library for estimating model size and energy consumption, assuming a 45\unit{nm} process~\cite{horowitz}. Despite the first dense layer having the most weights, the number of FLOPs and the energy consumption is significantly higher in the convolutional layers due to the much larger number of multiply-accumulate operations performed.
The per-layer summaries does not include results for batch normalization or pooling layers, as the contribution from these are negligible in comparison. The total model energy and bit size includes all layers of the model. 

The training is performed minimizing the categorical crossentropy loss~\cite{Goodfellow-et-al-2016} using the Adam optimizer~\cite{kingma2017adam}. 
The optimal learning rate is obtained using the hyperparameter optimization described above, found to be 0.003, and is set as the starting learning rate. 
If there is no improvement in the loss for five epochs, the learning rate is reduced by 90\% until a minimum learning rate of $10^{-6}$ is reached.
The batch size is 1,024 and the training takes at most 100 epochs. 
Early stopping is enabled when no improvement in the validation loss is observed across ten epochs.

\section{Compression by pruning}
\label{sec:pruning}

\begin{figure}[t!]
    \centering
    \includegraphics[width=0.47\textwidth]{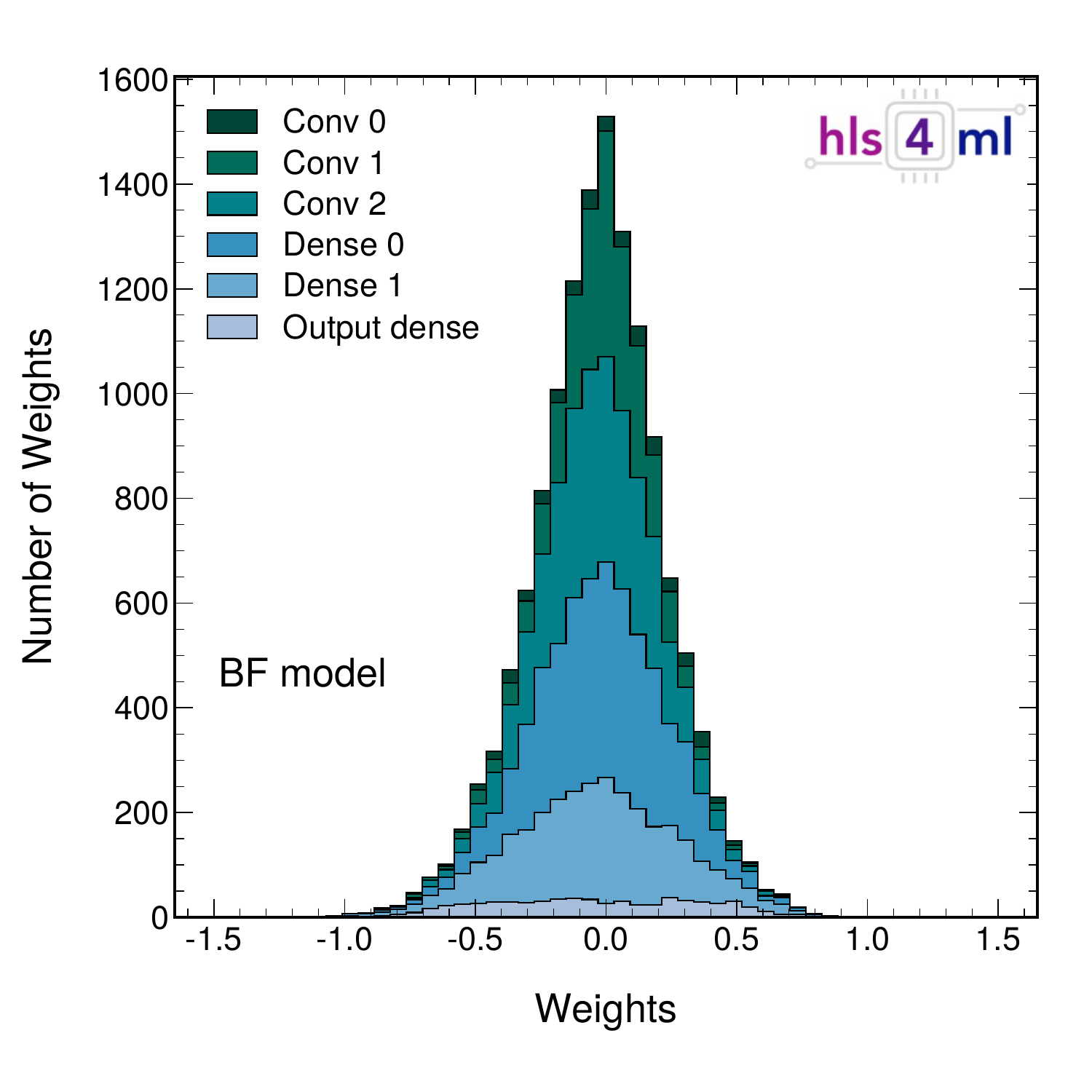}
    \includegraphics[width=0.47\textwidth]{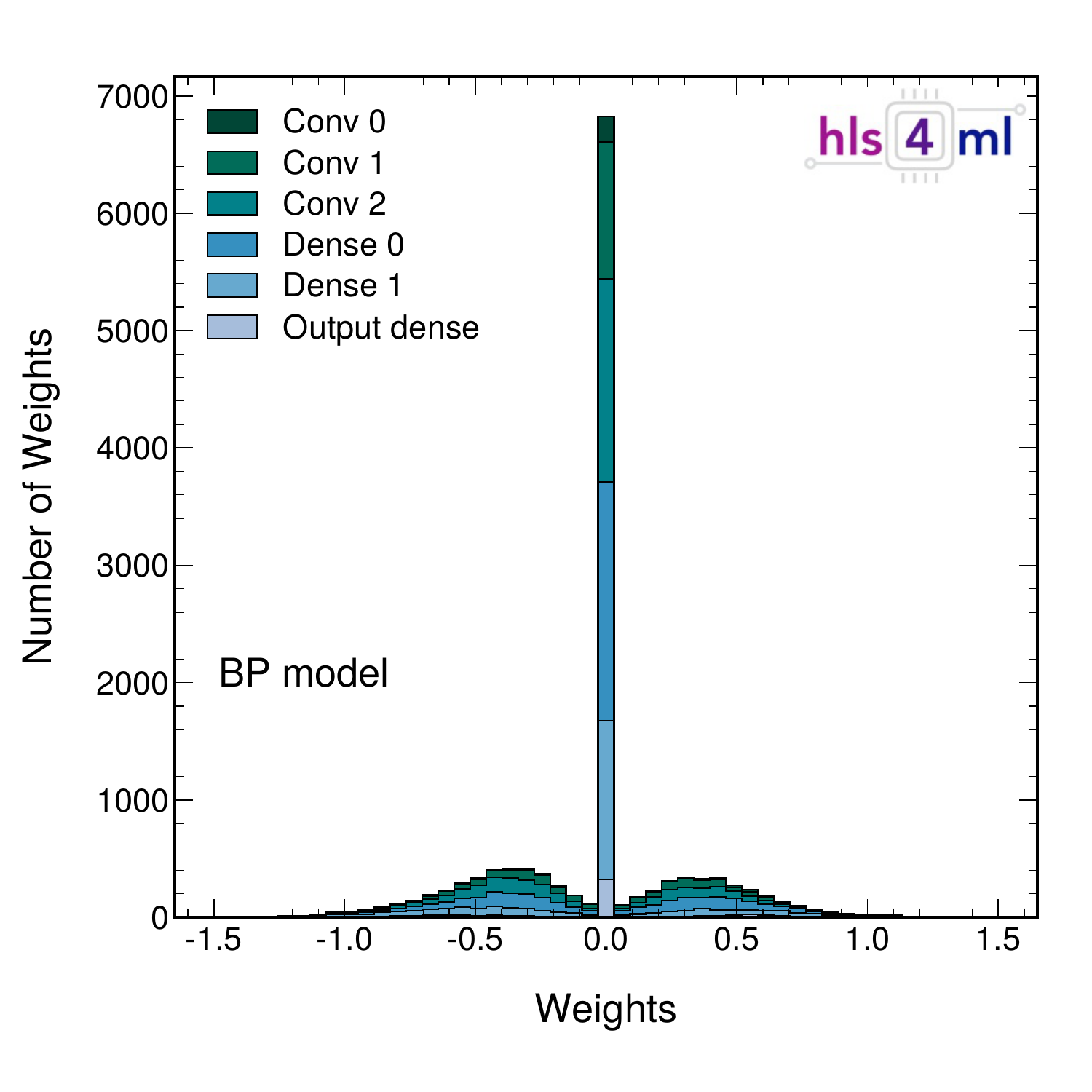}
    \caption{The weights per layer for the the Baseline Floating-point (BF) model (left) and the Baseline Pruned (BP) model (right). 
    The BP model is derived by starting from the BF model and repeating the training while applying a pruning procedure with a target sparsity of 50\% for each layer.\label{fig:weights_pruning}}
\end{figure}

Weight pruning is an established strategy to compress a neural network and consequently reducing its resource utilization. 
One strategy, magnitude-based pruning, consists of eliminating redundant weights in the weight tensors by setting the value of the smallest weights in a tensor to zero~\cite{DBLP:journals/corr/HanMD15,NIPS1989_250,2017arXiv171201312L, DBLP:journals/corr/HanPTD15,DBLP:journals/corr/YangCS16a,Duarte:2018ite}. 
All zero-weight multiplications are omitted by the HLS library when translating the network into firmware, consequently saving significant FPGA resources.
% The \hlsfml doesn't perform any specific operation to prune the zero-value weights. 
% Instead, it relies on the underlying optimization applied by the HLS library, which is designed to omit $\times 0$ operations. 
 
% Pruning can be done either (1) after training, setting to zero 
% all weights below some fixed threshold or below some percentile of the weight distribution, or (2) during training, where small values are set to zero based on their magnitude in an iterative way. 
% Post-training pruning is more effective when the weights are encouraged to assume small values through $L_1$ regularization in the loss function during training and then setting to zero these weights once the training is complete. 
% This process can be repeated until the target sparsity is achieved. 
% Previous applications of this procedure in the context of \hlsfml development are discussed in~\cite{Duarte:2018ite}.

% While applying post-training pruning in a cycle of training iterations allows the network to adapt to the progressive reduction of available weights, in-training pruning provides even more adaptive power, generally resulting in better performance. In this work, we focus on in-training pruning.
% % and is therefore the method of choice for this study.

Pruning is enforced using the \TensorFlow pruning API, a \Keras-based interface consisting of a simple drop-in replacement of \Keras layers. 
A sparsity of 50\% is targeted, meaning only 50\% of the weights are retained in the pruned layer and the remaining ones are set to zero. 
Before pruning, the weights of each layer are initialized to the weights of the corresponding model without pruning (i.e. \emph{fine-tuning pruning}), ensuring the model is in a stable minimum before removing weights deemed unimportant. 
Each model is pruned starting from the 10th epoch, with the target sparsity gradually increasing to the desired 50\% with a polynomial decay of the pruning rate~\cite{zhu2017prune}.

By pruning the BF model layers as listed in Table~\ref{tab:flopsPerLayer} to a target sparsity of 50\%, the number of FLOPs required when evaluating the model, can be significantly reduced. We refer to the resulting model as the Baseline Pruned (BP) model.

The distribution of the weight values per layer for the BF and BP models are shown in Figure~\ref{fig:weights_pruning}. 
The effect of pruning is seen by comparing the two distributions: the smallest magnitude weights of the BF weight distribution migrate to the spike at zero in the BP weight distribution, while the two tails remain populated, with most of the weights falling in the interval $[-1.5, 1.5]$.

\begin{figure}[t!]
    \centering
    \includegraphics[width=0.47\textwidth]{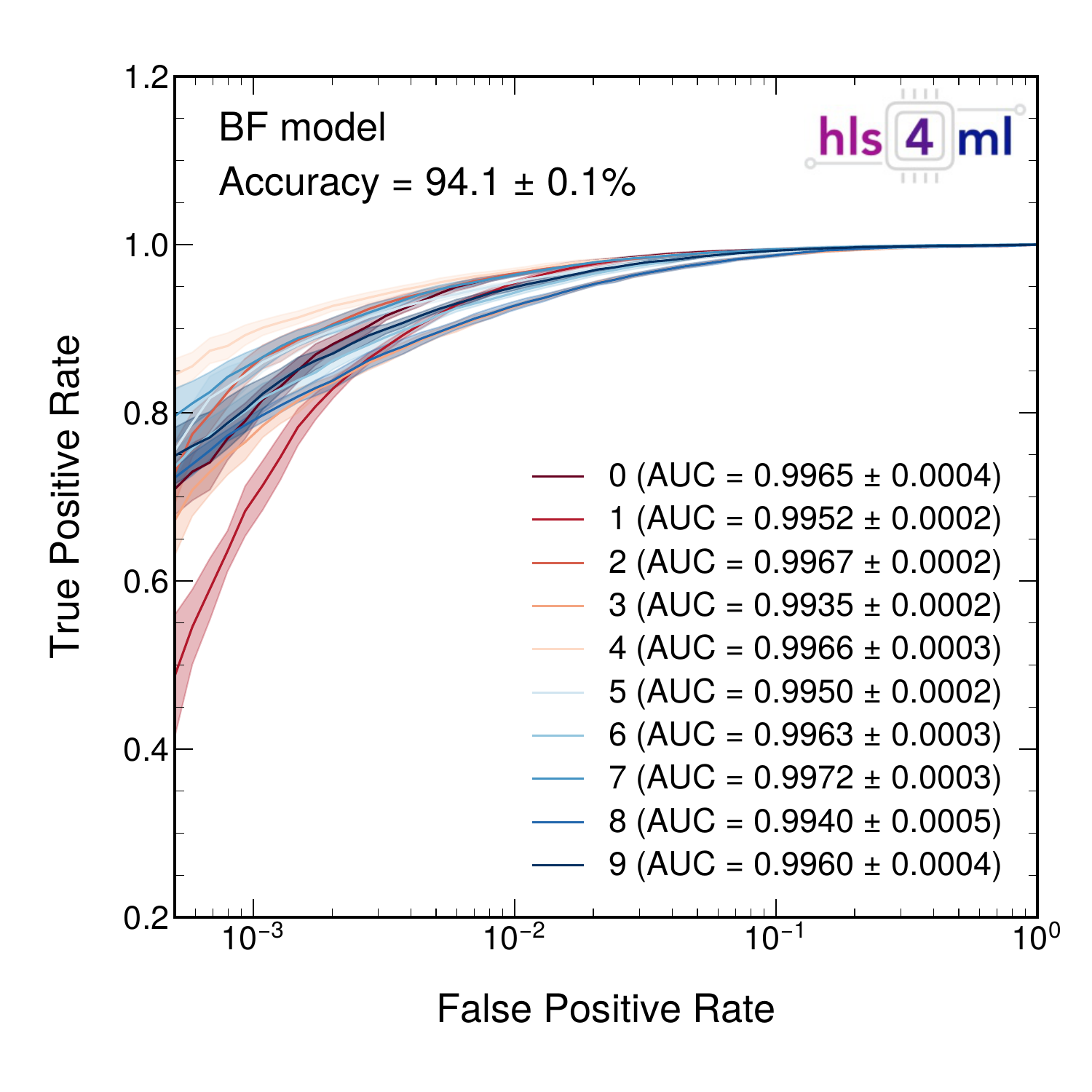}
    \includegraphics[width=0.47\textwidth]{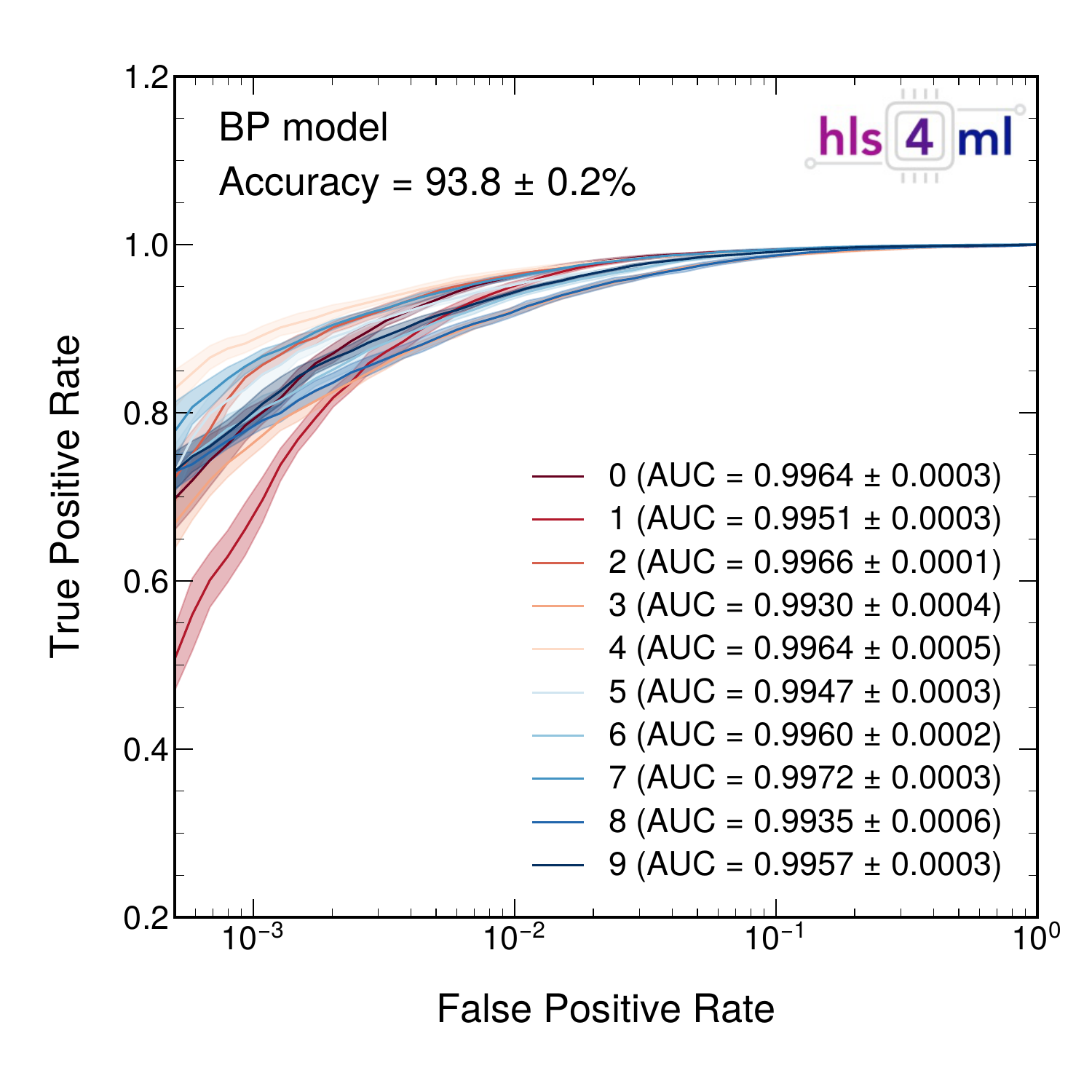}
    \caption{ROC curves of false positive rate (FPR) versus true positive rate (TPR) for the Baseline Floating-point (BF) model (left) and the Baseline Pruned (BP) model (right).
    Training is performed using $k$-fold cross validation, with $k=10$. 
    For each digit, the solid line corresponds to the mean and the band to the standard deviation across the 10 folds. The area under the curve (AUC) is reported in the legend and is defined as the mean across the ten folds with an uncertainty given by the standard deviation.\label{fig:roc_baseline}}
\end{figure}

Figure~\ref{fig:roc_baseline} compares the classification performance of the BF and BP models. 
Specifically, it shows the receiver operating characteristic (ROC) curves and the corresponding area under the curve (AUC) for each digit. 
In addition, we consider the model accuracy, i.e. how often the predictions (after taking the $\argmax$ of the output neurons) equals the labels.
%, defined as $(1-FPR+TPR)/$ where $FPR$ ($TPR$) is the false (true) positive rate.
For each ROC, the solid line corresponds to the mean across the 10 folds and the uncertainty to the standard deviation. The mean accuracy and standard deviation across the 10 folds is also reported on the plot. 
Despite removing 50\% of the weights for the BP model, the model accuracy is comparable between the BF and BP models.

These models serve as our reference models. 
The accuracy, latency and resource consumption of these will be discussed in Section~\ref{sec:fpgaporting}. 
In general, we observe a significant reduction in FPGA resource consumption for pruned models, as zero-weight multiplications are optimized away by the HLS compiler. 
Because pruning has little impact on the model accuracy (as demonstrated in Figure~\ref{fig:roc_baseline}), pruning is always recommended before translation into FPGA firmware with \hlsfml.

\section{Compression by quantization}
\label{sec:quantization}

To further limit the model footprint, we reduce the numerical precision of the model weights before FPGA deployment.
During training, one typically relies on single- or double-precision floating-point arithmetic, i.e. 32 or 64 bit precision.
% Specifically, all training described in this paper were carried out using single-precision floating-point arithmetic on a GeForce RTX 2080 Ti Nvidia GPU.
However, when deploying a deep neural network on FPGA, reduced precision fixed-point arithmetic (quantization) is often used in order to minimize resource consumption and latency. 
It has been shown that deep neural networks experience little accuracy loss when QAT is applied, even up to binary quantization of weights~\cite{DBLP:journals/corr/HubaraCSEB16}.

When a quantized model is deployed on an FPGA, all its weights, biases, and activation functions are converted to fixed-point precision before being deployed. This is referred to as post-training quantization (PTQ). The chosen precision is a new tunable hyperparameter.
The \hlsfml library allows users to specify different numerical precisions for different components of the network (known as \emph{heterogeneous quantization}). 
For instance, it is found that severe PTQ of the activation functions typically results in a greater reduction of accuracy than severe PTQ of the weights~\cite{DiGuglielmo:2020eqx}.
By default, \hlsfml assumes 16 total bits for every layer, 6 of which are dedicated to the integer part ($\langle 16, 6\rangle$ precision).

In this paper, we consider two approaches to network quantization: PTQ of a floating-point model, and QAT, resulting in a model already optimized for fixed-point precision. 
Both methods will be described in the following and the result on hardware discussed in detail in Section~\ref{sec:fpgaporting}. 
To summarize, we observe a significant reduction in accuracy using PTQ, with no prediction power remaining below a bit width of 14. 
Using QAT, however, high accuracy is maintained down to extremely narrow bit widths of 3--4. 
The latency and resource consumption are similar for the two methods (with certain caveats that will be discussed in Section~\ref{sec:fpgaporting}), and QAT is therefore the preferred solution for model quantization before deployment with \hlsfml.

\subsection{Post-training quantization}

The \hlsfml library converts model weights and biases from floating-point to fixed-point precision, applying the same quantization to the whole network or setting the precision per layer and per parameter type. 
% By default, \hlsfml assumes 16 total bits, 6 of which are dedicated to the integer part ($\langle 16, 6\rangle$ precision).
Bit width and number of integer bits must be tuned carefully to prevent compromising the model accuracy. 
For each component, an appropriate precision is selected by studying the floating-point weight profiles, i.e. the range of input or output values spanned by the testing data for the trained model, component by component. 
In order to minimize the impact of quantization on accuracy, the precision can be tuned so that the numerical representation adequately covers the range of values observed in the floating-point activation profile. 

\begin{figure}[t!]
    \centering
    \includegraphics[width=0.47\textwidth]{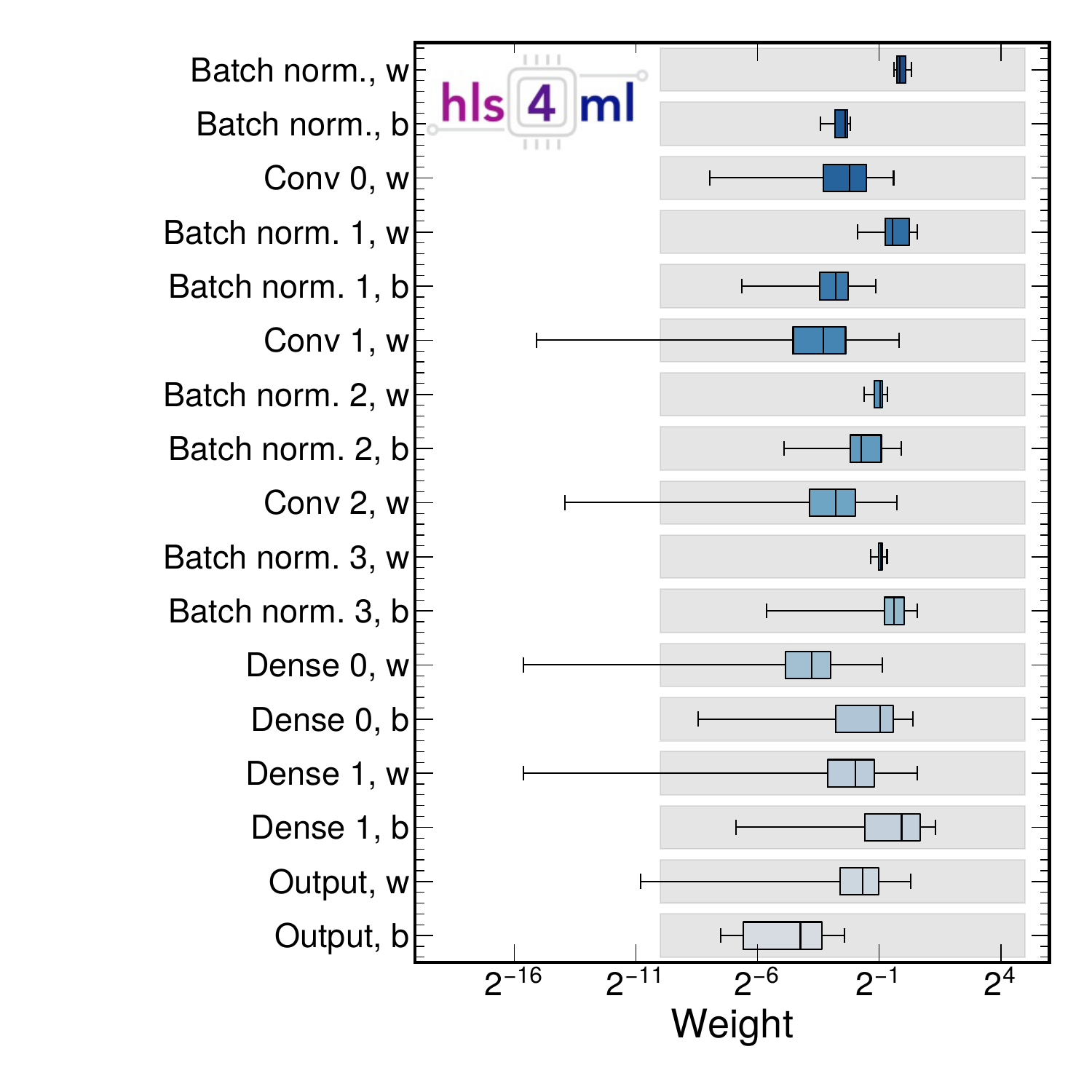}
    \includegraphics[width=0.47\textwidth]{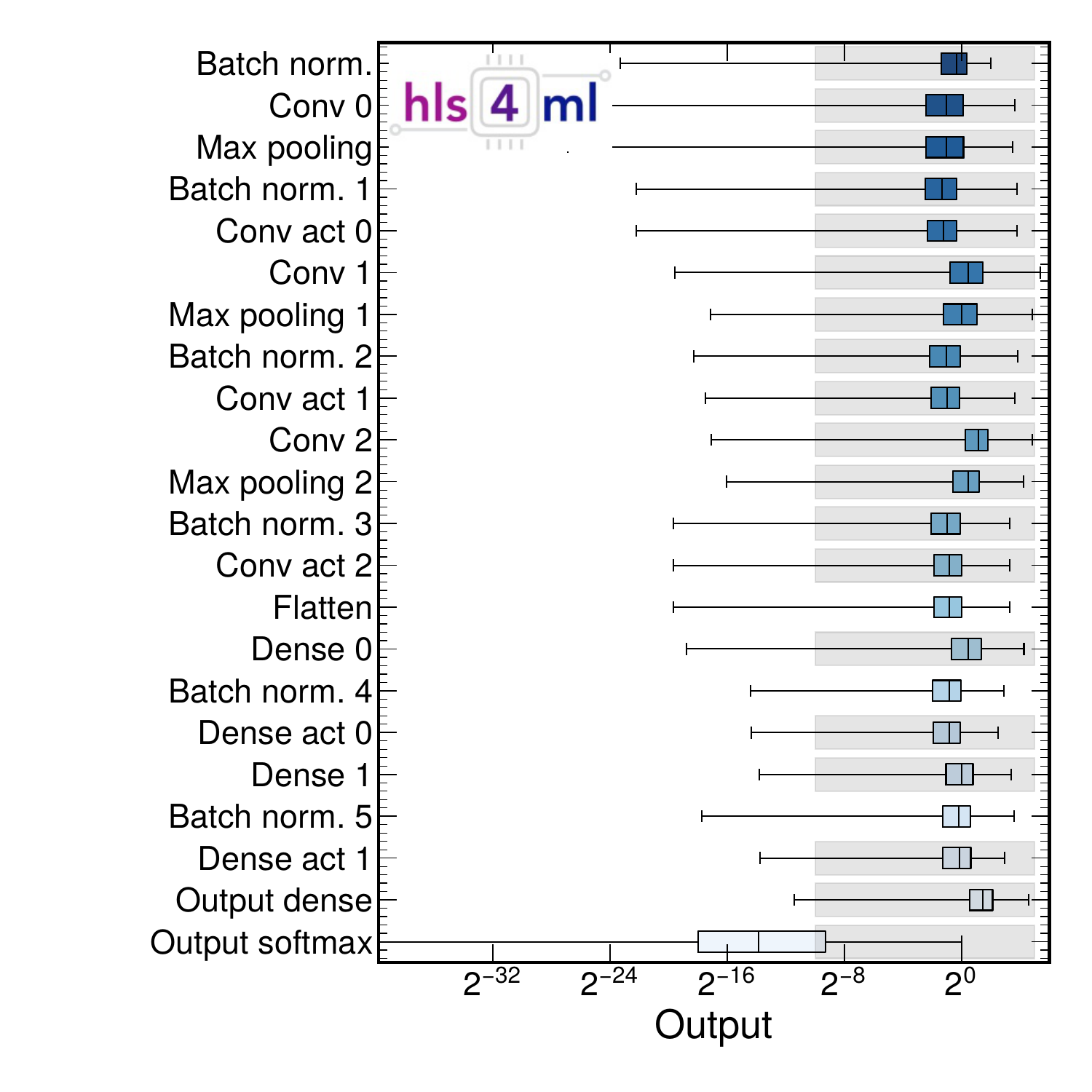}\\
    % \includegraphics[width=0.47\textwidth]{figures/Profile_pruned_full_0_weights.pdf}
    % \includegraphics[width=0.47\textwidth]{figures/Profile_pruned_full_0_activations.pdf}
        % The range covered by post-training quantization for a $\langle 16,6 \rangle$ precision is shown in gray.
    \caption{Layer weight (left) and output (right) numerical values per layer for the Baseline Floating-point (BF) model for a subset of the test data. The gray band represents the coverage of the default precision of $\langle 16, 6\rangle$.
 \label{fig:baseline_weight_profile}}
\end{figure}

As an example, the by-layer weight profiles of the BF model is shown in Fig.~\ref{fig:baseline_weight_profile}, for both layer weights (left) and outputs (right) using the testing data. 
The last letter in the label indicates which type of weight is being profiled, where $w$ is for weights and $b$ is for bias. 
Learnable bias parameters are only included in the final dense layer. 
The other bias terms are introduced by the fusing of a batch normalization and a dense layer. 
The grey bands illustrate the numerical range covered by the default $\langle 16,6 \rangle$ precision. 
No gray band is visible for the Flatten layer as no operations changing the data is performed. 
Also no gray band is visible for the 4th and 5th batch normalization layer outputs as these are fused with the dense layers.

Typically, extreme PTQ results in a sizeable accuracy loss. 
The increased spacing between representable numbers enforces a severe weight rounding that leads to a significant reduction in the model accuracy once the resolution becomes too coarse. 
The amount of compression one can reach by this procedure is balanced by the need to preserve the model accuracy, and how much model accuracy reduction can be tolerated is an application-specific question.

We use PTQ to generate a range of compressed models, further discussed in Section~\ref{sec:fpgaporting}, scanning bit widths from 16 to 1, with 6 integer bits.

\begin{figure}[t!]
    \centering
    \includegraphics[width=0.47\textwidth]{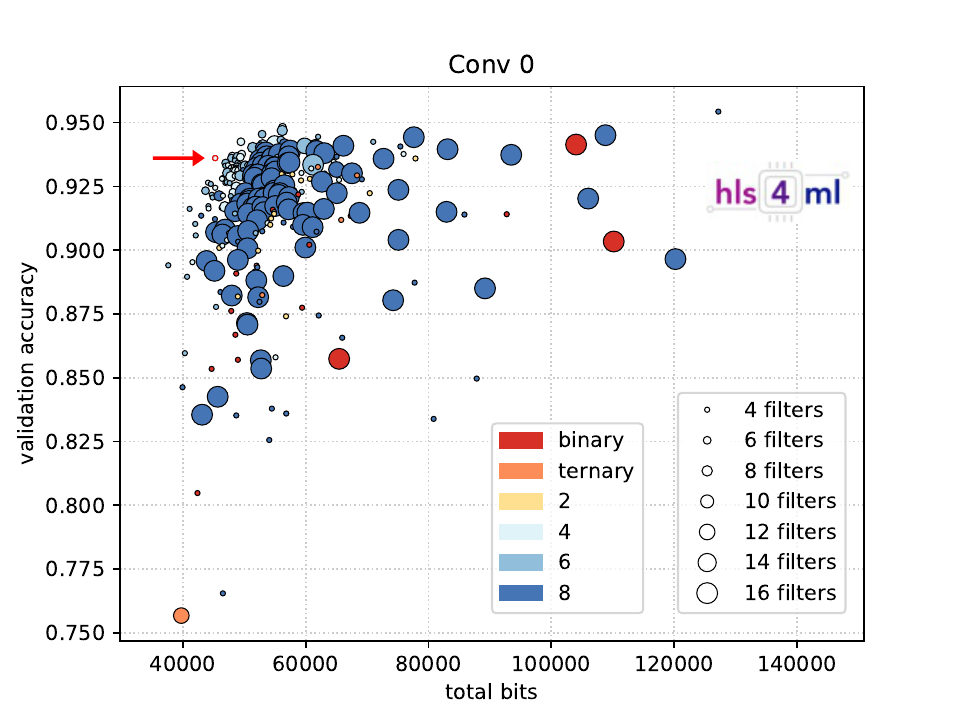}
    \includegraphics[width=0.47\textwidth]{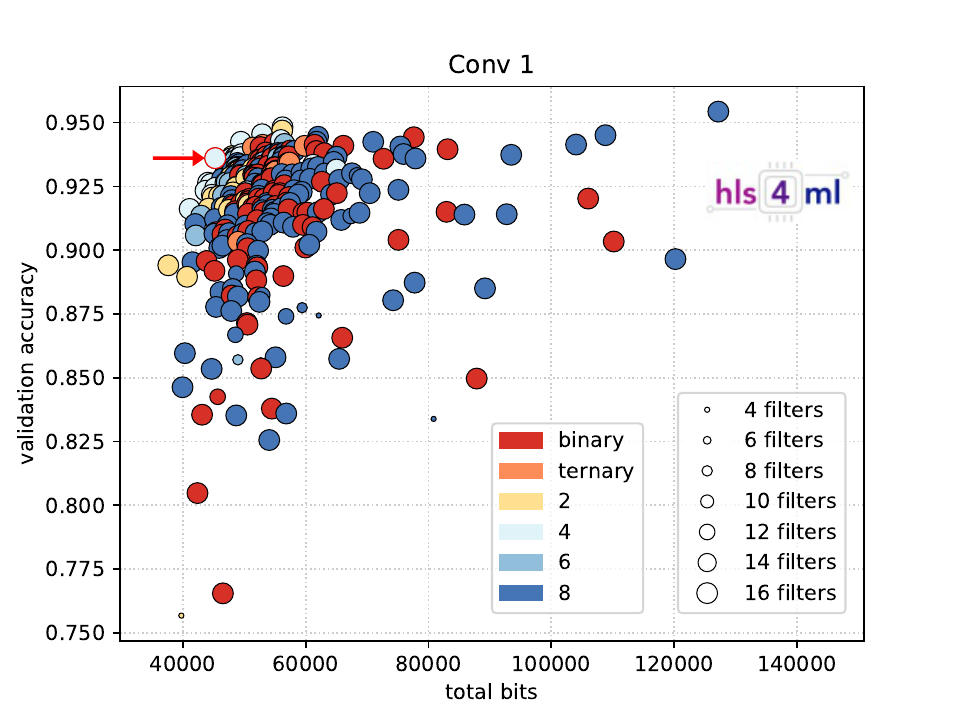} \\
    \includegraphics[width=0.47\textwidth]{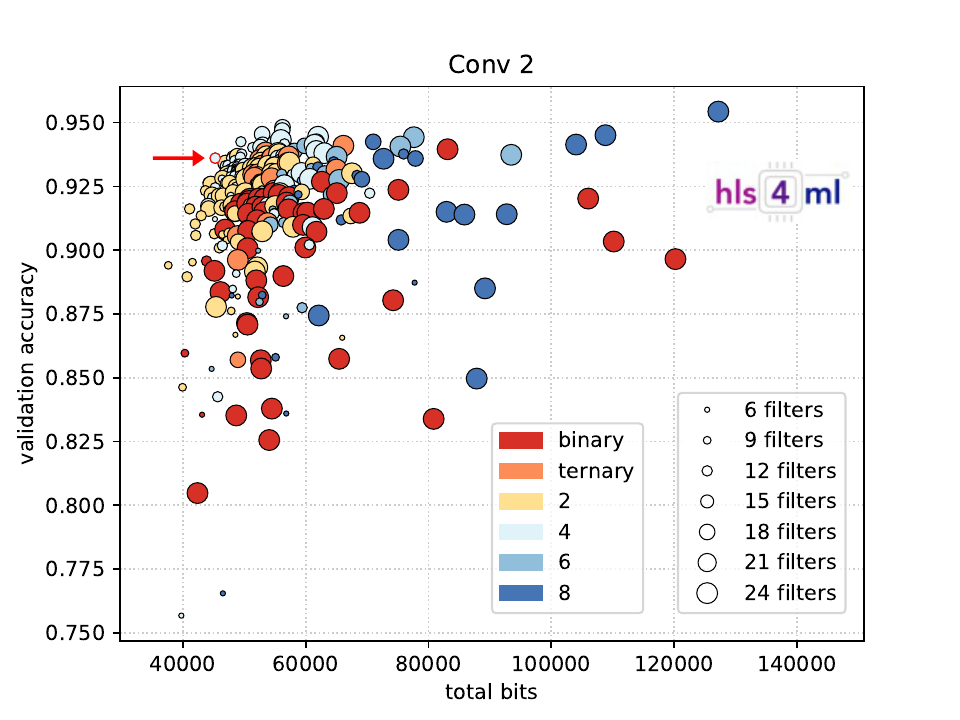}
    \caption{Relation between accuracy and model bit size for an ensemble of trial models, resulting from a Bayesian optimization performed over layer quantizers and number of filters, using \AutoQKeras. 
    Each figure corresponds to a given quantization and filter configuration tested for the first (top left), second (top right) and third (bottom) convolutional layer. 
    The size of each marker corresponds to the number of filters tested for that layer, and the color to the quantization (binary, ternary or mantissa quantization). 
    The red arrow indicates the model yielding the best accuracy versus size trade-off.}
     \label{fig:autoq}
\end{figure}

\subsection{Quantization-aware training}

QAT~\cite{DBLP:journals/corr/abs-1712-05877} is an efficient procedure to limit accuracy loss while reducing the numerical precision of the network components.
Here, quantized weights and biases are used in the training during the forward pass, while full precision is used in the backward pass in order to facilitate the drift towards the optimal point in the loss minimization (known as the \emph{straight-through estimator})~\cite{NIPS2015_5647}. 
The \hlsfml library supports QAT through its interface to \QKeras~\cite{Coelho:2020zfu}. 

We train a range of quantized \QKeras models using the same architecture as in Fig.~\ref{fig:architecture}, imposing a common bit width across the model. 
We scan the bit width from 16 to 3, as well as train a ternary and a binary quantized model. 
We refer to these models as QKeras (Q) models. In addition, we train pruned versions of these models, targeting a sparsity of 50\%. These are referred to as QKeras Pruned (QP) models.

Only convolutional layers, dense layers, and activation functions are quantized. 
The batch normalization layers are not quantized during training, as support for the \QKeras quantized equivalent of the \Keras batch normalization layer is not supported in \hlsfml at the time of this writing. 
Support for this is planned for a future version of \hlsfml. 
Batch normalization layers in the QAT models are therefore set to the default precision of $\langle 16,6 \rangle$ by \hlsfml.
The final softmax layer is also kept at the default precision  of $\langle 16,6 \rangle$ in order to not compromise the classification accuracy.

Finally, we define a heterogeneously quantized model using \AutoQKeras~\cite{Coelho:2020zfu}, a library for automatic heterogeneous quantization. 
The \AutoQKeras library treats the layer precision as a hyperparameter, and finds the quantization
which minimizes the model bit size while maximizing the model accuracy. 
By allowing \AutoQKeras to explore different quantization settings for different parts of a given network, we obtain an optimal heterogeneously quantized \QKeras model.
A Bayesian optimization is performed over a range of quantizers available in \QKeras, targeting a 50\% reduction in model bit size. 
At the same time, the number of filters per convolutional layer and neurons per dense layer is re-optimized as quantization tends to lead to a preference for either (1) more filters as information is lost during quantization or (2) less filters due to some filters effectively being the same after quantization.

The optimization process is shown in Fig.~\ref{fig:autoq}, where the model bit size versus the model validation accuracy is shown for all the models tested in the automatic quantization procedure, showing the different quantization configurations for each of the convolutional layers. The size of the markers correspond to the number of filters used for a given convolutional layer in that trial. The colors correspond to different type of quantizers (binary, ternary of mantissa quantization using different bit widths).
The model yielding the best accuracy versus size trade-off is marked by a red arrow. The number of filters per convolutional layer for the selected model is (4, 16, 12), compared to the original (16, 16, 24) for the BF and BP models, and the number of neurons per dense layer is (15, 16) compared to (42, 64) in the original model.  
Table~\ref{tab:autoq} summarizes the quantization configuration found to be optimal by \AutoQKeras, and the corresponding model energy consumption estimated using \QTools. We note that this model uses almost 90\% less energy than the original. 
We train two versions of this model: an unpruned version (AQ), and a pruned version (AQP). 
The latter model is the same as AQ, but additionally pruned to a target sparsity of 50\%.

\begin{table}[t!]
 \caption{Per-layer heterogeneous quantization configuration obtained with \AutoQKeras, the total estimated energy consumption and model size in bits of the AutoQ (AQ) model. 
 The energy is estimated assuming a 45 nm process using \QTools.\label{tab:autoq}}
  \centering 
\resizebox{\textwidth}{!}{
\begin{tabular}{c|cccccccccccc|c|c}
\multirow{2}{*}{Model} & \multicolumn{12}{c}{Precision per layer} & Energy [nJ] & Bit size\\
 & Conv2D         & ReLU    & Conv2D   & ReLU     & Conv2D & ReLU & Dense & ReLU & Dense & ReLU & Dense & Softmax\\
\hline
AQ & $\langle4, 0\rangle$ & $\langle3, 1\rangle$ & $\langle4, 0\rangle$ & $\langle3, 1\rangle$ & $\langle4, 0\rangle$ & $\langle8, 4\rangle$ & $\langle4, 0\rangle$ & $\langle4, 2\rangle$& $\langle4, 0\rangle$& $\langle8, 2\rangle$ & $\langle6, 0\rangle$ & $\langle16, 6\rangle$ & 465 & 45,240
\end{tabular}}
\end{table} 

\begin{figure}[t!]
    \centering
    \includegraphics[width=0.47\textwidth]{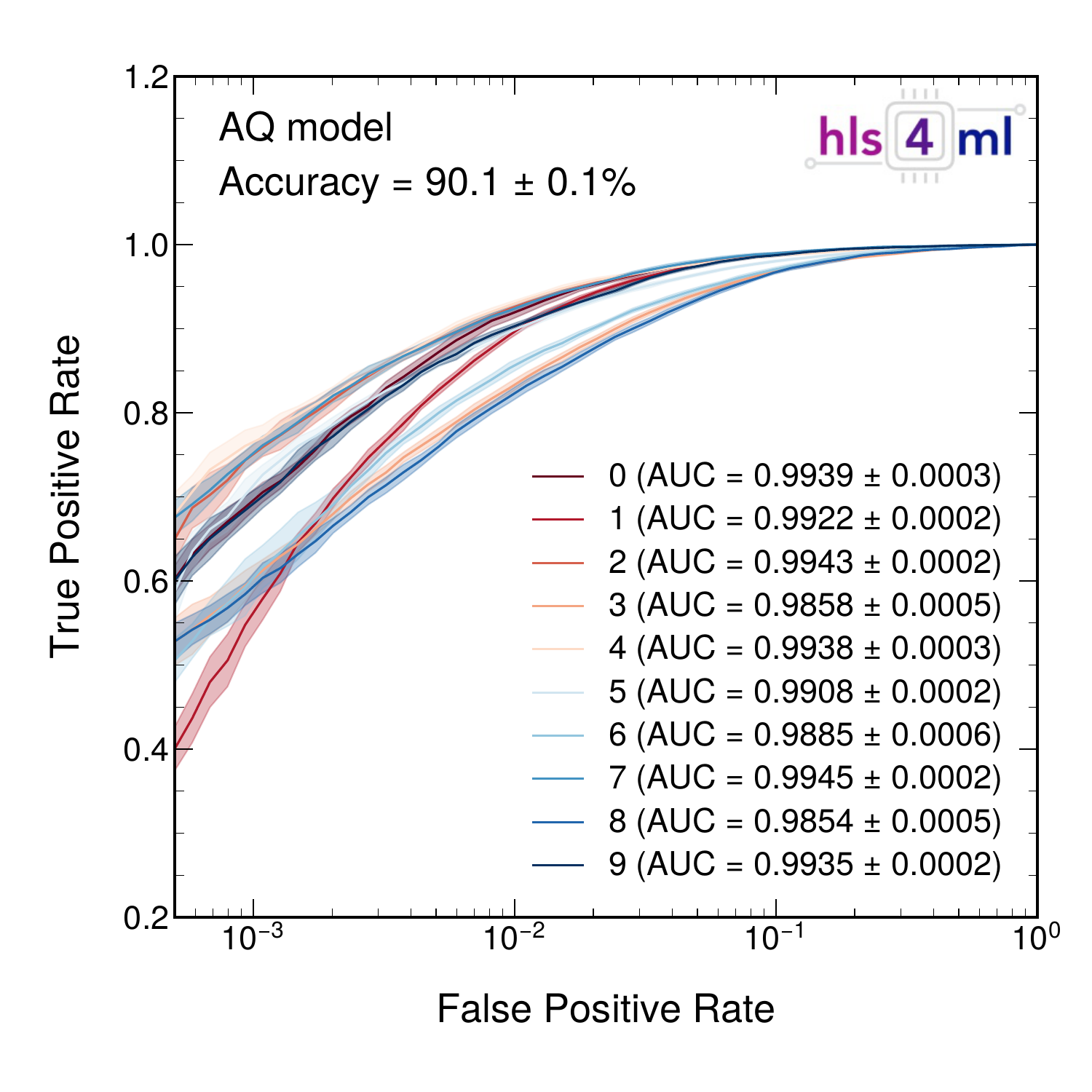}
    \includegraphics[width=0.47\textwidth]{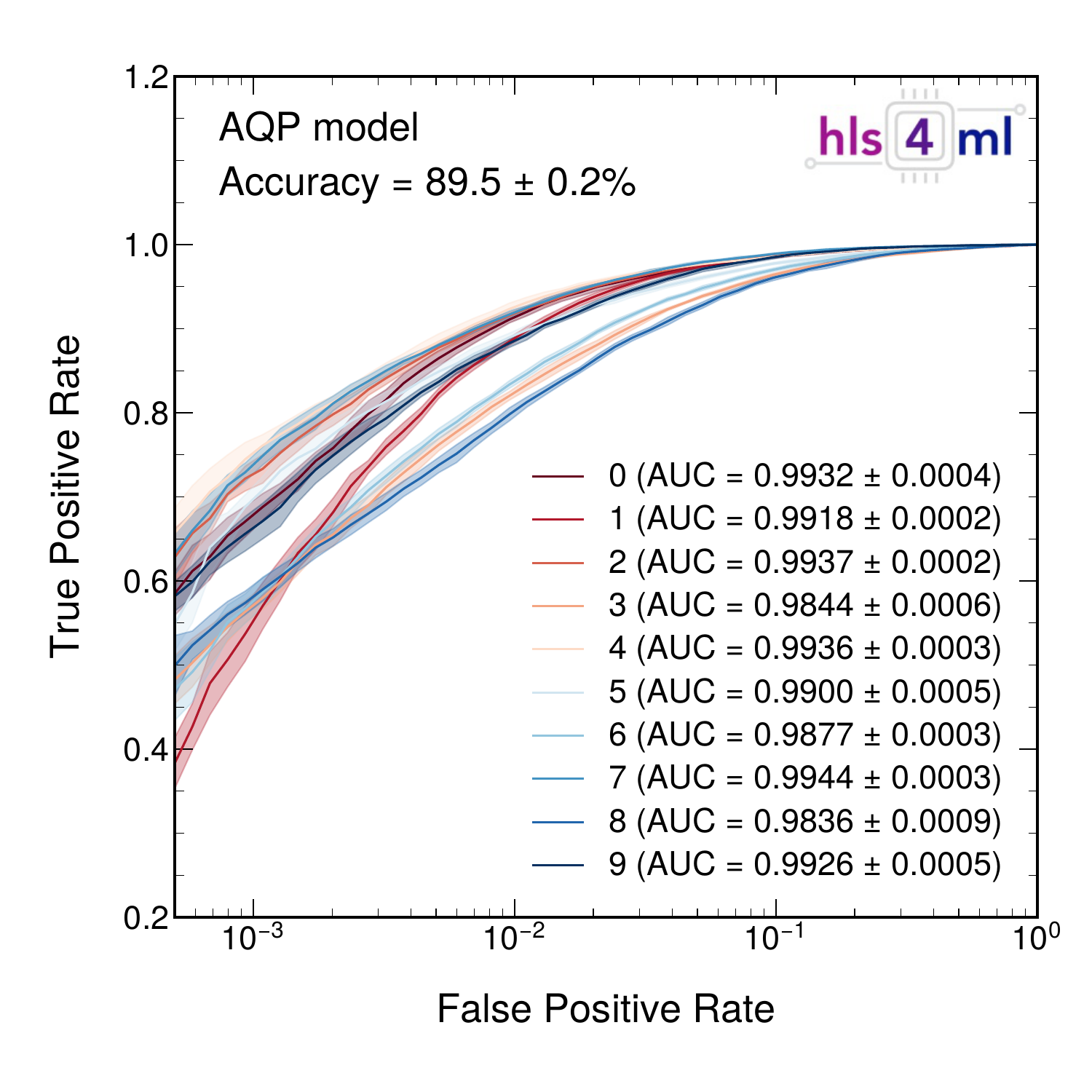}
    \caption{ROC curves of false positive rate (FPR) versus true positive rate (TPR) for the AutoQ (AQ) model (left) and the AutoQ Pruned (AQP) model (right). 
    Training is performed using $k$-fold cross validation, with $k=10$. 
    For each digit, the solid line corresponds to the mean across the ten folds and the band to the standard deviation.
    The mean area under the curve (AUC) across the ten folds is reported in the legend. 
    The mean accuracy and it's standard deviation across the ten folds is reported as well.\label{fig:ROC_quantized}}
\end{figure}

Figure~\ref{fig:ROC_quantized} shows the ROC curves for the AQ and AQP models. 
The curves show a slightly lower classification accuracy than those in Fig.~\ref{fig:roc_baseline}, with AUCs differing by approximately 1\%.

\begin{figure}[t!]
    \centering
    \includegraphics[width=0.47\textwidth]{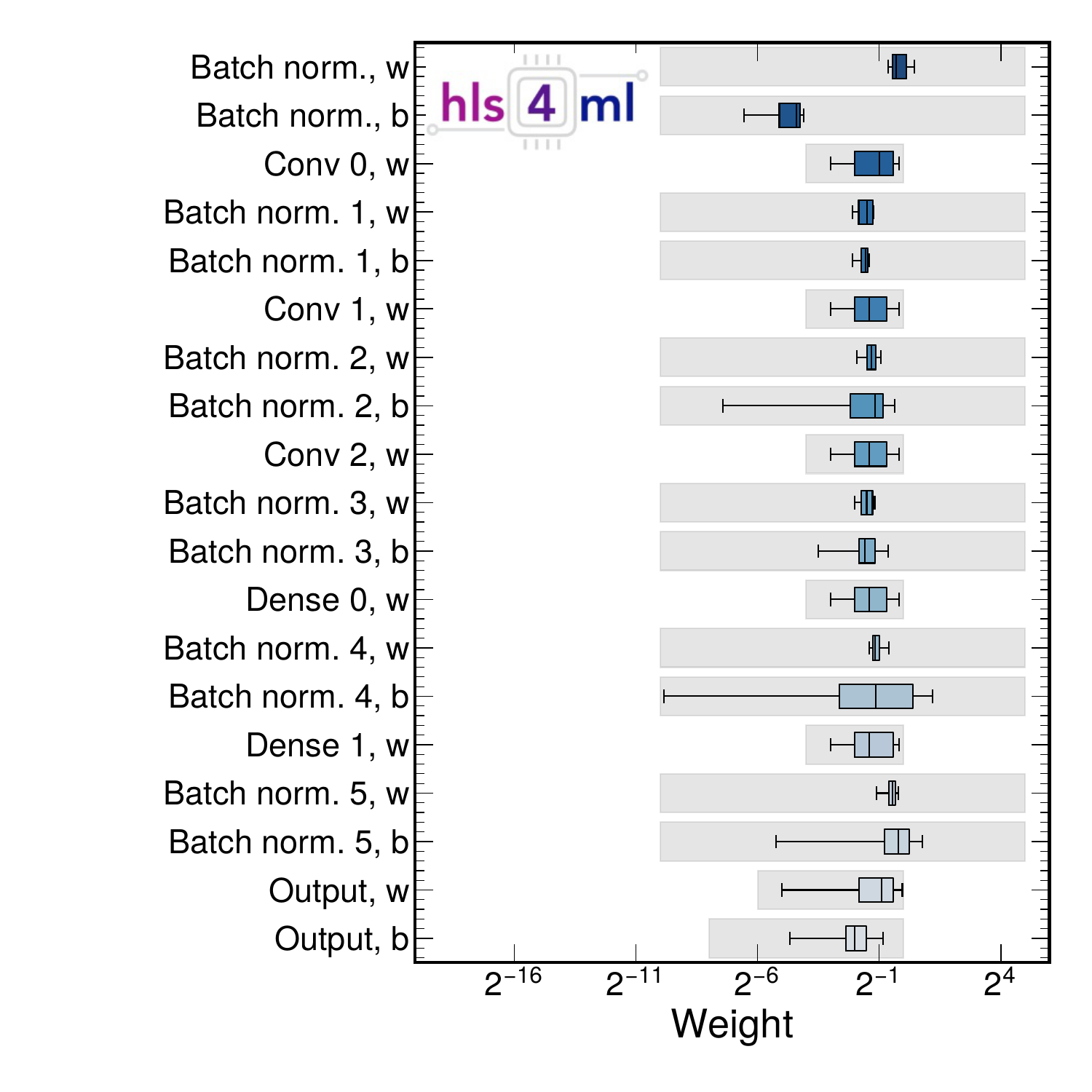}
    \includegraphics[width=0.47\textwidth]{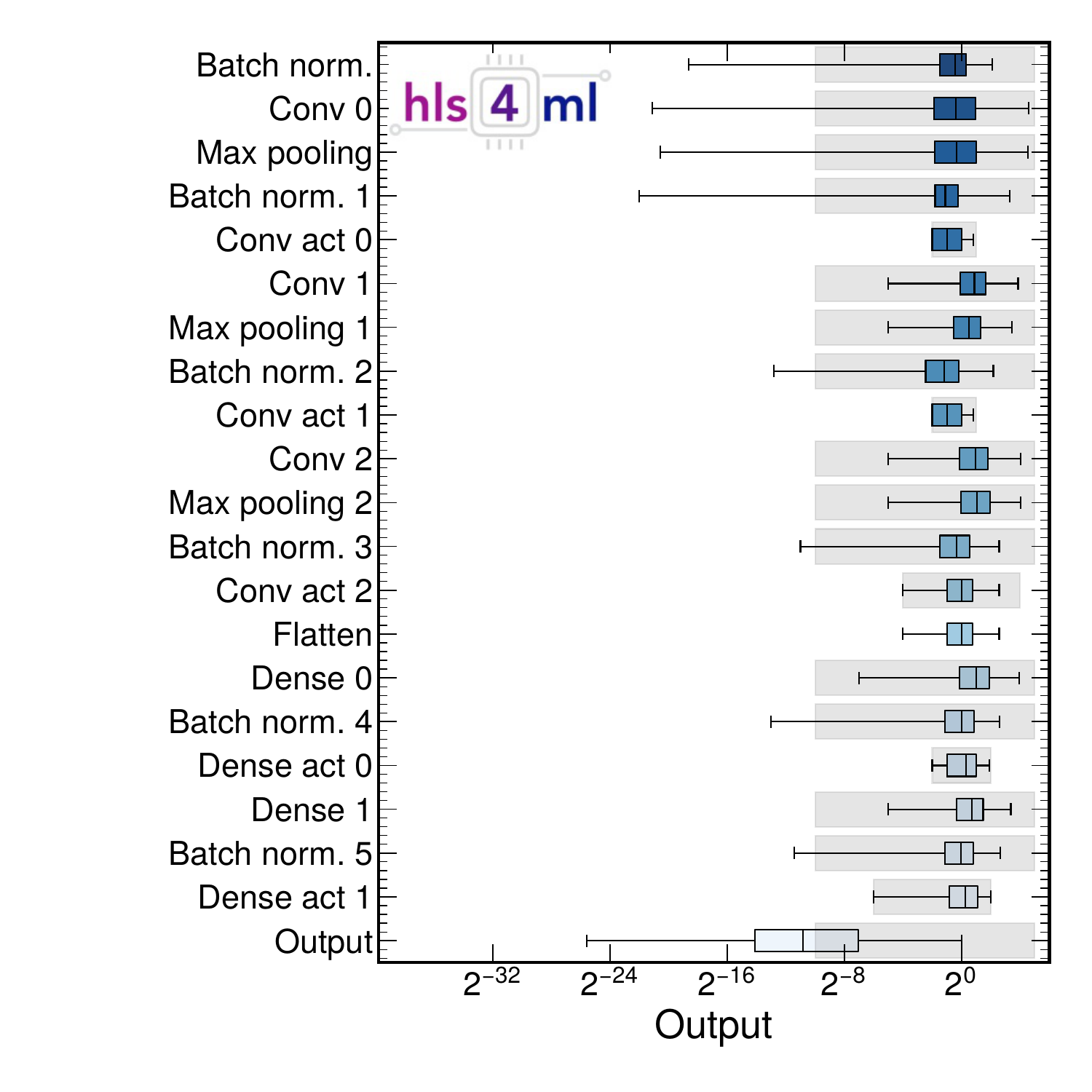}\\
    % \includegraphics[width=0.47\textwidth]{figures/Profile_pruned_full_0_weights.pdf}
    % \includegraphics[width=0.47\textwidth]{figures/Profile_pruned_full_0_activations.pdf}
        % The range covered by post-training quantization for a $\langle 16,6 \rangle$ precision is shown in gray.
    \caption{Layer weights (left) and output (right) numerical values per layer for the AutoQ (AQ) model using a subset of the training data. 
    The gray band represents the range covered by the fixed precision per layer in \hlsfml.
 \label{fig:autoq_weight_profile}}
\end{figure}

The numerical values spanned by the AQ model is shown in Figure~\ref{fig:autoq_weight_profile} for layer weights (left) and outputs (right). In contrast to those showed in Fig.~\ref{fig:baseline_weight_profile}, different bit widths are now used for the different layers, in correspondence with the bit width used in \QKeras. 

\begin{figure}[b!]
    \centering
    \includegraphics[width=0.47\textwidth]{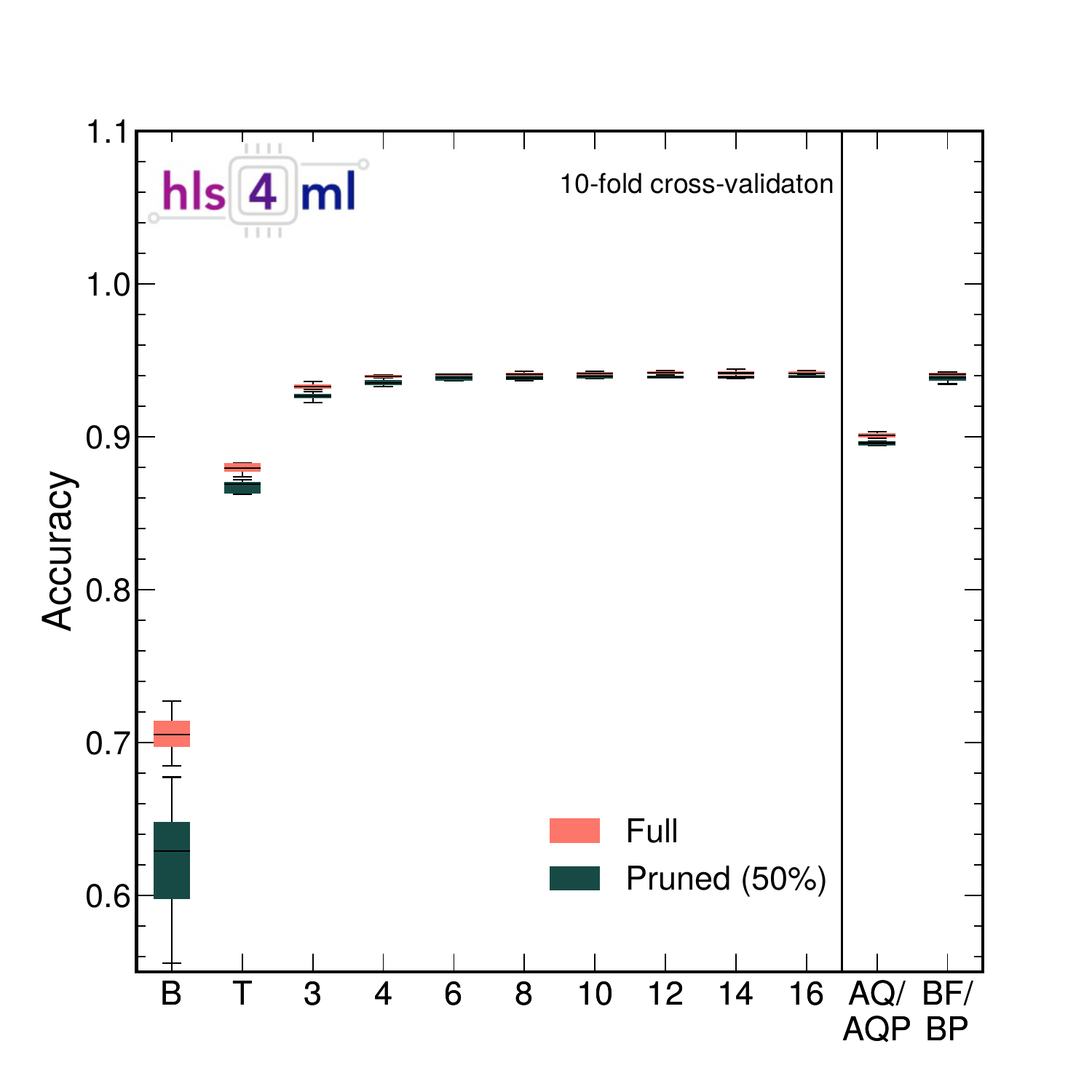}
    \caption{Accuracy for unpruned (red) and pruned (yellow) models for binary (B) and ternary (T) precisions, homogeneously quantized models with bit widths between 3 and 16, the heterogeneously quantized models AutoQ (AQ) and AutoQ Pruned (AQP), compared to the Baseline Floating-point (BF) and Baseline Pruned (BP) models. 
    The black line represents the median, and the box extends from the lower to upper quartile values across the 10 folds. \label{fig:accuracy_quantized}}
\end{figure}

Figure~\ref{fig:accuracy_quantized} summarizes the effects of pruning and quantization. 
Here, we show the median accuracy and upper and lower quartiles across the 10 folds of the unpruned (red) and pruned (green) quantized models, for different choices of bit widths and for the AQ (AQP) models. 
The unquantized baseline models are shown for reference (BF or BP). 
For bit widths above four, pruning to 50\% sparsity has very little impact on the model accuracy. At very low bit widths, however, pruning negatively impacts the model performance.
The accuracy is constant down to four bit precision, with marginal accuracy loss down to three bits. 
% Statistical uncertainty due to the choice of training set is also small: less than 1\%, for bit widths down to three. 
Using ternary quantization, the model accuracy drops to 87--88\% and has a higher statistical uncertainty. 
When quantizing down to binary precision, the model accuracy is reduced to 72\% for the unpruned model and 64\% for the pruned model. 
The significant reduction in accuracy due to pruning for binary networks is due to too little information being available in the network to accurately classify unseen data. 
A large spread in model accuracy for the binary network across the 10 folds is observed, indicating that the model is less robust to fluctuations in the training dataset. 
As demonstrated in~\cite{DiGuglielmo:2020eqx}, this can be mitigated by increasing the model size (more filters and neurons per layer). 
The AQ models obtain a slightly lower accuracy than the baselines, but uses, as will be demonstrated in Section~\ref{sec:fpgaporting}, significantly fewer resources. 

Due to the results above, it is recommended that users prune and quantize models using QAT through \QKeras, before proceeding with FPGA deployment with \hlsfml.

\section{FPGA porting}
\label{sec:fpgaporting}

\begin{figure}[hb!]
    \centering
    \includegraphics[width=0.47\textwidth]{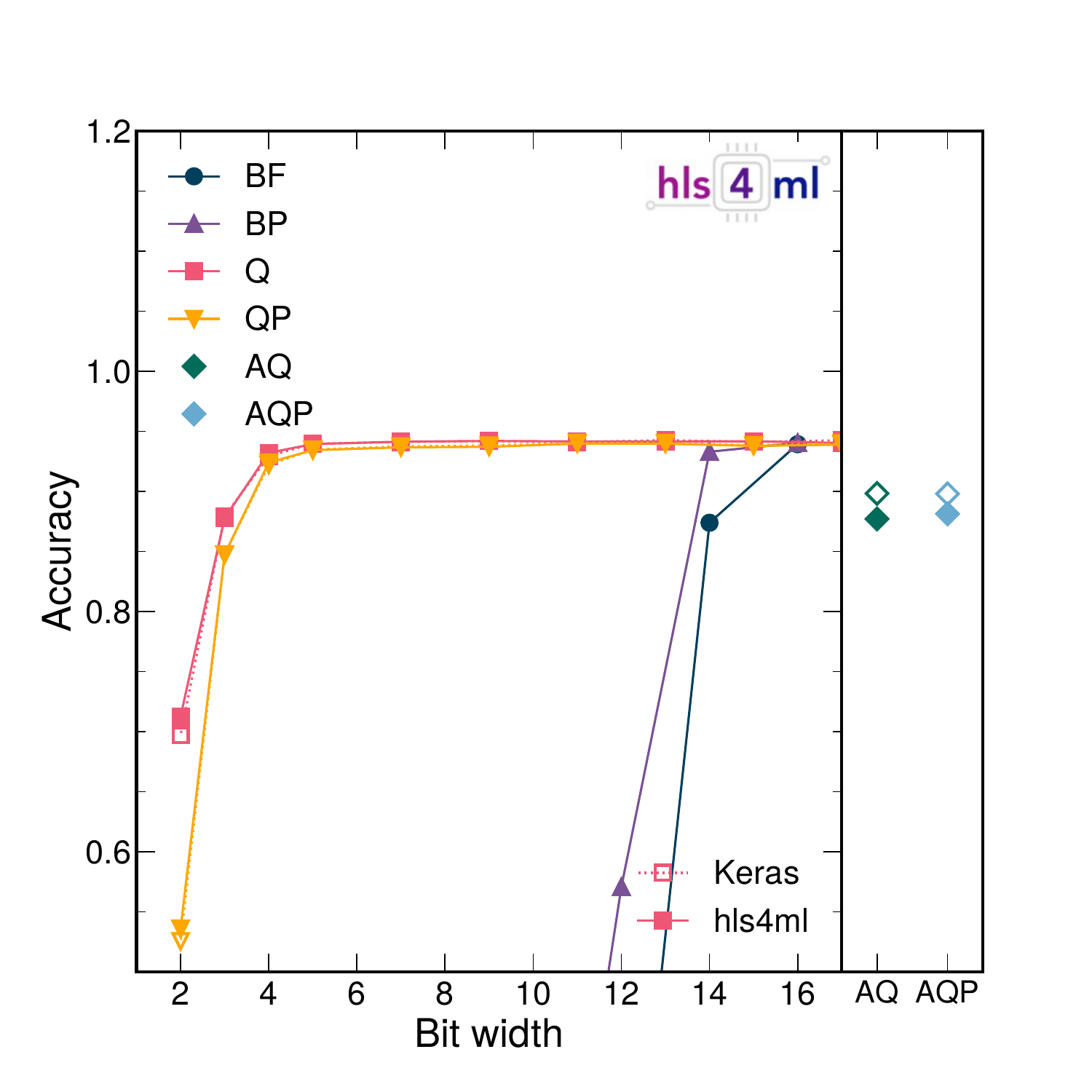}
    \caption{Model accuracy as a function of bit width for the Baseline Floating-point (BF), Baseline Pruned (BP), QKeras (Q) and QKeras Pruned (QP) models. 
    The heterogeneously quantized models AutoQ (AQ) and AutoQ Pruned (AQP) are shown in the sidebar.}
    \label{fig:accuracy_all}
\end{figure}

The models described above are translated into firmware using \hlsfml version 0.5.0, and then synthesized with Vivado HLS 2020.1, targeting a Xilinx Virtex UltraScale+ VU9P (\texttt{xcvu9pflgb2104-2L}) FPGA with a clock frequency of 200\unit{MHz}. 
For the \QKeras quantized models, the sign is not accounted for when setting the bit width per layer during QAT, so layers quantized with total bit width $b$ in \QKeras are therefore implemented as fixed-point numbers with total bit width $b+1$ in \hlsfml. 
We compare the model accuracy, latency, and on-chip resource consumption. 
The accuracy after translating the model into \texttt{C/C++} code with \hlsfml (solid line) for the different models, is shown in Figure~\ref{fig:accuracy_all} and compared to the accuracy evaluated using \Keras. 
No pre-synthesis results are shown for the BF and BP models, as these are quantized \emph{during} synthesis. 
Nearly perfect agreement in evaluated accuracy before and after synthesis is observed for the Q and QP models and the translation into fixed-point precision is lossless.

While the accuracy of the Q and QP models trained via QAT remains high down to a bit width of three, the accuracy of the PTQ models fall off sharply with decreasing bit width and have almost no discrimination power for bit widths smaller than 14. 
PTQ has a higher negative impact on the unpruned models, indicating that rounding errors are the biggest cause for accuracy degradation. 
The heterogeneously quantized models AQ and AQP have slightly lower accuracy than the baseline $\langle 16,6 \rangle$ model.

\begin{figure}[ht]
    \centering
    \includegraphics[width=0.47\textwidth]{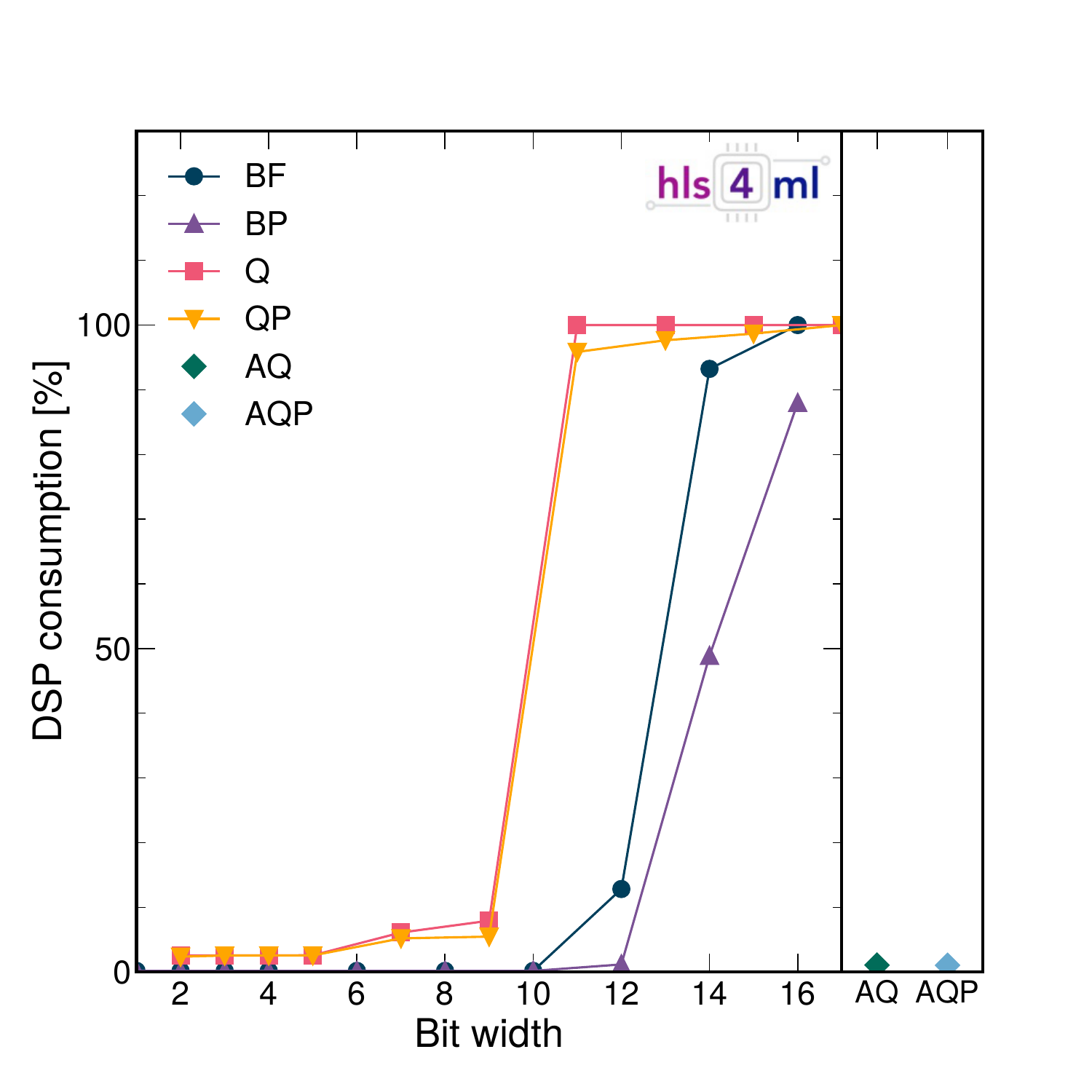}
    \includegraphics[width=0.47\textwidth]{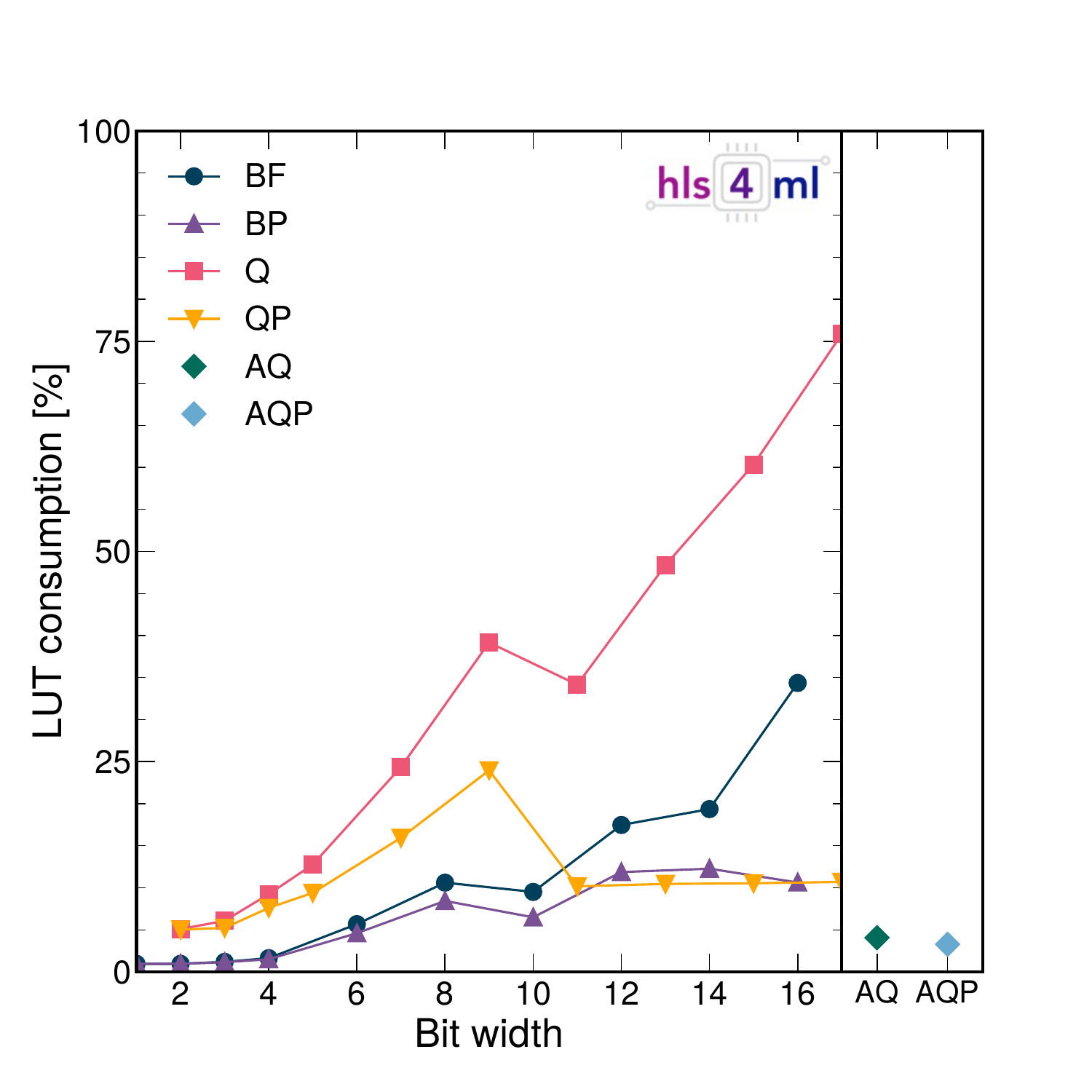}\\
    \includegraphics[width=0.47\textwidth]{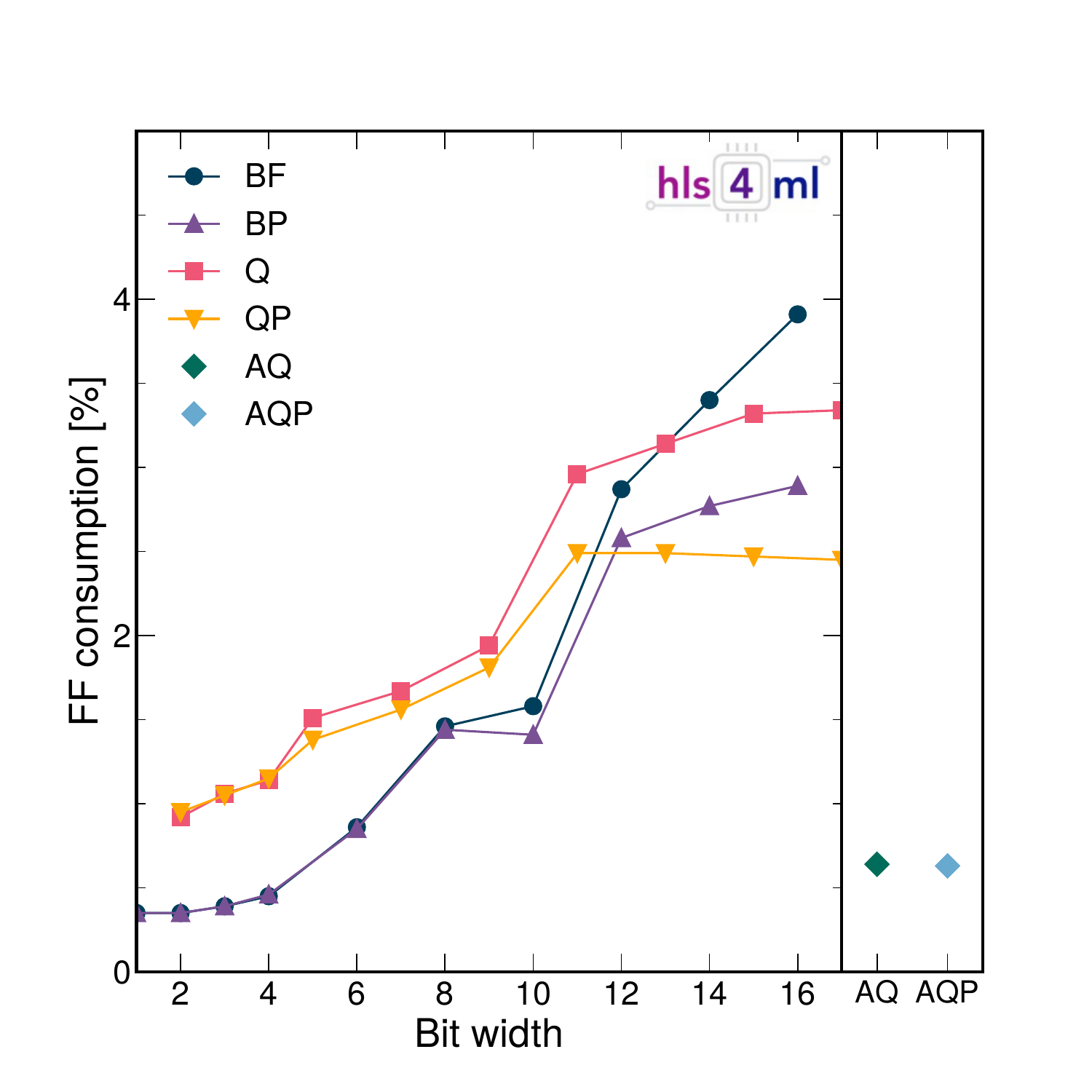}
    \includegraphics[width=0.47\textwidth]{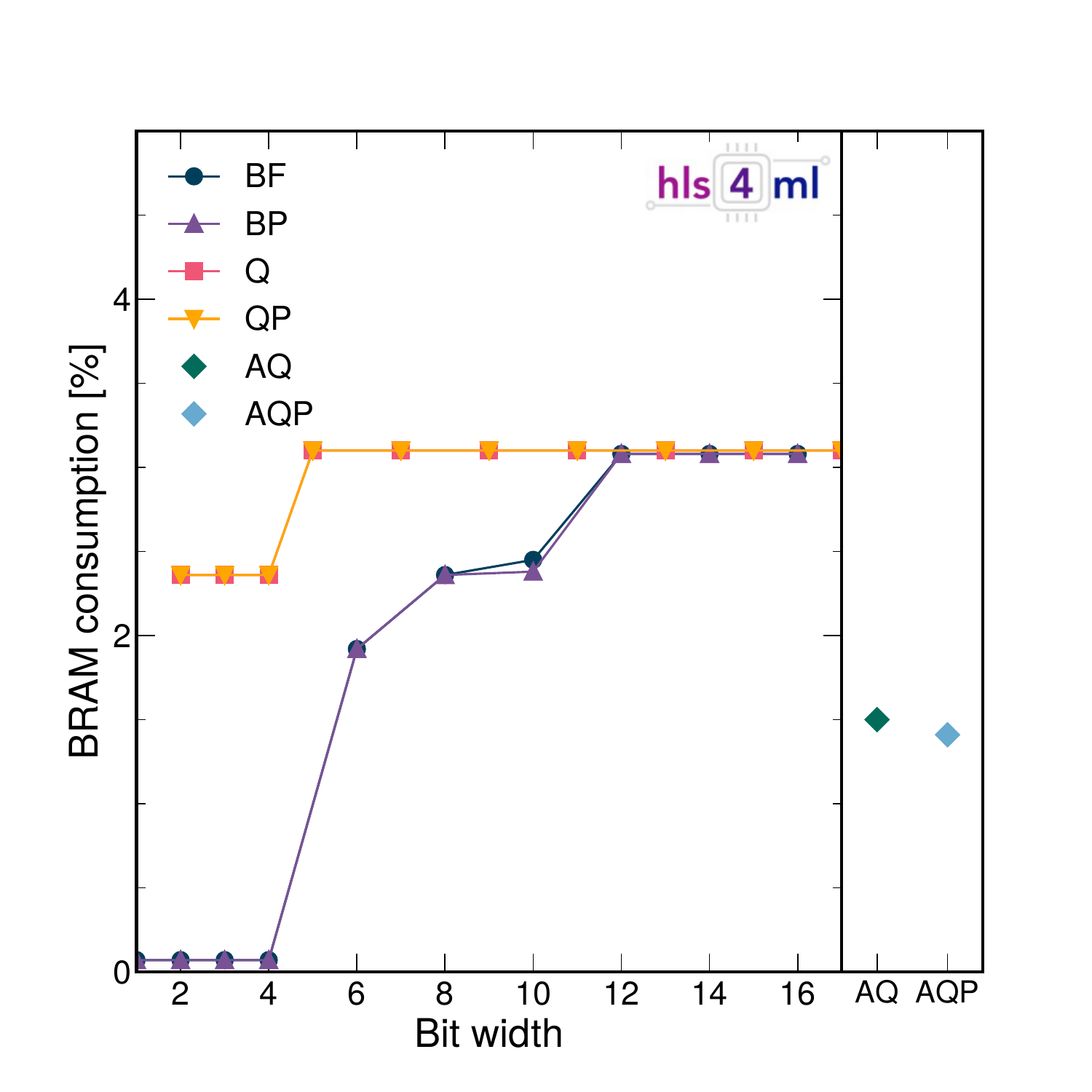}
        \caption{Resource consumption as a function of bit width for the Baseline Floating-point (BF), Baseline Pruned (BP), QKeras (Q), and QKeras Pruned (QP) models. The heterogeneously quantized AutoQ (AQ) and AutoQ Pruned (AQP) models are displayed in the right sub-plot. The model DSP (top left), LUT (top right), FF (bottom left) and BRAM (bottom right) consumption is shown. 
         \label{fig:hls4mlscan_resources}}
\end{figure}

We then study the resource consumption and latency of the different models after logic-synthesis. 
The resources available on the FPGA are digital signal processors (DSPs), lookup tables (LUTs), BRAMs, and flip-flops (FFs). 
In Fig.~\ref{fig:hls4mlscan_resources}, the resource consumption relative to the total available resources is shown.
Here, a fully parallel implementation is used where each multiplier is used exactly once, which can be achieved by setting the \emph{reuse factor} $R$~\cite{Duarte:2018ite} to 1 for each layer in \hlsfml.

The DSP consumption is slightly higher for the Q and QP models than the BF and BP models due to the batch normalization layers in the QAT models being fixed to $\langle 16, 6\rangle$.

Below a bit width of 10, the DSP consumption is significantly reduced as multiplications are performed using LUTs. 
DSPs are usually the limiting resource for FPGA inference, and we observe that through QAT, the DSP consumption can be reduced from one hundred percent down to a few percent with no loss in model accuracy (as demonstrated in Fig.~\ref{fig:accuracy_all}). 
Above a bit width of 10, almost all the DSPs on the device are in use for the Q and QP models. This routing is a choice of Vivado HLS during optimization of the circuit layout.
This is also the reason why pruning appears to have relatively little impact for these models: the DSPs are maximally used and the remaining multiplications are performed with LUTs. 
% This becomes evident when studying the LUT consumption as a function of bit width. 
The QP models use significantly fewer LUT resources than the unpruned equivalent. 
The point where most multiplications are moved from DSPs to LUTs is marked by a steep drop in DSP consumption starting at a bit width of 10. 

The heterogeneously quantized models, AQ and AQP, consume very little FPGA resources, comparable to that of the Q and QP models quantized to a bit width of three.  All models use very few FFs, below 4\% of the total budget. 
The BRAM consumption is also small and below 4\% for all models. 
% For the Q and QP models, the same amount of BRAMs is used down to a bit width of four, and then is further reduced. 
% For the BF and BP models, BRAM consumption falls off steadily with bit width. 
Some dependence on bit width can be traced back to how operations are mapped to the appropriate resources through internal optimizations in HLS. 
Depending on the length and the bit width of the FIFO buffers used for the convolutional layer sliding window, HLS will decide whether to place the operation on BRAMs or LUTs and migration between the two is expected. 
Most of the BRAMs, are spent on \emph{channels}, the output of different layers.

\begin{figure}[ht]
    \centering
    \includegraphics[width=0.47\textwidth]{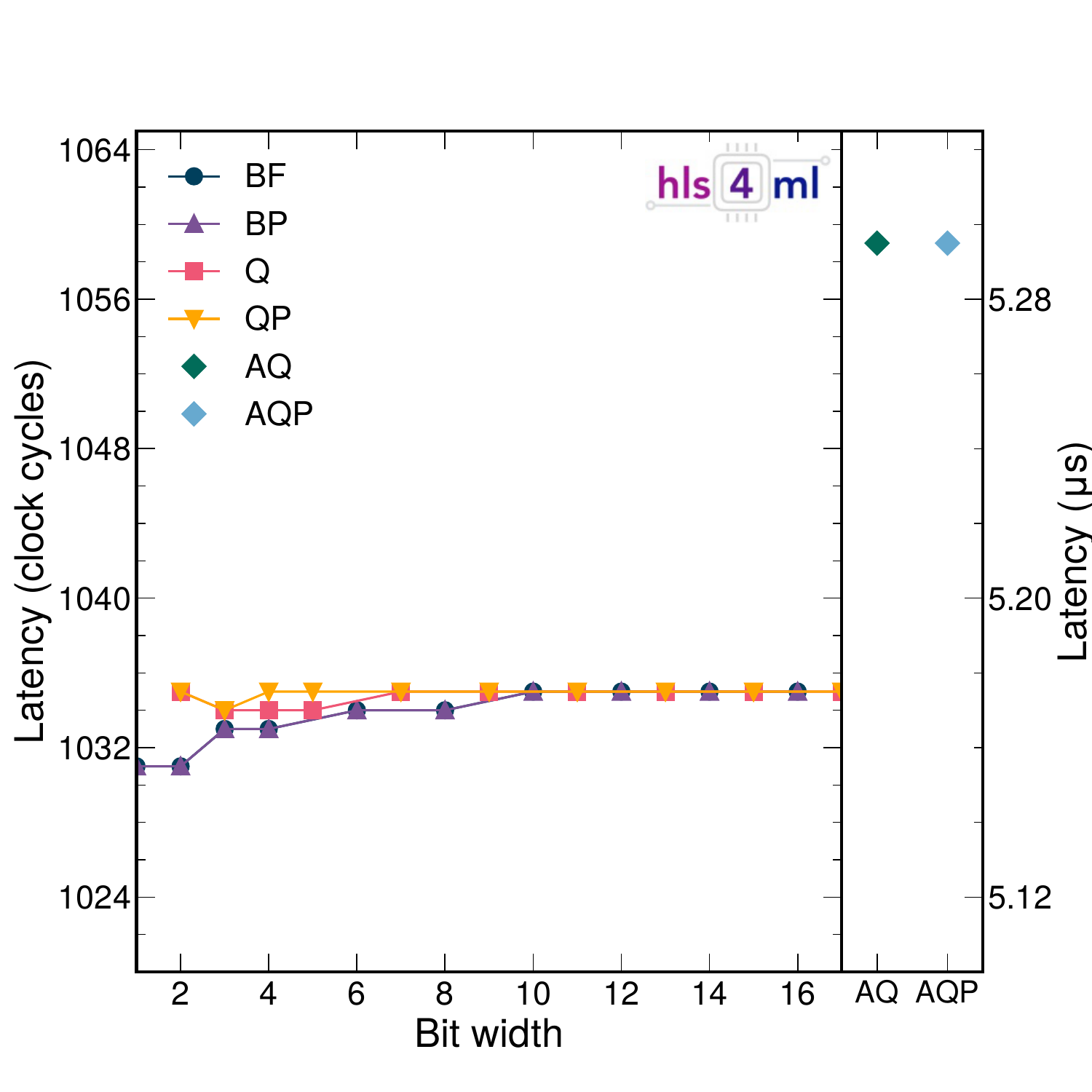}
    \includegraphics[width=0.47\textwidth]{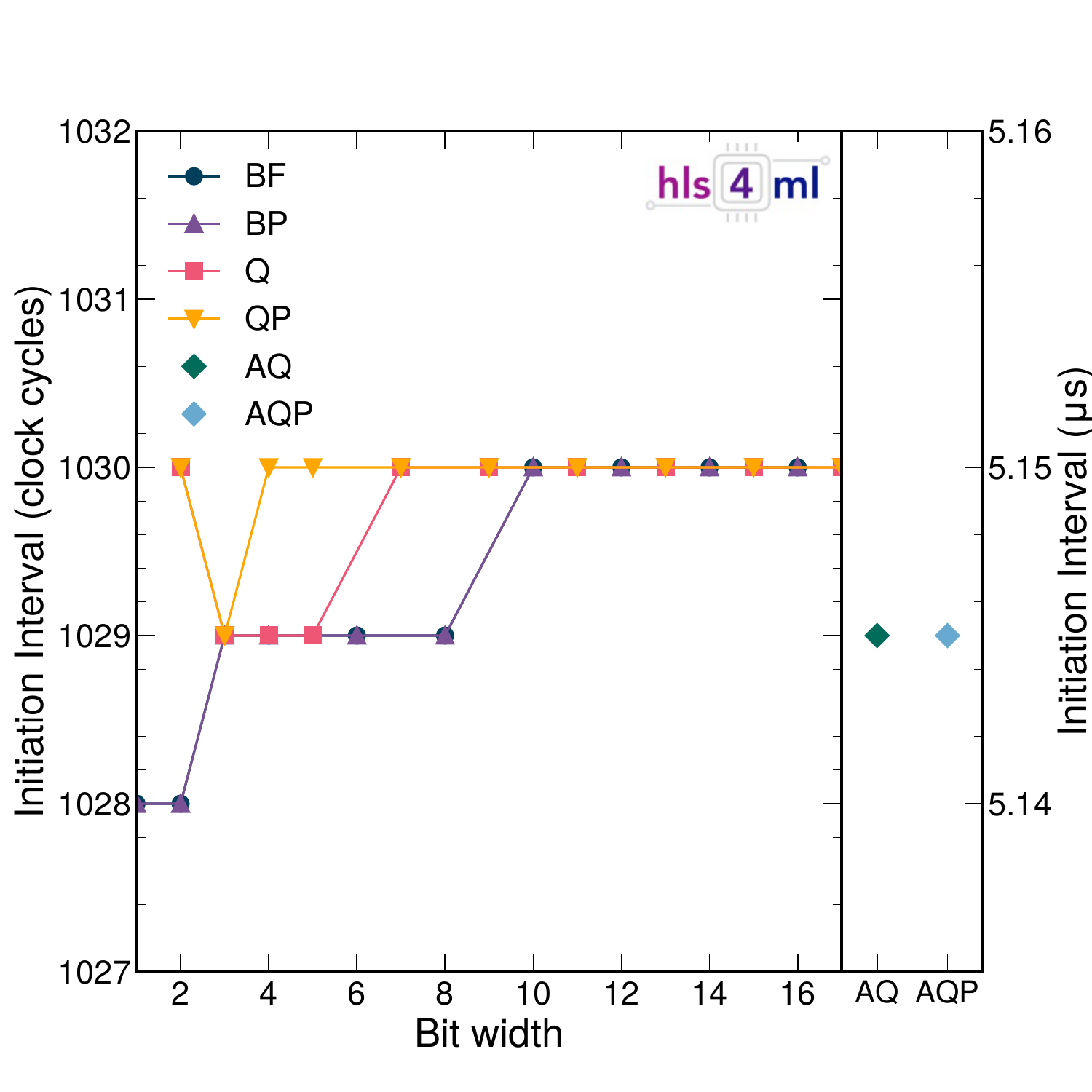}
        \caption{The model latency (left) and initiation interval (right) as a function of bit width for the Baseline Floating-point (BF), Baseline Pruned (BP), QKeras (Q), and QKeras Pruned (QP) models. The heterogeneously quantized AutoQ (AQ) and AutoQ Pruned (AQP) models are displayed in the right sub-plot. \label{fig:hls4mlscan_latency}}
\end{figure}

The latency and II for all models is shown in Figure~\ref{fig:hls4mlscan_latency}. 
A total latency of about 5\unit{$\mu$s} is observed for all models, similar to the II. 
The latency is independent of bit width when running at a fixed clock period. 
We leave it for future studies to explore running the board at higher clock frequencies.
% However, narrower bit widths do allow the FPGA to operate at a higher frequency. 
% {\bf add plot/discussion here}.

A summary of the accuracy, resource consumption and latency for the Baseline Floating-point (BF) and Baseline Pruned (BP) models quantized to a bit width of 14, the QKeras (Q) and QKeras Pruned (QP) models quantized to a bit width of 7 and the heterogeneously quantized AutoQ (AQ) and AutoQ Pruned (AQP) models, is shown in Table~\ref{tab:fpgasummary}. Resource utilization is quoted as a fraction of the total available resources on the FPGA, and the absolute number of resources used is quoted in parenthesis. The accuracy of the post-training quantized BF and BP models drops below 50\% for bit widths narrower than 14 and can not be used for inference. The QAT models, Q and QP, quantized to a bit width of 7 maintain a high accuracy despite using only a fraction of the available FPGA resources. The models using the fewest resources are the AQ and AQP heterogeneously quantized models, reducing the DSP consumption by 99\% while maintaining a relatively high accuracy. Finding the best trade-off between model size and accuracy in an application-specific way can be done using \AutoQKeras, as demonstrated in Sec.~\ref{sec:quantization}.  

\begin{figure}[h!]
    \centering
    \includegraphics[width=0.41\textwidth]{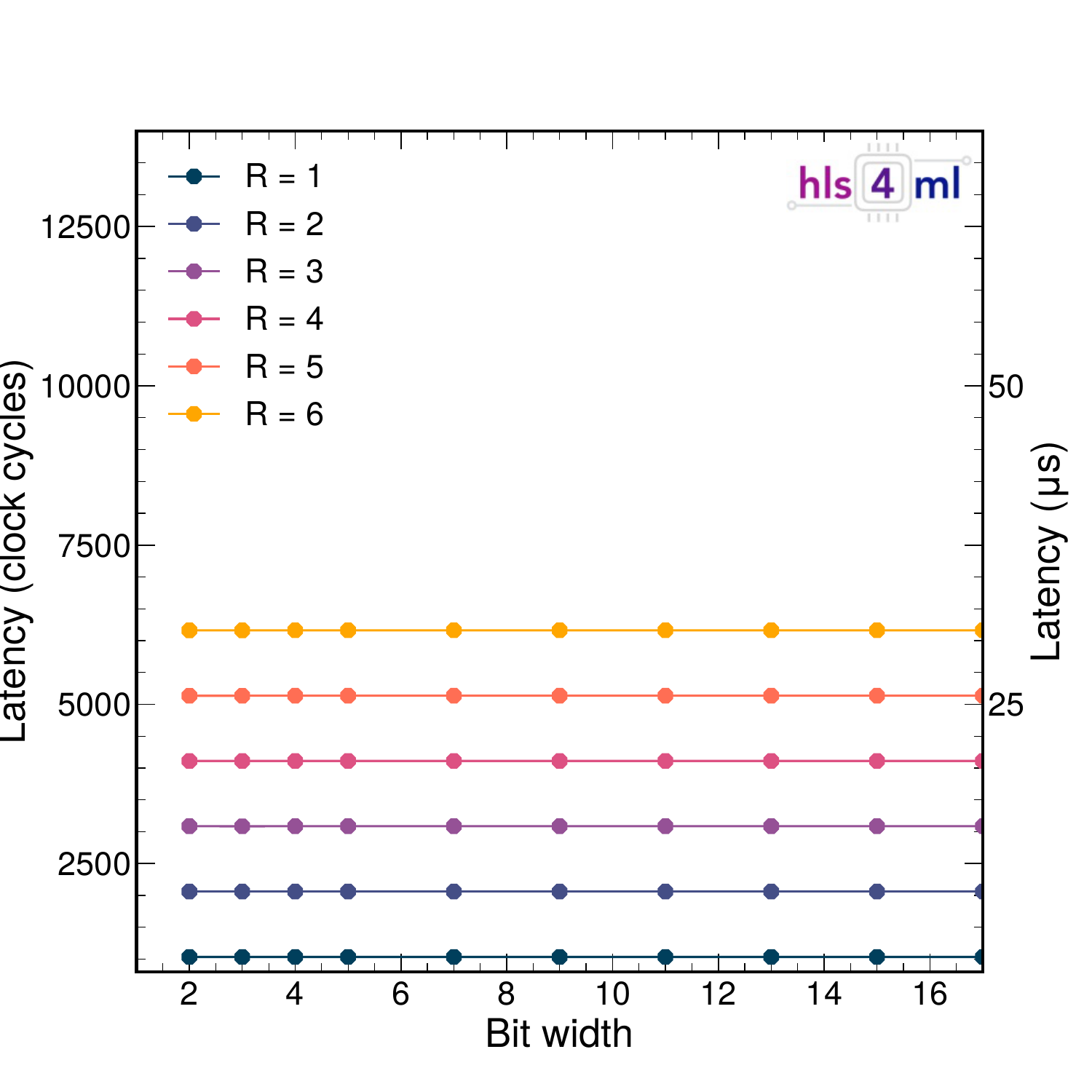}
    \includegraphics[width=0.41\textwidth]{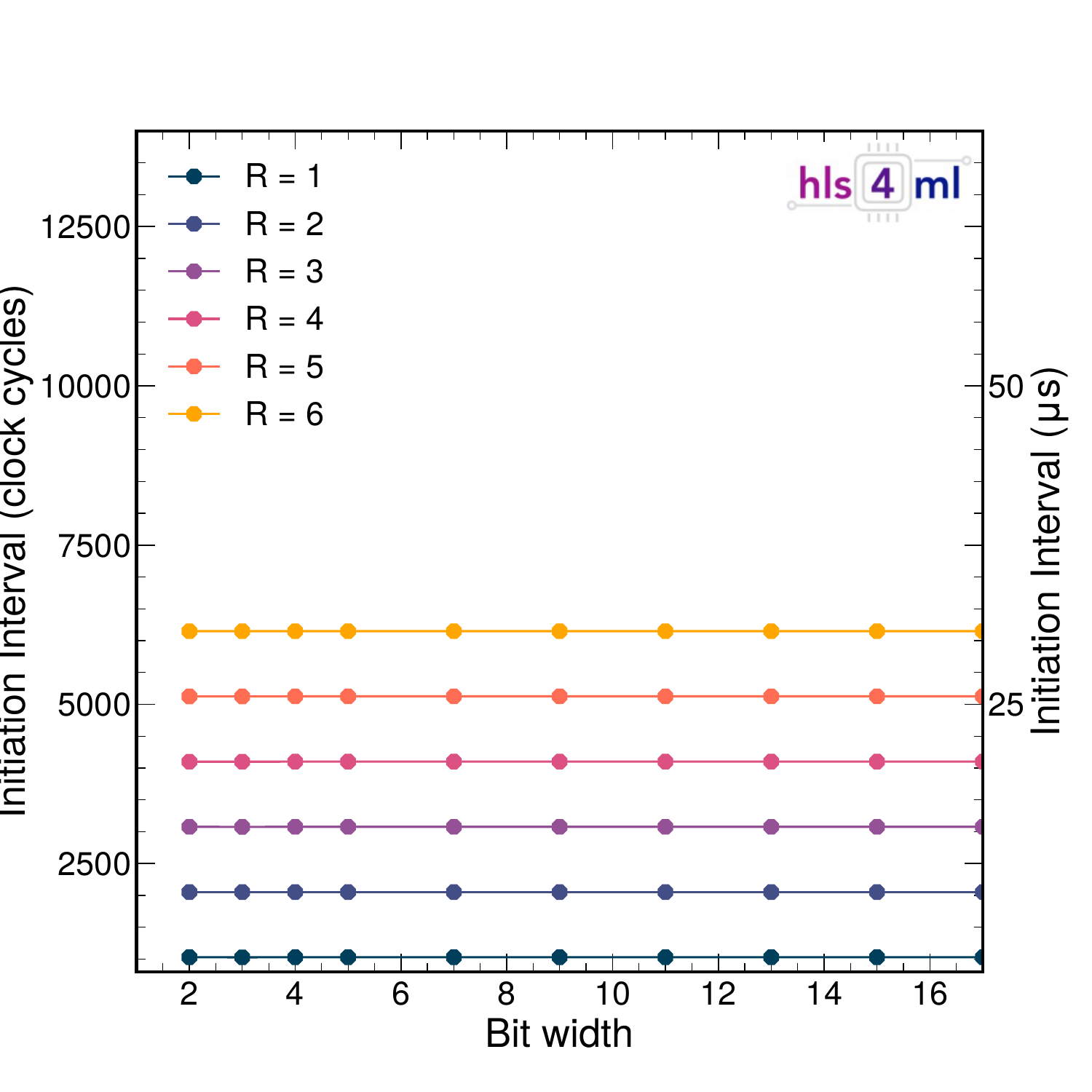}\\
    \includegraphics[width=0.41\textwidth]{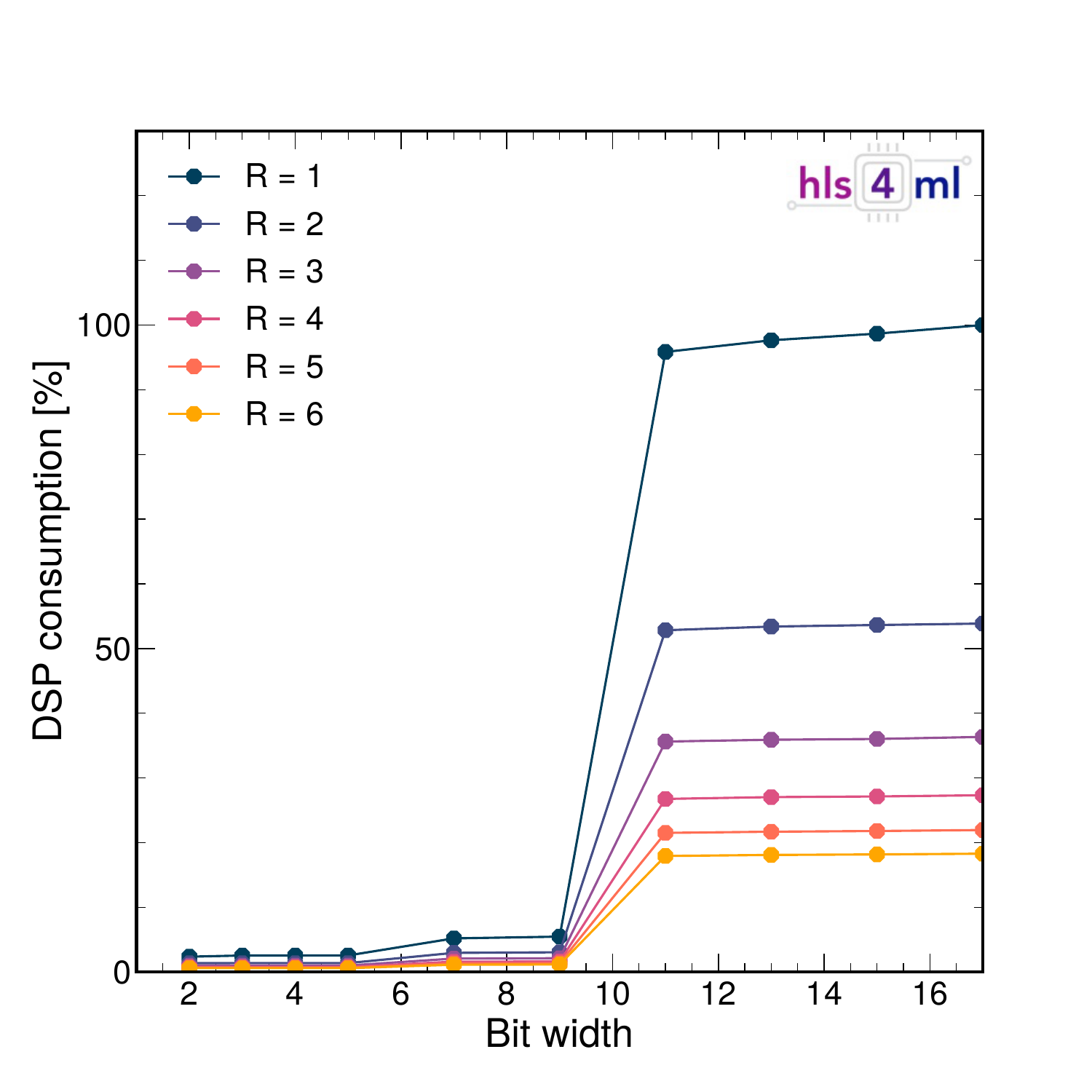}
    \includegraphics[width=0.41\textwidth]{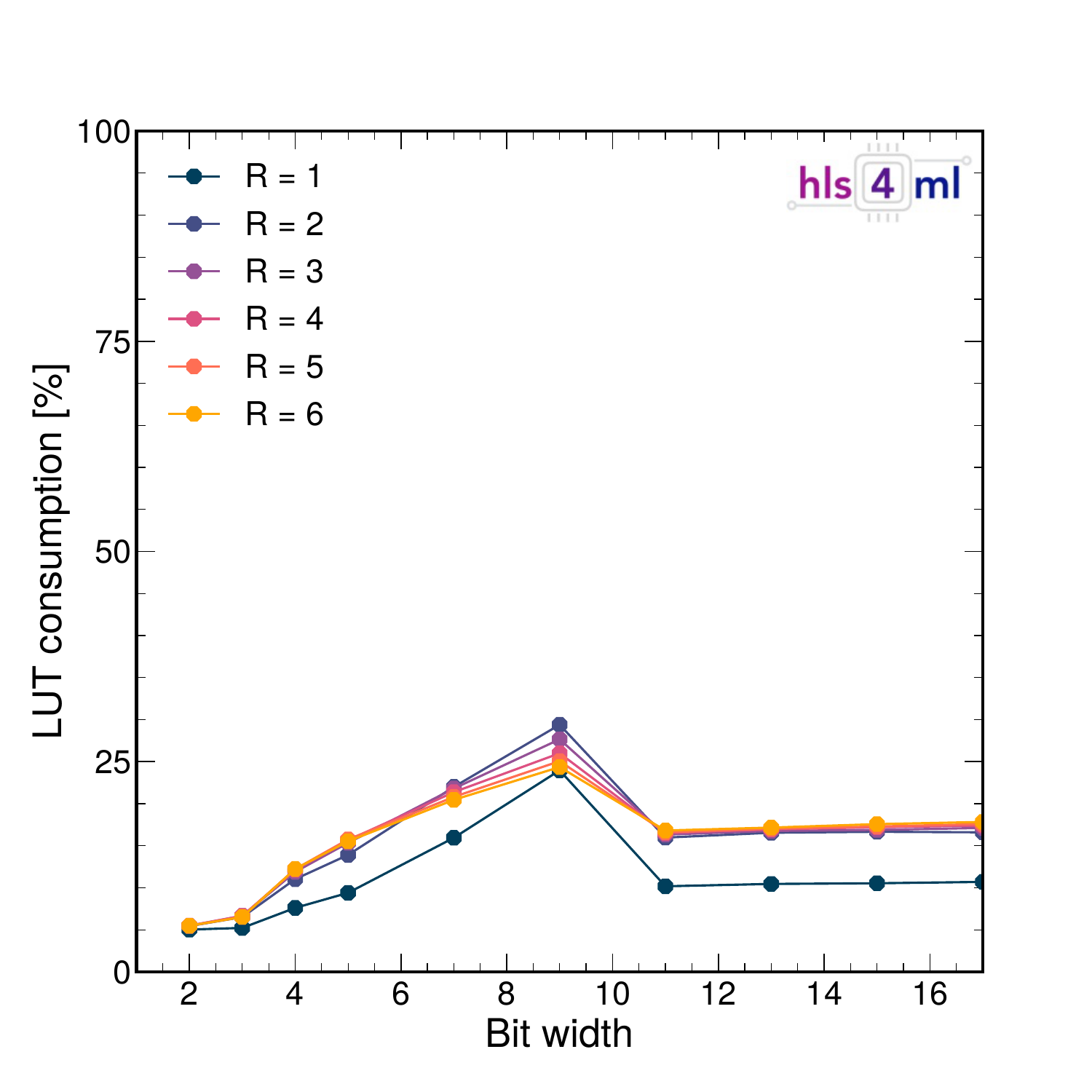}\\   
    \includegraphics[width=0.41\textwidth]{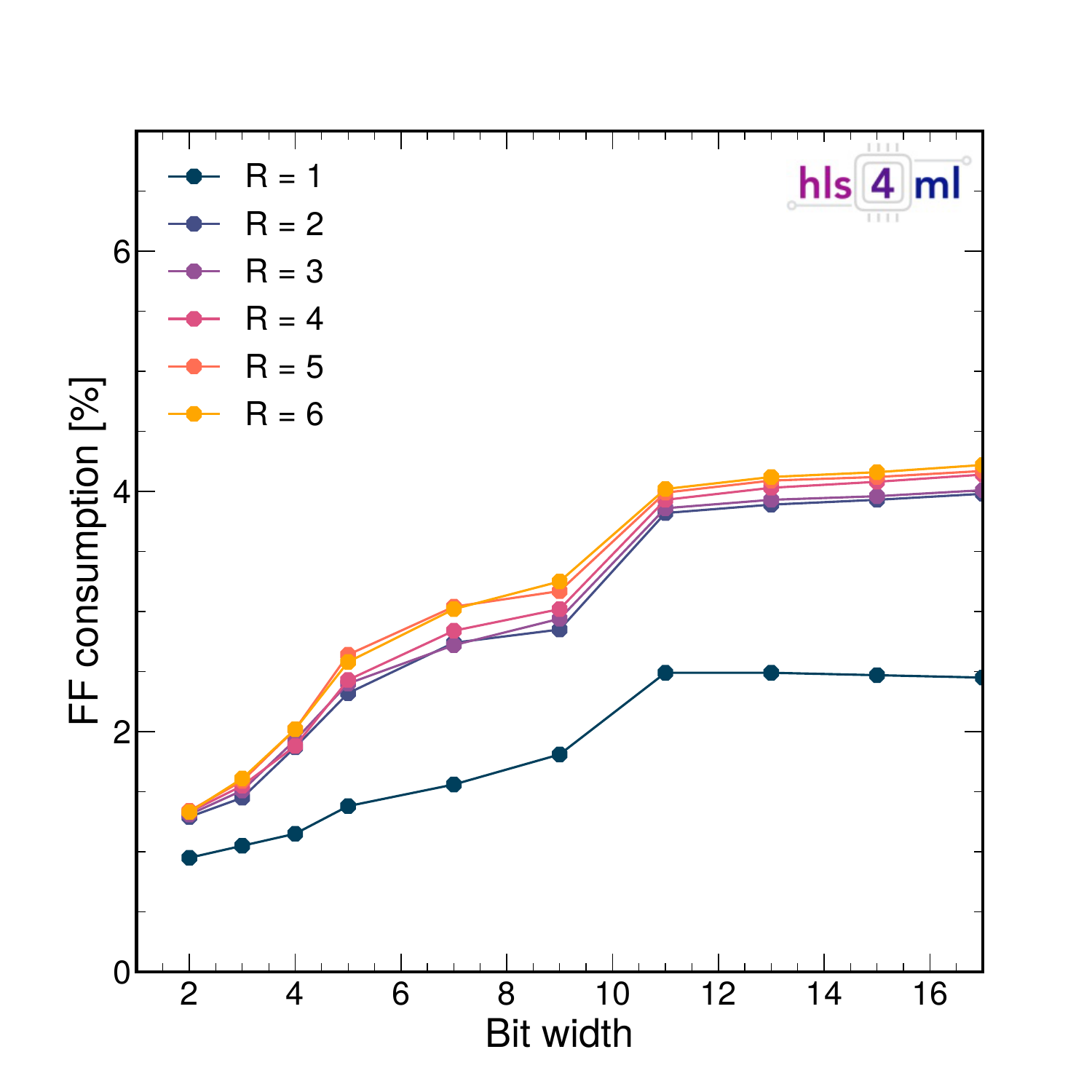}
    \caption{Latency (top left), initiation interval (top right), DSP (middle left), LUT (middle right), FF (bottom) consumption as a function of bit width and for different reuse factors for the QKeras Pruned (QP) models. \label{fig:pruned_quant_per_reuse}}
\end{figure}

\begin{table}[ht!]
 \caption{Accuracy, resource consumption and latency for the Baseline Floating-point (BF) and Baseline Pruned (BP) models quantized to a bit width of 14, the QKeras (Q) and QKeras Pruned (QP) models quantized to a bit width of 7 and the heterogeneously quantized AutoQ (AQ) and AutoQ Pruned (AQP) models. The numbers in parentheses correspond to the total amount of resources used. \label{tab:fpgasummary}}
  \centering 
\resizebox{\textwidth}{!}{
\begin{tabular}{l|ccccccc}
\multicolumn{7}{l}{FPGA: Xilinx Virtex UltraScale+ VU9P}\\
\hline
Model     & Accuracy &   DSP       & LUT        &FF       & BRAM     & Latency [cc]  & II [cc]\\
\hline
BF 14-bit &  0.87   & 6,377 (93.2\%)   & 228,823 (19.4\%) & 80,278 (3.4\%)  & 66.5 (3\%)   & 1,035 (5.2\unit{$\mu$s})          & 1,030   \\
BP 14-bit &  0.93   & 3,341 (48.9\%)   & 145,089 (12.3\%) & 65,482 (2.8\%)  & 66.5 (3\%)   & 1,035 (5.2\unit{$\mu$s})          & 1,030   \\
Q  7-bit  &  0.94   & 175 (2.5\%)      & 150,981 (12.8\%) & 35,628 (1.5\%)  & 67.0 (3\%)   & 1,034 (5.2\unit{$\mu$s})          & 1,029   \\ 
QP 7-bit  &  0.94   & 174 (2.5\%)      & 111,152 (9.4\%)  & 32,554 (1.4\%)  & 67.0 (3\%)   & 1,035 (5.2\unit{$\mu$s})          & 1,030   \\
AQ        &  0.88   & 72 (1.0\%)       & 48,027 (4.0\%)   & 15,242 (0.6\%)  & 32.5 (2\%)   & 1,059 (5.3\unit{$\mu$s})          & 1,029   \\
AQP       &  0.88   & 70 (1.0\%)       & 38,795 (3.3\%)   & 14,802 (0.6\%)  & 30.5 (1\%)   & 1,059 (5.3\unit{$\mu$s})          & 1,029   \\
% \toprule
% {\bf Face Detector NullHop ~\cite{nullhop}} &  -   & - & - & - & - &26.4\unit{$\mu$s}& - \\
% {\bf Face Detector \hlsfml }&  -       &   56.8  (3,885)  &0.95 (22,576)  & 5.67 (67,054) & 2.08 (2,160)    & 1,316 (6.6\unit{$\mu$s})          & 1,302   \\
\end{tabular}}
\end{table} 
To further reduce the resource consumption, the reuse factor $R$ can be increased. 
This comes at the cost of higher latency. % and the user must eventually make a trade-off between the two depending on the specific application.
The model latency and resource consumption as a function of bit width and for different reuse factors for the QP models are shown in Figure~\ref{fig:pruned_quant_per_reuse}. 
The latency and II increase with $R$, while the DSP consumption goes down. The LUT consumption is minimally affected by the reuse factor, consistent with the results reported in~\cite{Duarte:2018ite}. 
The BRAM consumption is the same for all reuse factors, around 3\%, and therefore not plotted. The corresponding study for the BF, BP and Q models can be found in Appendix~\ref{app:reusescan}.

\begin{figure}[h!]
    \centering
    \includegraphics[width=0.47\textwidth]{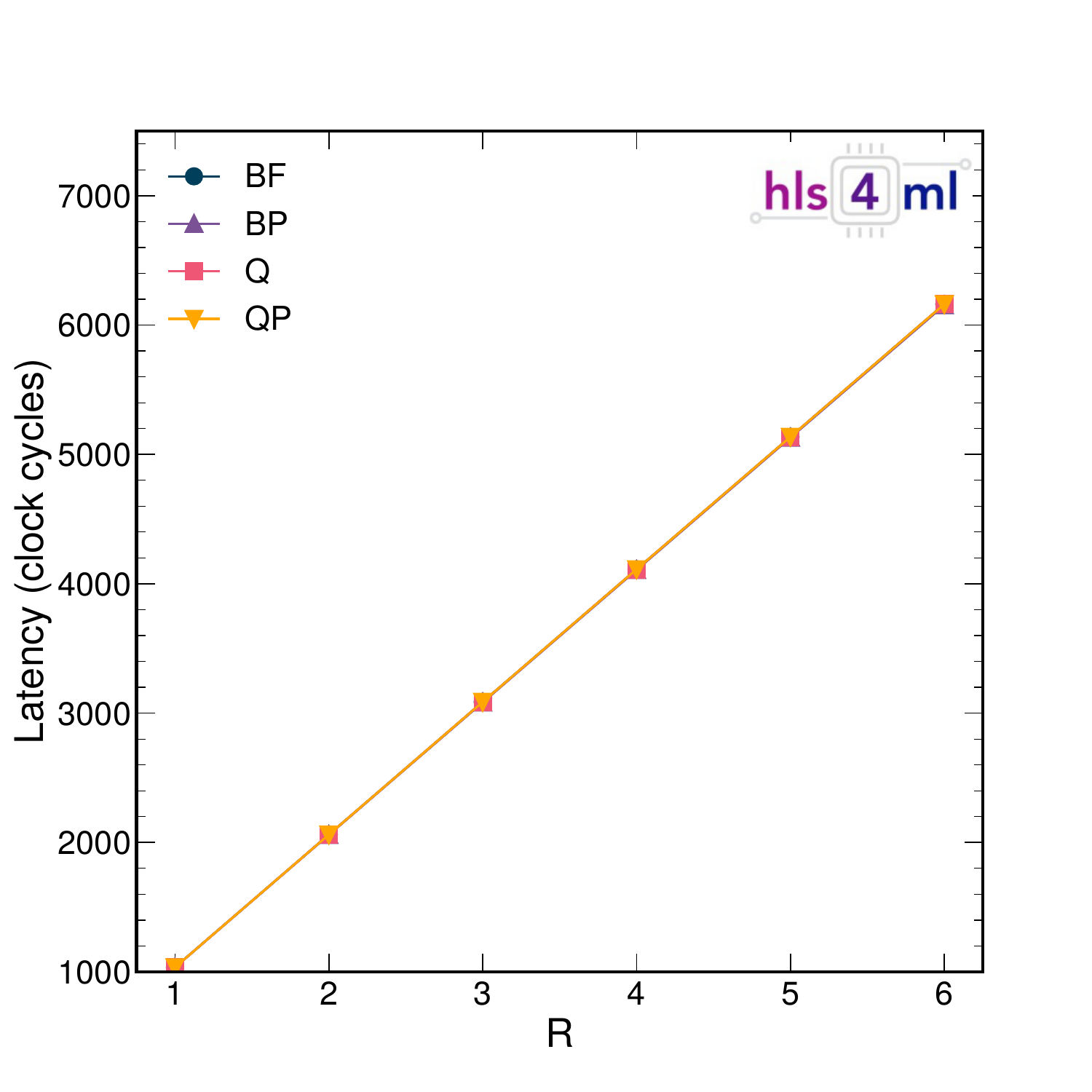}
    \includegraphics[width=0.47\textwidth]{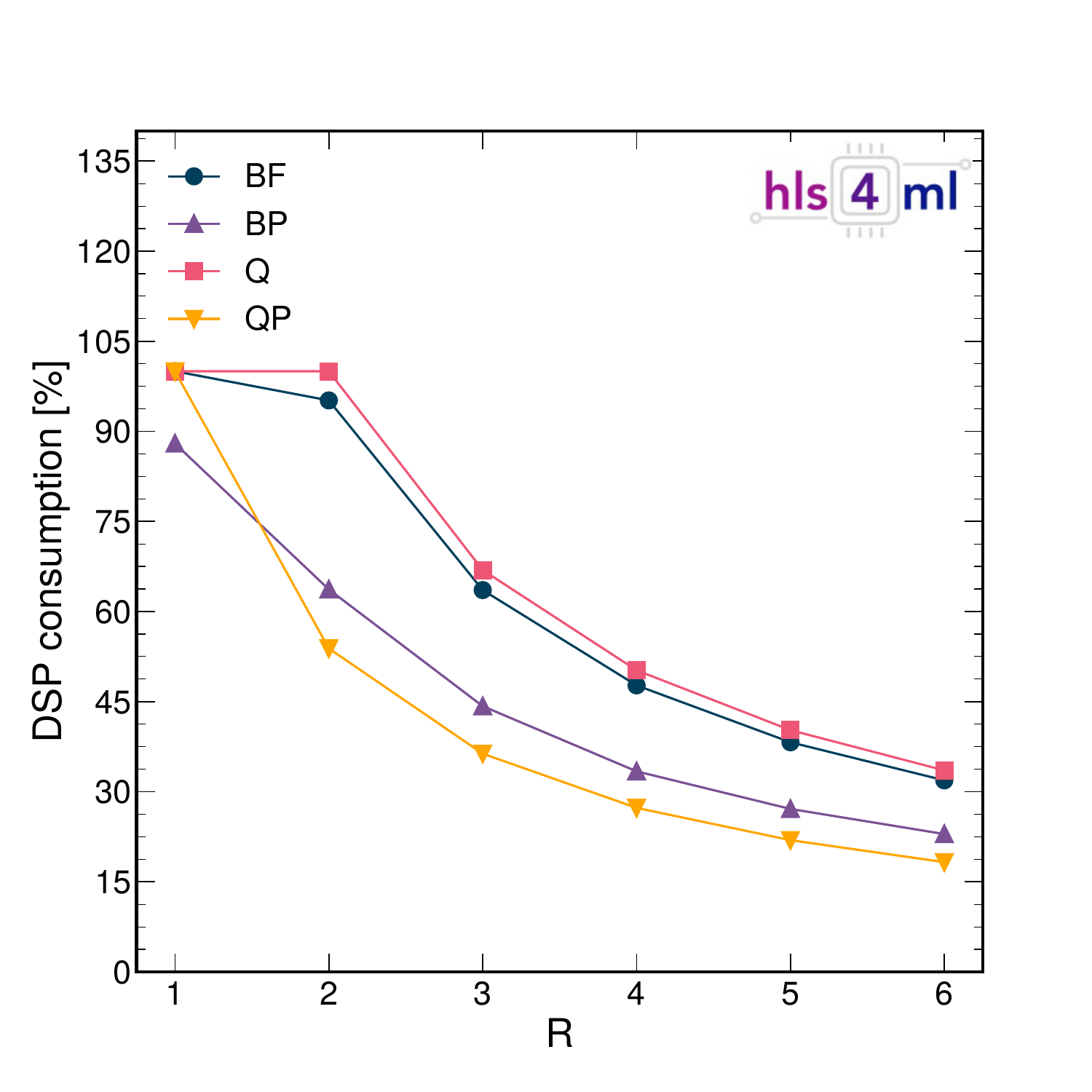}\\
    \includegraphics[width=0.47\textwidth]{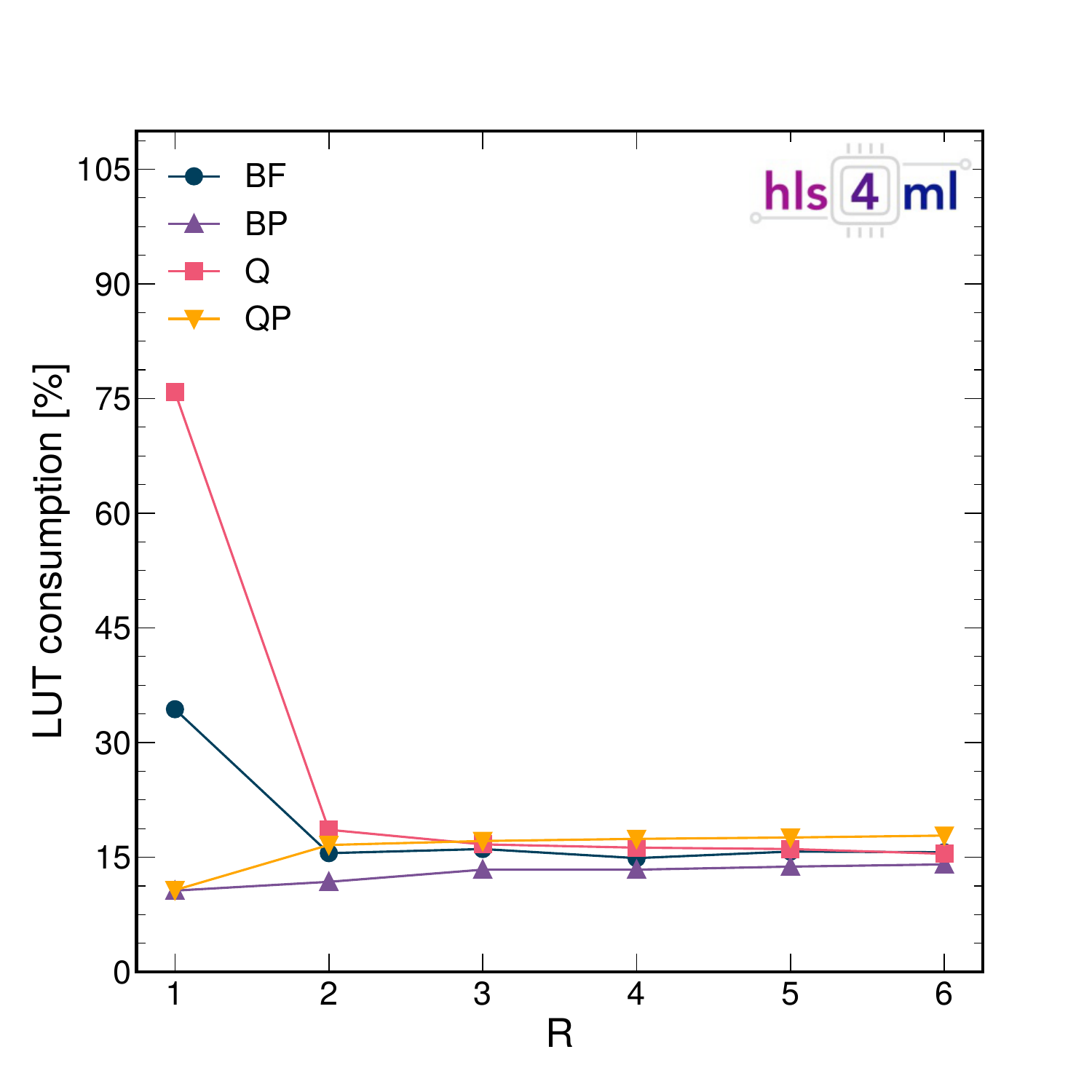}
    \includegraphics[width=0.47\textwidth]{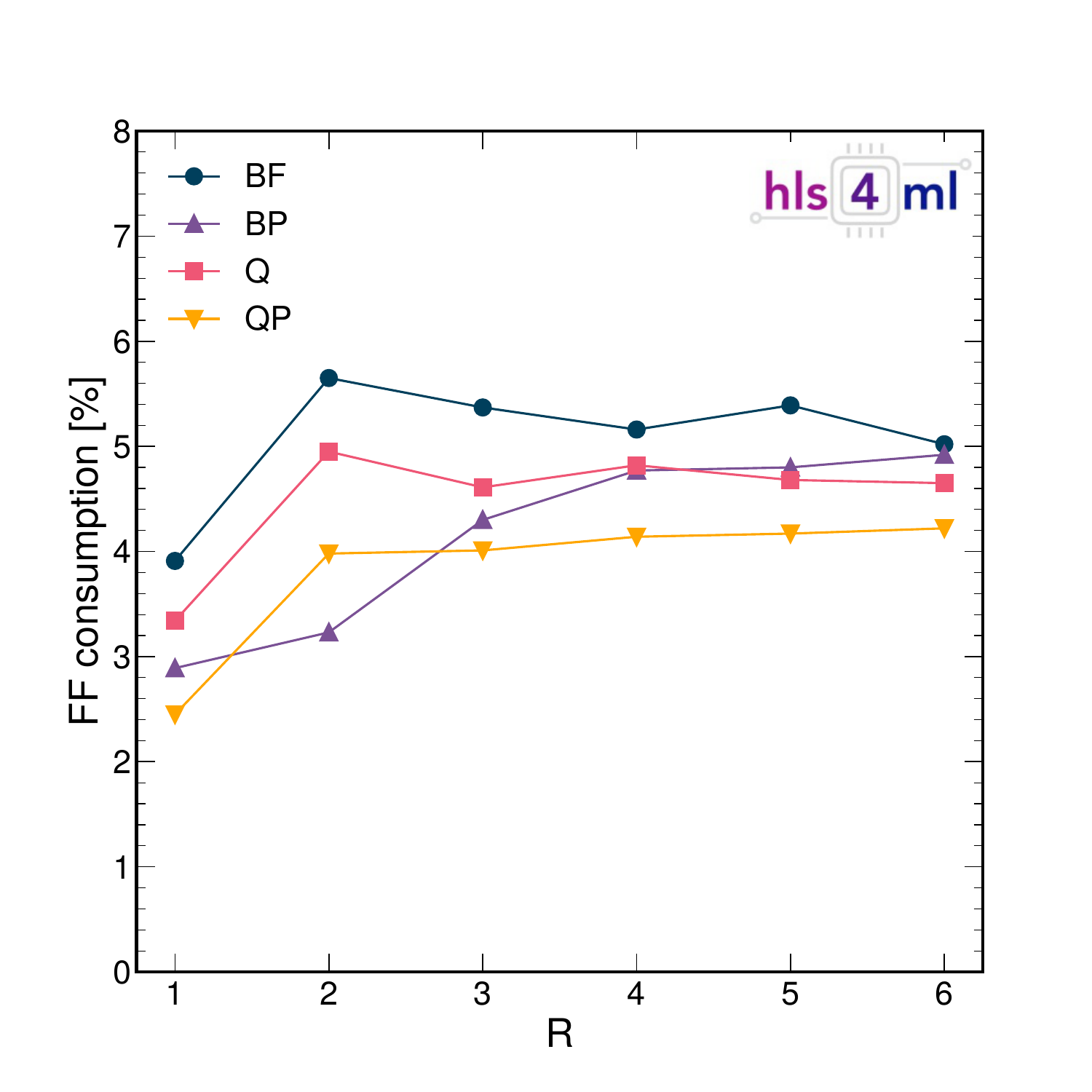}
    \caption{The model latency (top left), DSP (top right), LUT (bottom left) and FF (bottom right) consumption as a function of reuse factor for a fixed bit width of 16 for the Baseline (BF), Baseline Pruned (BP), QKeras (Q), and QKeras Pruned (QP) models. \label{fig:reuse_bw16}}
\end{figure}

A summary of the latency and resource consumption for different reuse factors for all the models at a fixed bit width of 16 is shown in Figure~\ref{fig:reuse_bw16}.
The latency has a linear dependence on the reuse factor, as expected because each multiplier is used in series one reuse factor at the time. 
The DSP consumption decreases as $\sim 1/R$ for all models. 
The first point deviates from this as the maximum number of DSPs are in use, effectively reaching a plateau. 
The LUT consumption is high for a reuse factor of one, complimenting the ceiling reached in DSP consumption at a reuse of one, since the multiplications that do not fit on DSP are moved to LUTs. 
The FF consumption is flat as a function of reuse factor. 
The BRAM consumption does not depend on the reuse factor and is the same for all models, around $3\%$. We leave it up to \hlsfml users to find the optimal trade-off between inference latency and resource consumption for a given application through tuning of the reuse factor.

Although particle physics experiments mostly use large FPGAs, the \hlsfml library can be readily used for smaller FPGAs, like those found on system-on-chip (SoC) or internet-of-things (IoT) devices, through increasing the reuse factor. 
To demonstrate this, we synthesize and deploy the smallest model that retains the original model accuracy, QP 7-bit, onto a low-cost TUL PYNQ-Z2 development board, equipped with a Xilinx Zynq XC7Z020 SoC (FPGA part number \texttt{xc7z020clg400-1}). 
This FPGA is significantly smaller than the Xilinx Virtex UltraScale+ VU9P, and consists of 13,300 logic slices, each with four 6-input LUTs and 8 FFs, 630\unit{kB} of BRAM, and 220 DSP slices. 
As expected, a large reuse factor is needed in order to fit the QP 7-bit model onto the Zynq XC7Z020. For a clock frequency of 100\unit{MHz}, the resulting inference latency is 171\unit{$\mu$s} and up to 2,831 image classifications per second.
This implementation uses a total of 91\% of the LUTs, 97\% of the DSPs, 33\% of the FFs, and 44\% of the BRAM.
A summary is provided in Table~\ref{tab:pynq}.
This demonstrates the flexibility of \hlsfml to accommodate SoC/IoT use cases, which can demand smaller FPGAs and tolerate millisecond latencies.

\begin{table}[ht!]
 \caption{Resource consumption and latency for the QP 7-bit model on a Xilinx Zynq XC7Z020 SoC.
 A clock frequency of 100\unit{MHz} is used. \label{tab:pynq}}
  \centering 
\resizebox{\textwidth}{!}{
\begin{tabular}{l|cccc|ccc}
\multicolumn{7}{l}{FPGA: Xilinx Zynq XC7Z020 SoC}\\
\hline
          &   DSP        & LUT              &FF                & BRAM (18\unit{kb})    & Latency [cc]               & II [cc]        & frame/s\\
\hline
Available & 220          & 53,200          & 106,400            & 280                  &  ---                      & ---           & --- \\
Used      &213 (97\%)    & 48,259 (91\%)   & 35,118 (33\%)      & 122 (44\%)           & 17,085 (171\unit{$\mu$s}) & 16,385        & 2,831  \\
\end{tabular}}
\end{table} 

\begin{figure}[h!]
    \centering
    \includegraphics[width=0.47\textwidth]{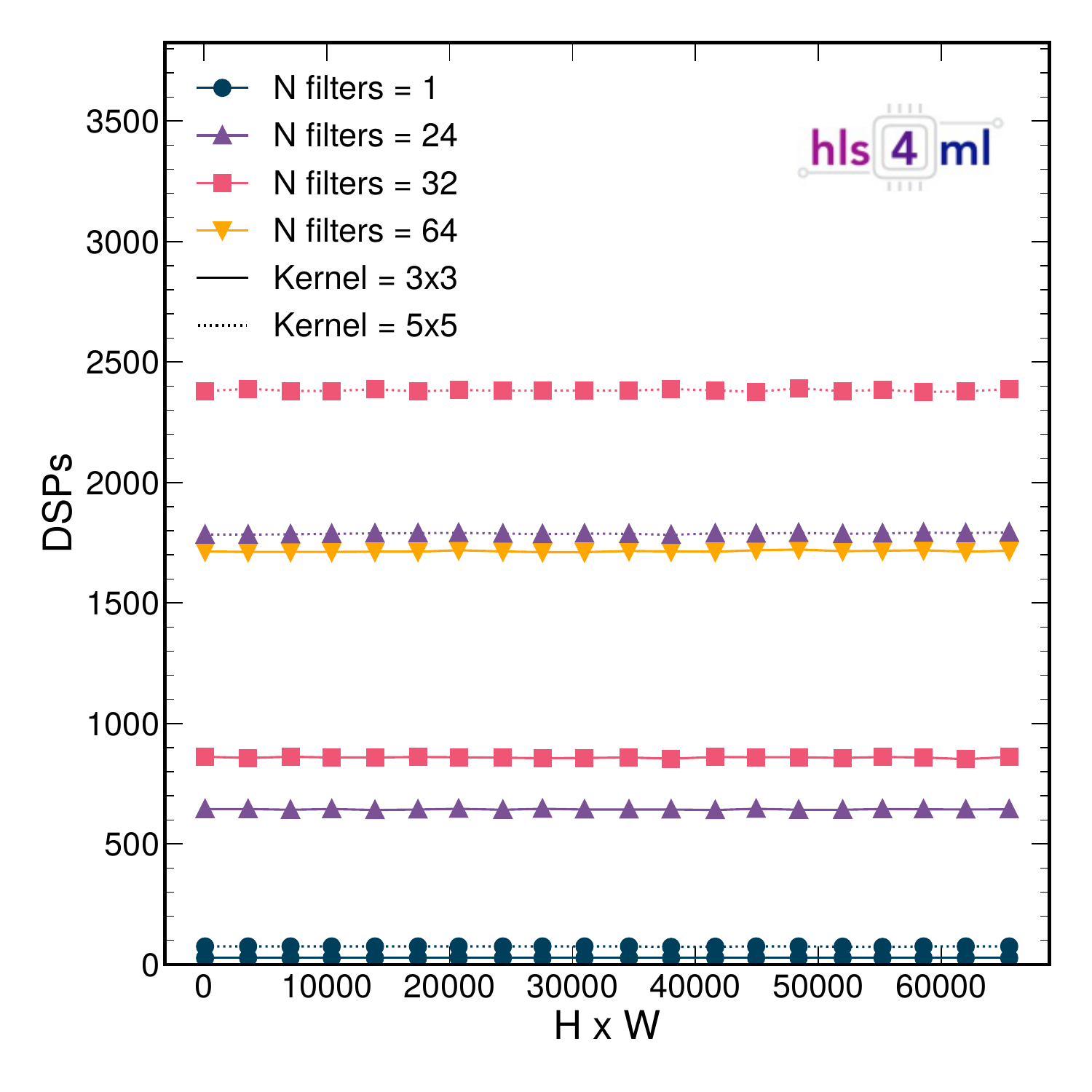}
    \includegraphics[width=0.47\textwidth]{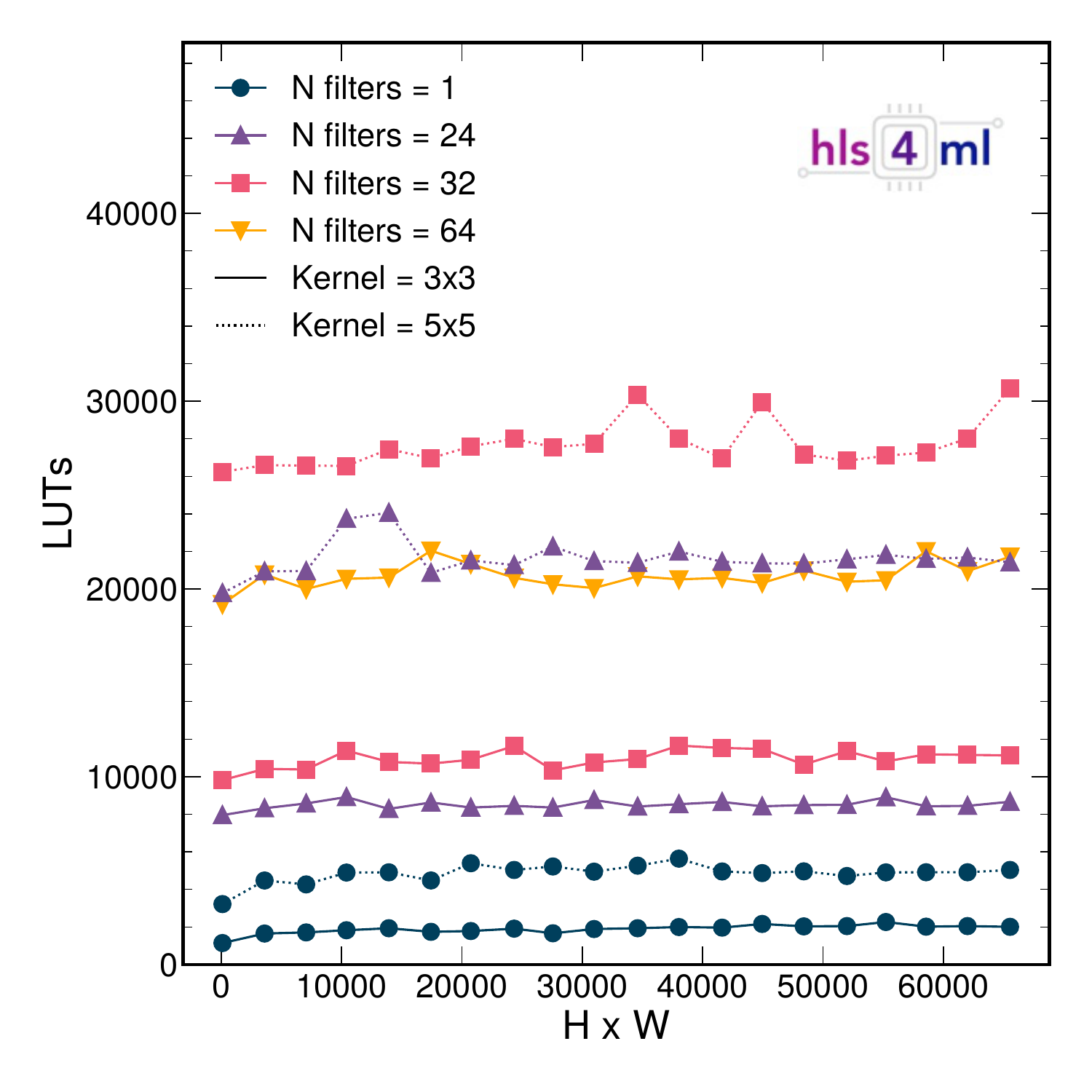}\\
    \includegraphics[width=0.47\textwidth]{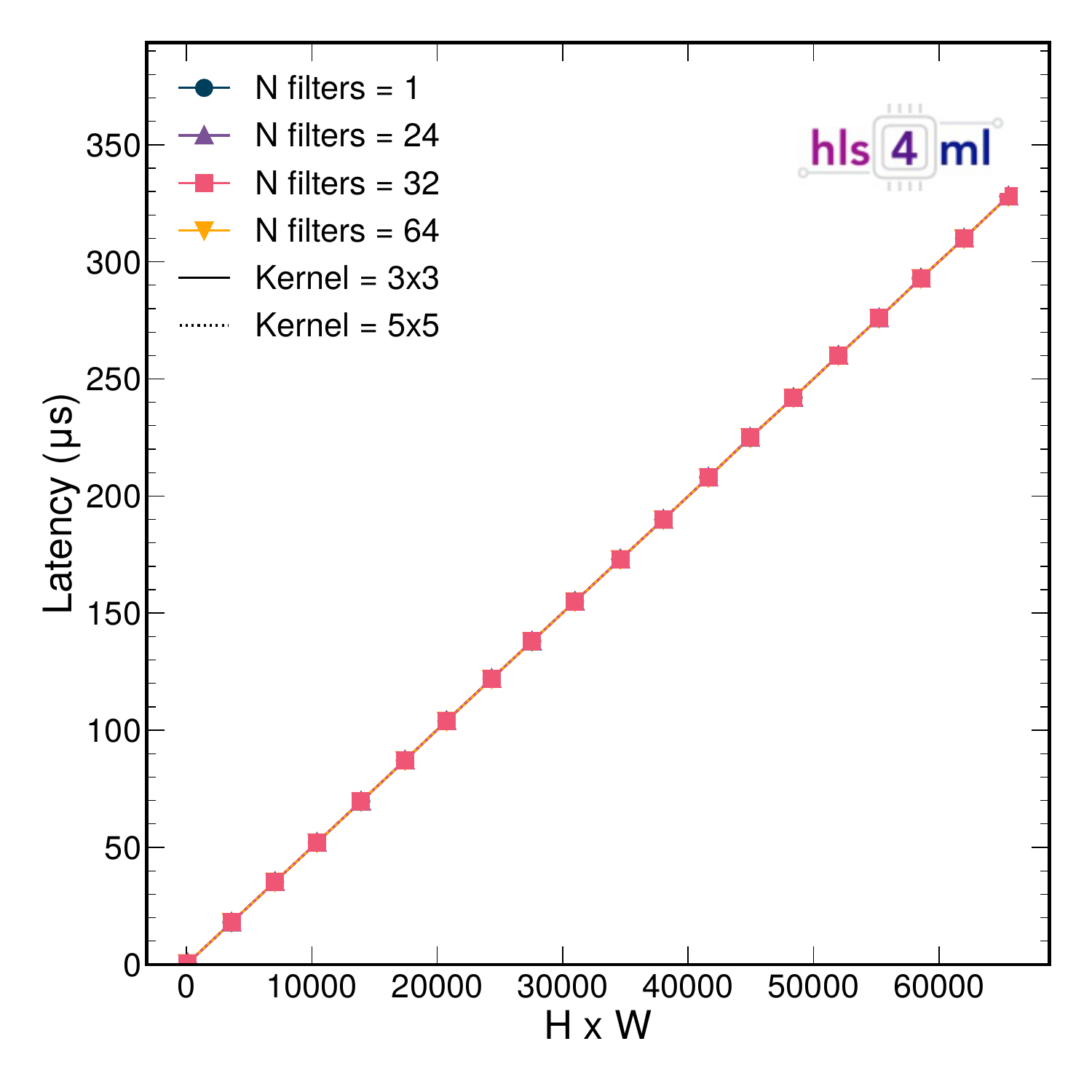}
    \caption{The DSP usage (top left), LUT usage (top right) and latency (bottom) as a function of the input image height ($H$) and width ($W$) for a single convolutional layer with varying kernel size and number of filters. 
    Three color channels are assumed such that the input size corresponds to ($H\times W\times3$). A default precision for all weights and outputs of $\langle 16,6 \rangle$ is assumed.  \label{fig:single_scan}}
\end{figure}

\begin{figure}[h!]
    \centering
    \includegraphics[width=0.47\textwidth]{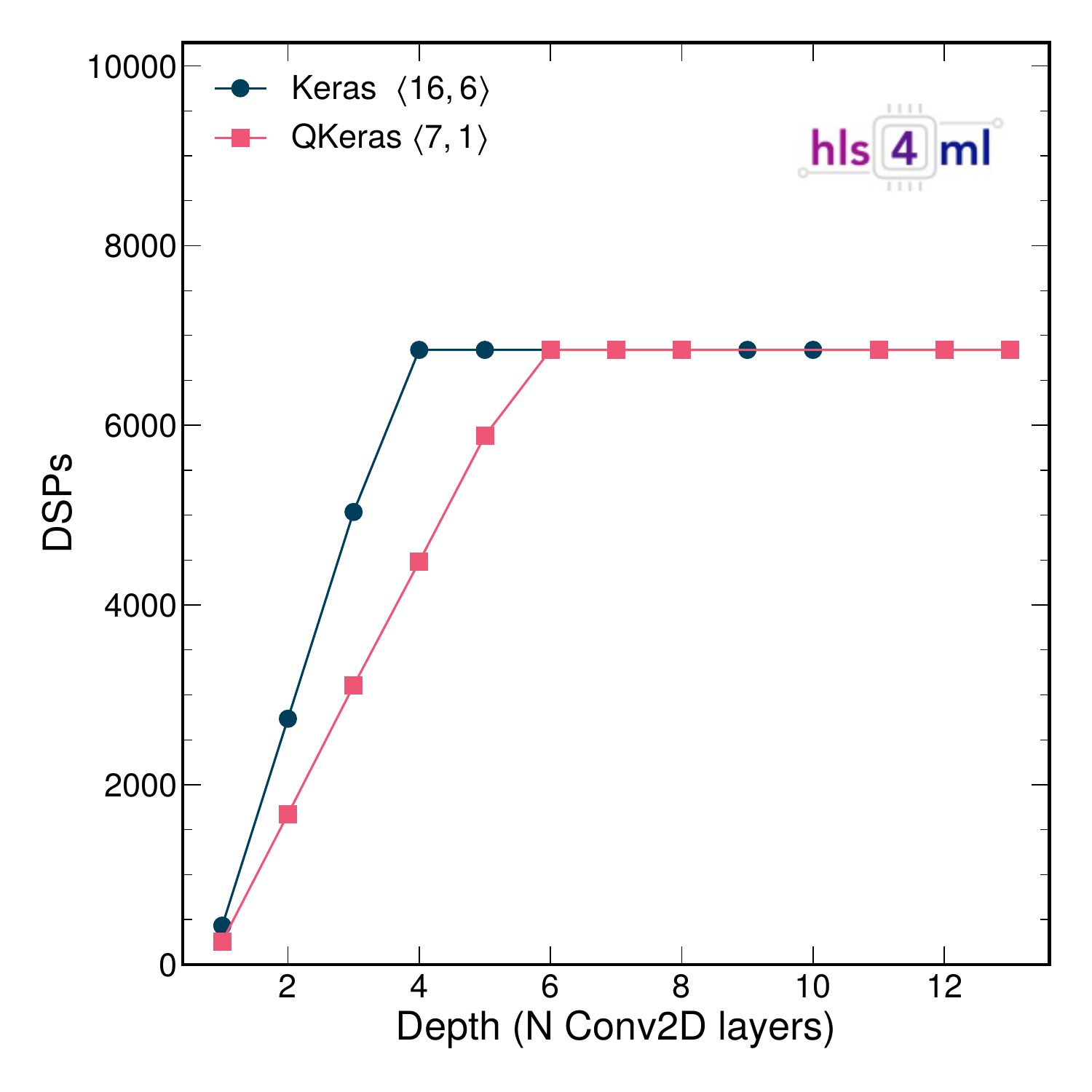}
    \includegraphics[width=0.47\textwidth]{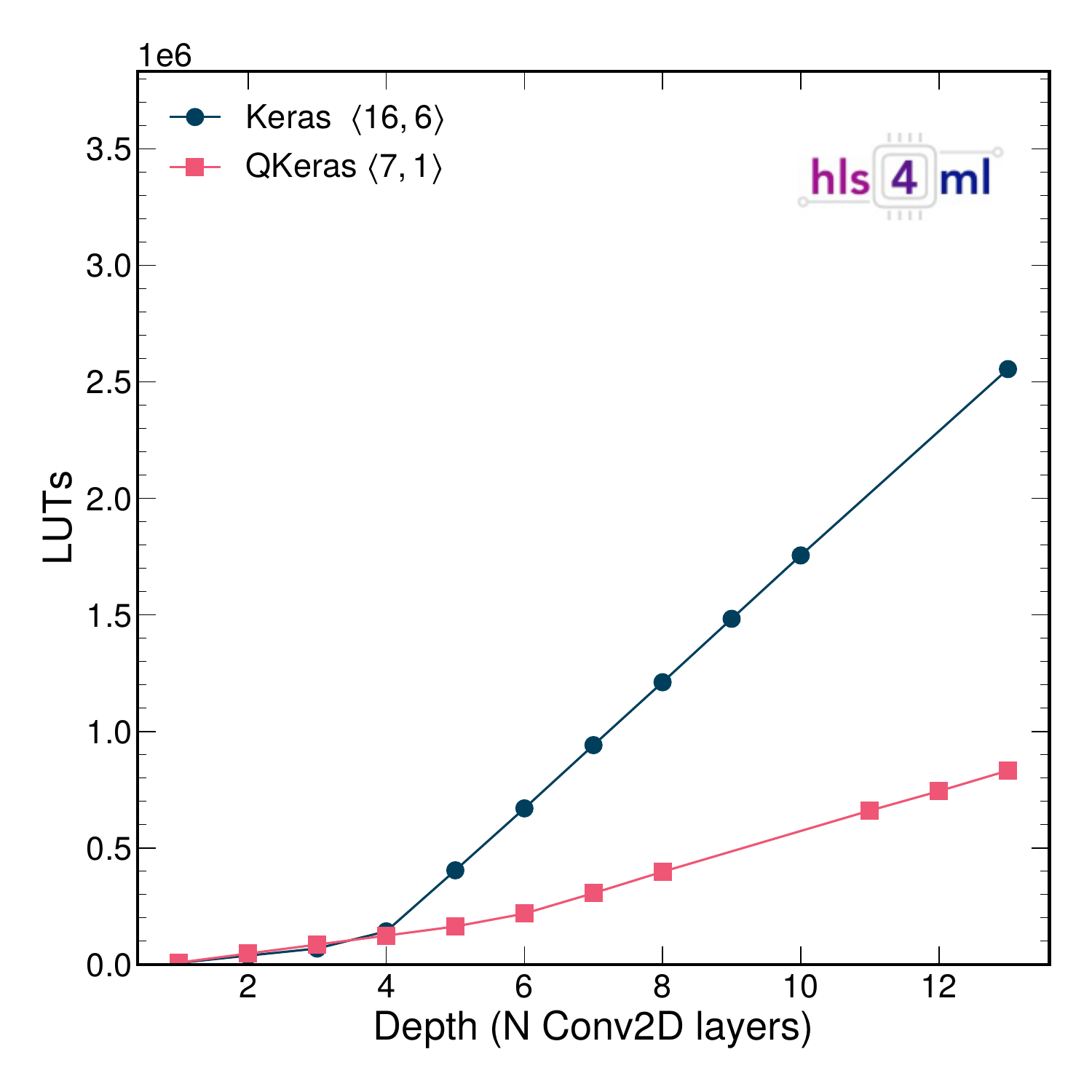}\\
    \includegraphics[width=0.47\textwidth]{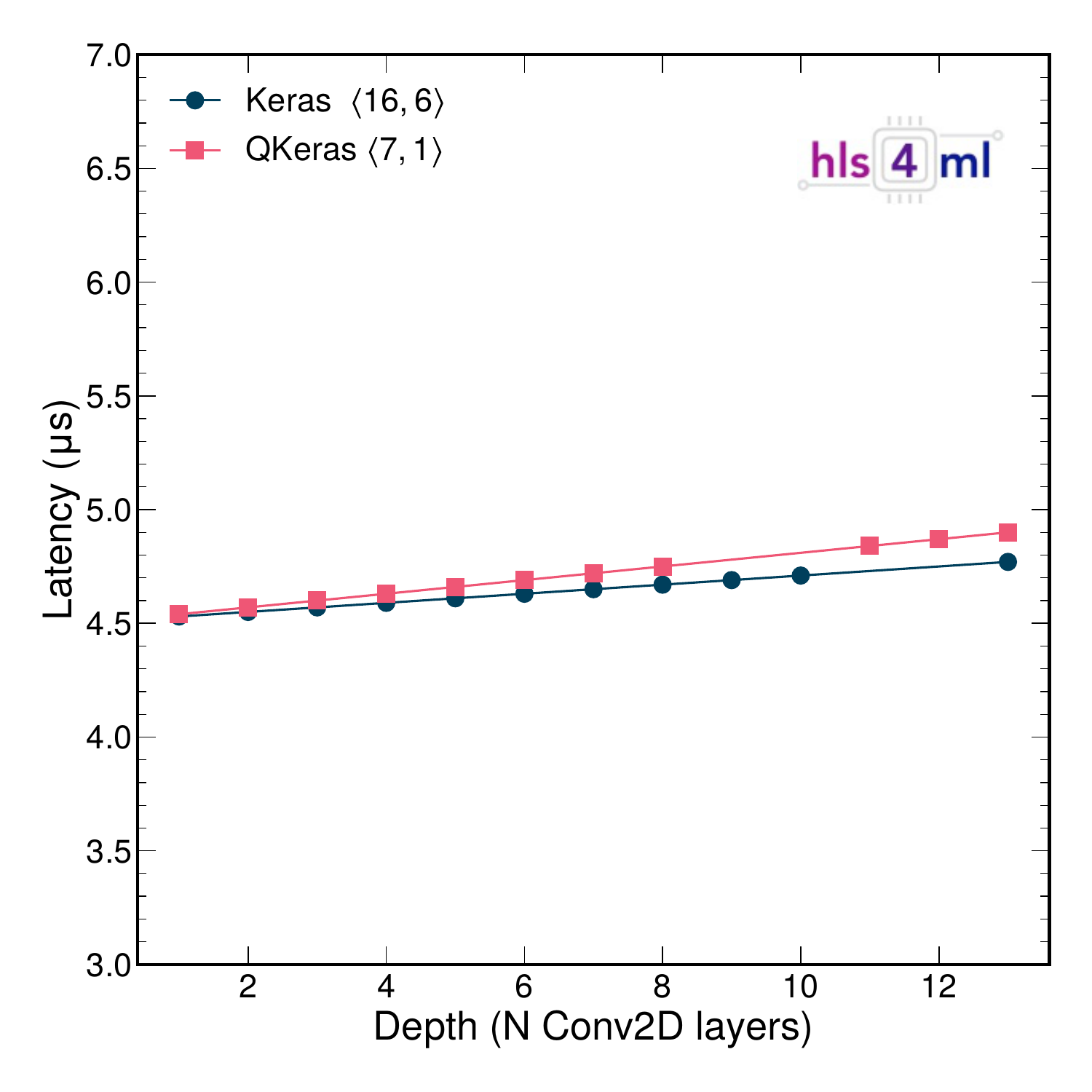}
    \caption{The DSP usage (top left), LUT usage (top right) and latency (bottom) as a function of the model depth for models using a precision of $\langle 16, 6\rangle$ and $\langle 7, 1\rangle$. An input size of ($30 \times 30\times 3$) is always assumed and each convolutional layer consists of 16 filters and a kernel size of ($3\times 3$).  \label{fig:scan_depth}}
\end{figure}

Finally, in Fig.~\ref{fig:single_scan} we study the resource consumption and latency as a function of the input size for a single convolutional layer with varying number of filters and kernel sizes. 
Three input channels are always assumed and the input height ($H$) and width ($W$) is varied between 10 and 256, such that the input size is $H \times W\times3$. 
A precision of $\langle 16,6 \rangle$ is assumed for all models to illustrate the dependency of latency/resources on the given layer configurations, although, as we have demonstrated above, resources can be significantly reduced using QAT. 
The DSP and LUT consumption is constant as a function of the input size, but increases with the number of filters used and the kernel size, due to the higher number of multiplications that need to be performed simultaneously. 
The latency increases linearly with the input size, but does not depend on the kernel size or the number of filters. 
We also show the latency as a function of the depth of the model in Fig.~\ref{fig:scan_depth}. For simplicity, we assume an input size of $30 \times 30\times 3$, 16 filters and a kernel size of $3\times 3$ for each convolutional layer.  
The precision is fixed to $\langle 16,6 \rangle$ or $\langle 7,1 \rangle$. The DSP consumption scales linearly with the model depth until the maximum number of DSPs are used. 
When all DSPs are in use, multiplications are moved onto LUTs, seen as a change of slope in the LUT consumption versus model depth. 
The inference latency increases linearly with the model depth.

Figures~\ref{fig:single_scan} and~\ref{fig:scan_depth} summarize how input size and model architecture affects the inference latency and resource consumption. 
Through increasing the reuse factor, smaller FPGAs can be targeted through a trade-off between latency and resource consumption. 
Support for QAT through and pruning further reduce the model footprint. 
The \hlsfml library is therefore capable of providing generic, multi-backend support for a wide range of hardware and latency constraints.

\section{Conclusions}
\label{sec:conclusion}

We have presented the extension of \hlsfml to support convolutional neural network (CNN) architectures for transpilation to FPGA designs, through a stream-based implementation of convolutional and pooling layers. 
A fully on-chip design is used in order to provide for microsecond latency applications, like those at the CERN Large Hadron Collider.
Taking as a benchmark example a CNN classifier trained on the Street View House Numbers Dataset, we show how compression techniques at training time (pruning and quantization-aware training) reduce the resource utilization of the FPGA-converted models, while retaining to a large extent the floating-point precision baseline accuracy. 
Once converted to FPGA firmware using \hlsfml, these models can be executed with 5\unit{$\mu$s} latency and a comparable initiation interval, while consuming less than 10\% of the FPGA resources. We demonstrate the flexibility and scalability of \hlsfml to accommodate CNN architectures of varying sizes, and offer solutions both for small System-on-Chip (SoC) FPGAs and for the larger FPGAs used in particle physics experiments.
This work enables domain machine learning (ML) specialists to design hardware-optimized ML algorithms for low-latency, low-power, or radiation-hard systems, for instance particle physics trigger systems, autonomous vehicles, or inference accelerators for space applications.

\section*{Acknowledgement}

 We acknowledge the Fast Machine Learning collective as an open community of multi-domain experts and collaborators. 
 This community was important for the development of this project. 

M.~P., S.~S. and V.~L. are supported by the European Research Council (ERC) under the European Union's Horizon 2020 research and innovation program (Grant Agreement No. 772369).
S.~J., M.~L., K.~P., and N.~T. are supported by Fermi Research Alliance, LLC under Contract No. DE-AC02-07CH11359 with the U.S. Department of Energy (DOE), Office of Science, Office of High Energy Physics.
P.~H. is supported by a Massachusetts Institute of Technology University grant. 
Z.~W. is supported by the National Science Foundation under Grants No. 1606321 and 115164.
J.~D. is supported by the DOE, Office of Science, Office of High Energy Physics Early Career Research program under Award No. DE-SC0021187.

\section*{Code availability statement}
The \hlsfml library is available at \url{https://github.com/fastmachinelearning/hls4ml} and archived in the Zenodo platform at \doi{10.5281/zenodo.4161550}.
The work presented here is based on the Bartsia release, version 0.5.0. 
For examples on how to use \hlsfml, the notebooks in \url{https://github.com/fastmachinelearning/hls4ml-tutorial} serve as a general introduction. 
The \QKeras library, which also includes \AutoQKeras and \QTools, is available at \url{https://github.com/google/qkeras}. 

\section*{Data availability statement}
The data that support the findings of this study are openly available. 
The SVHN dataset~\cite{Netzer2011} can be downloaded at \url{http://ufldl.stanford.edu/housenumbers} or through \TensorFlow Datasets at \url{https://www.tensorflow.org/datasets/catalog/svhn_cropped}.
\clearpage

\appendix
\section{Performance versus bit width and reuse factor}
\label{app:reusescan}
\begin{figure}[ht!]
    \centering
    \includegraphics[width=0.41\textwidth]{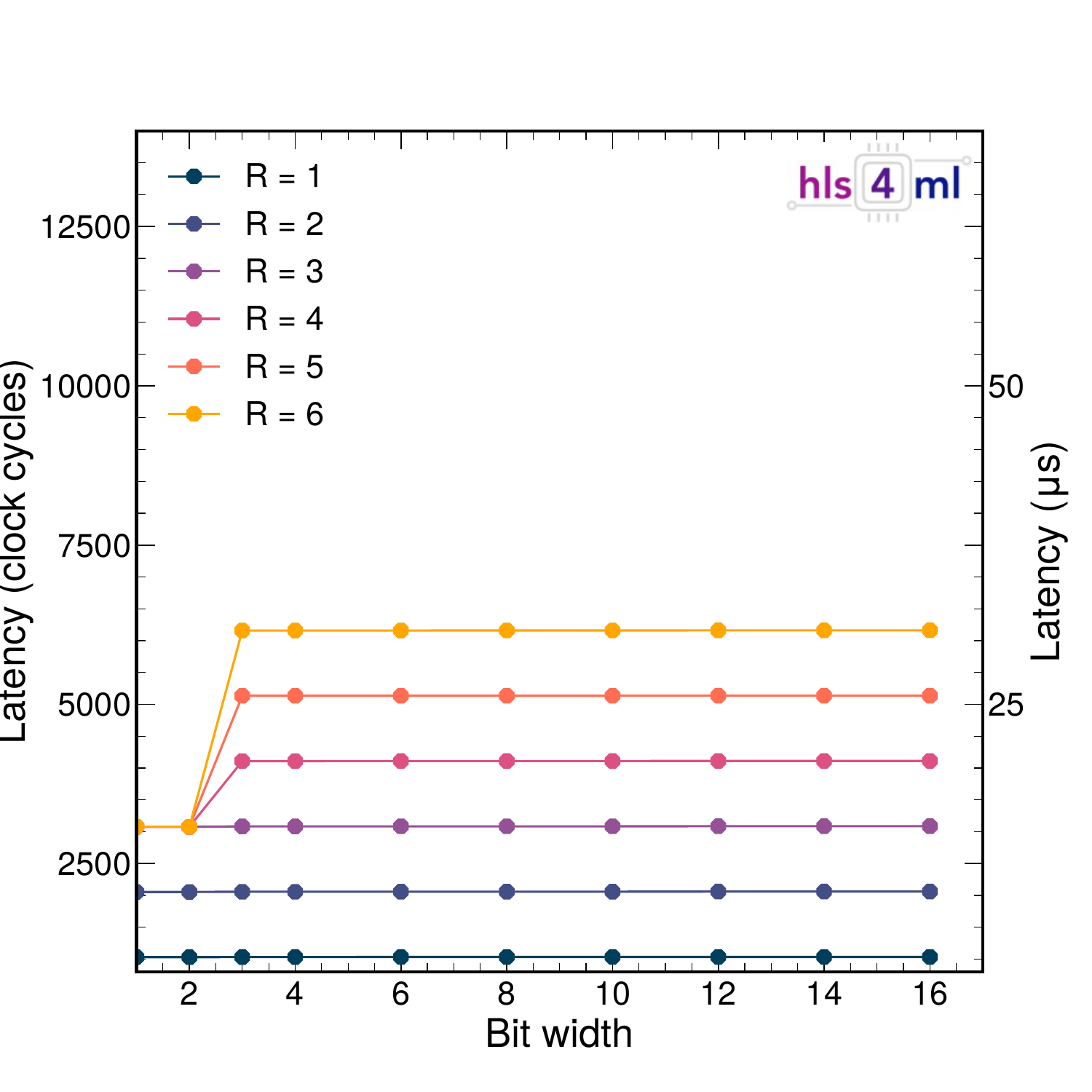}
    \includegraphics[width=0.41\textwidth]{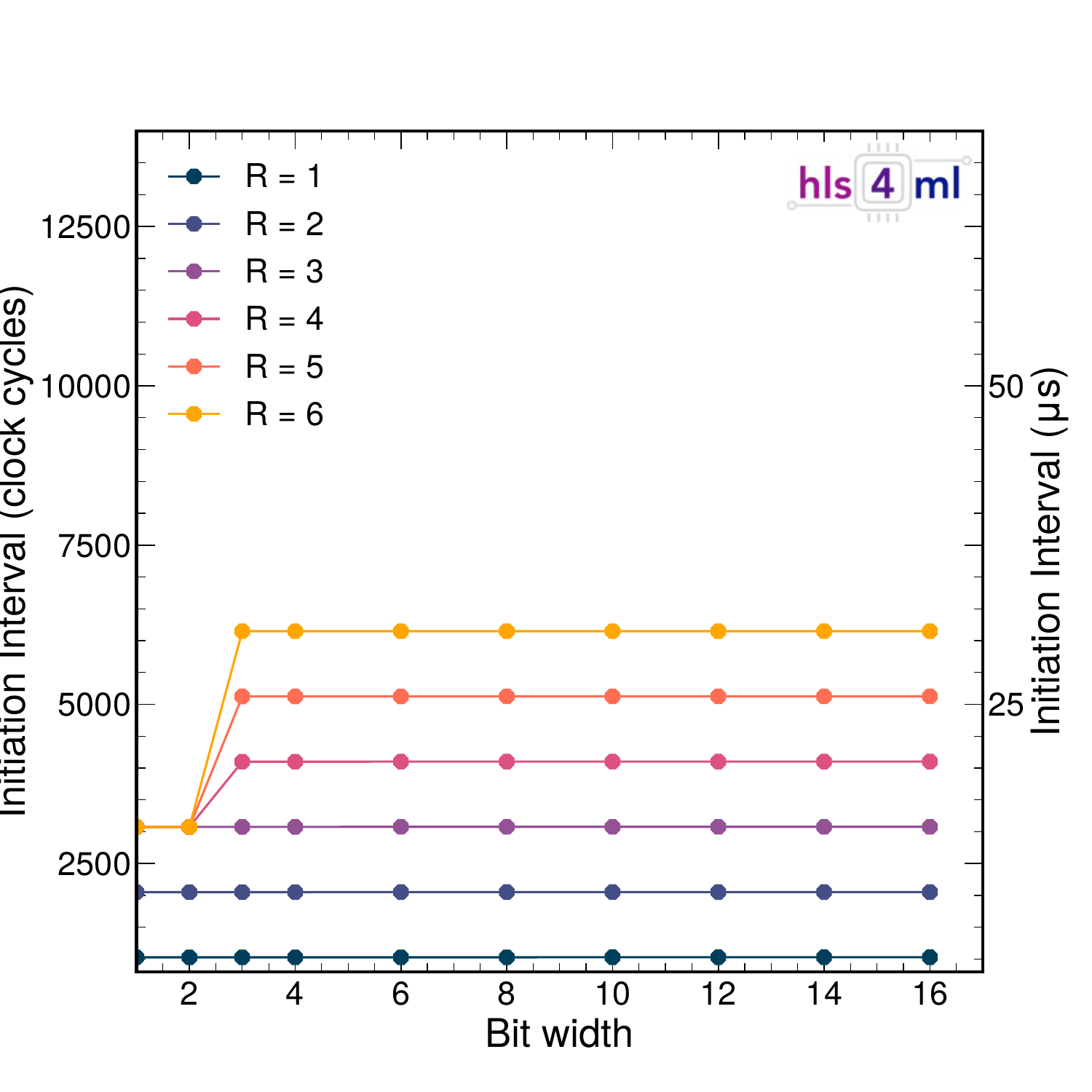}\\
    \includegraphics[width=0.41\textwidth]{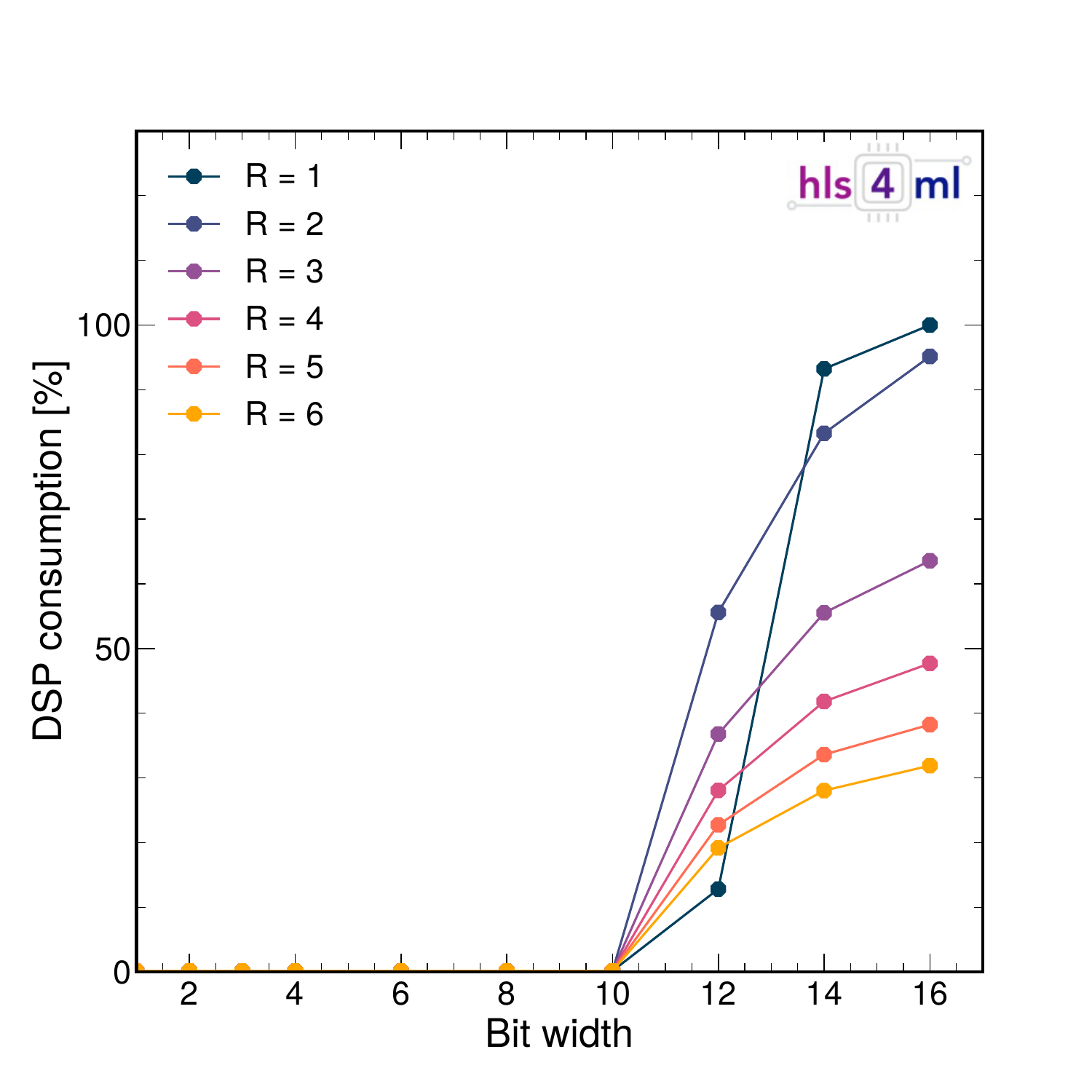}
    \includegraphics[width=0.41\textwidth]{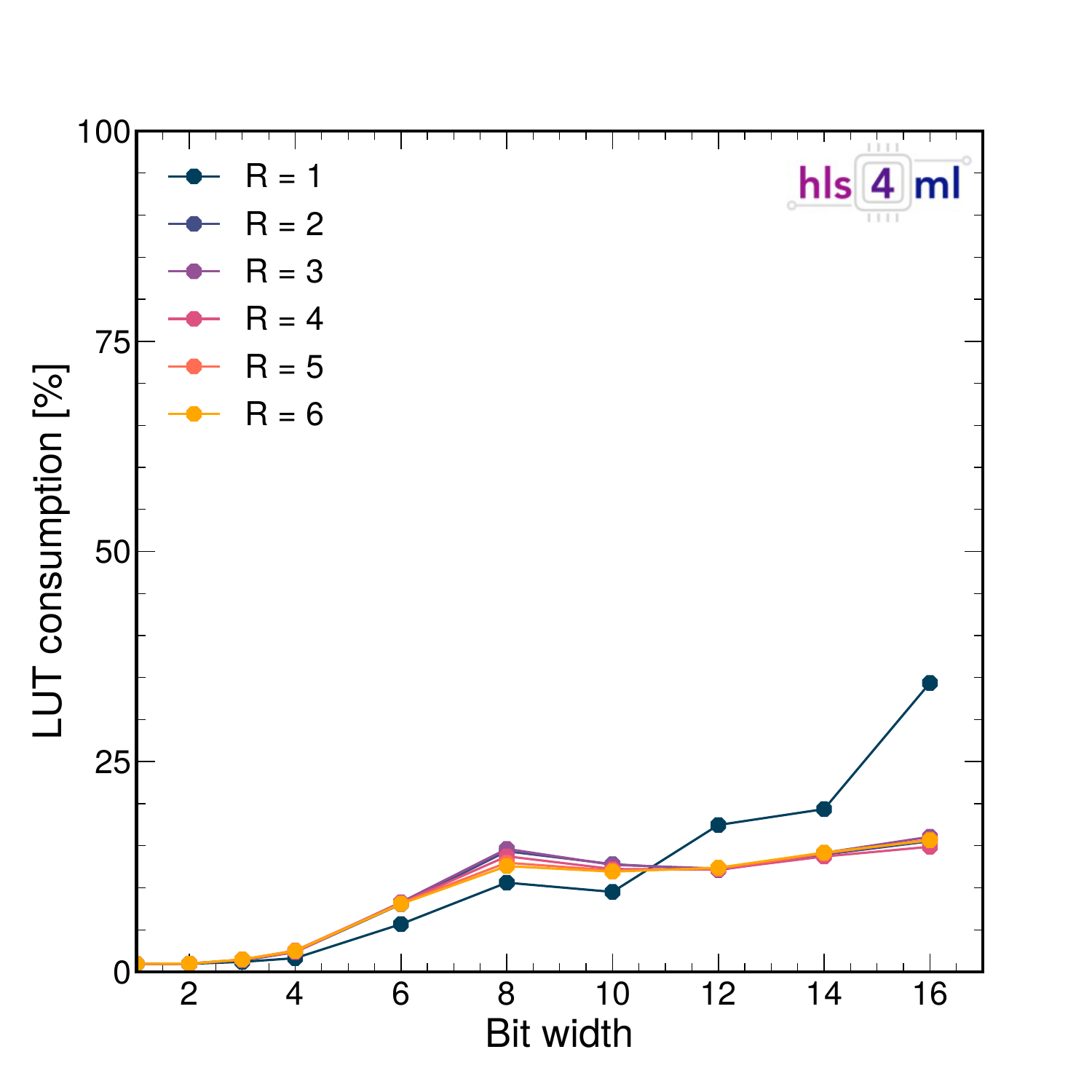}\\
    \includegraphics[width=0.41\textwidth]{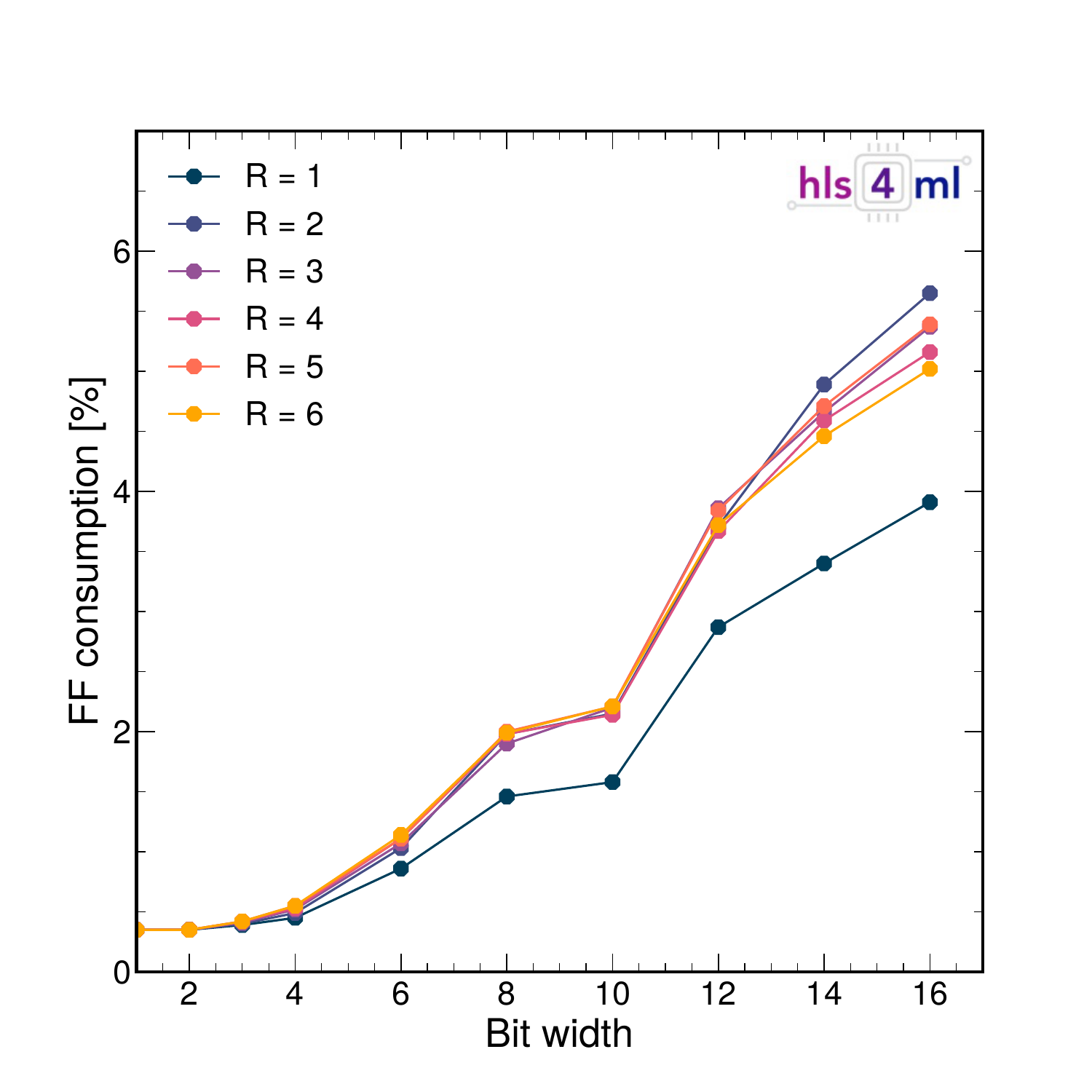}
    \caption{The model latency (top left), initiation interval (top right), DSP (middle left), LUT (middle right) and FF (bottom) consumption as a function of bit width and for different reuse factors for the Baseline Floating-point (BF) model. \label{fig:full_per_reuse}}
\end{figure}

\begin{figure}[ht!]
    \centering
    \includegraphics[width=0.41\textwidth]{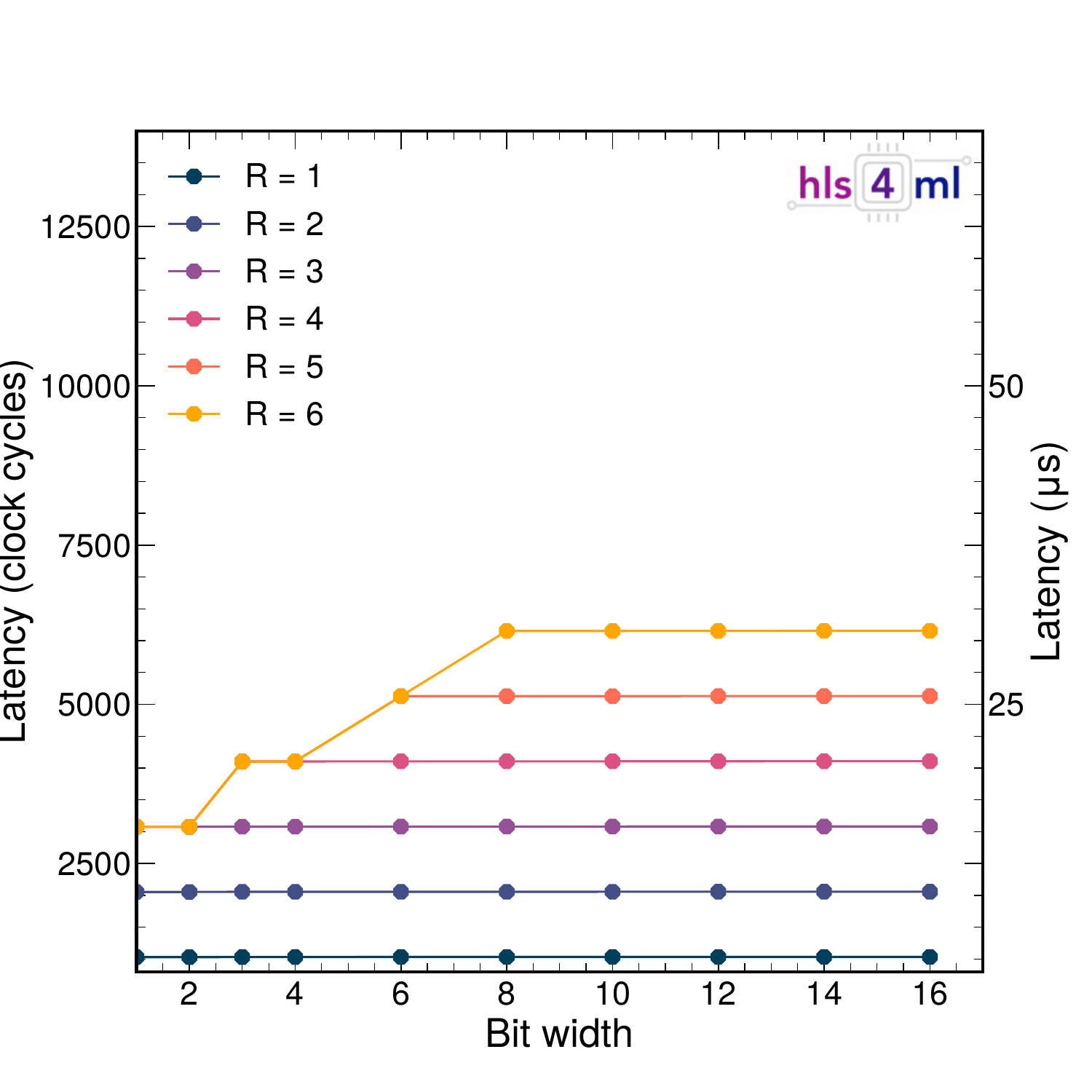}
    \includegraphics[width=0.41\textwidth]{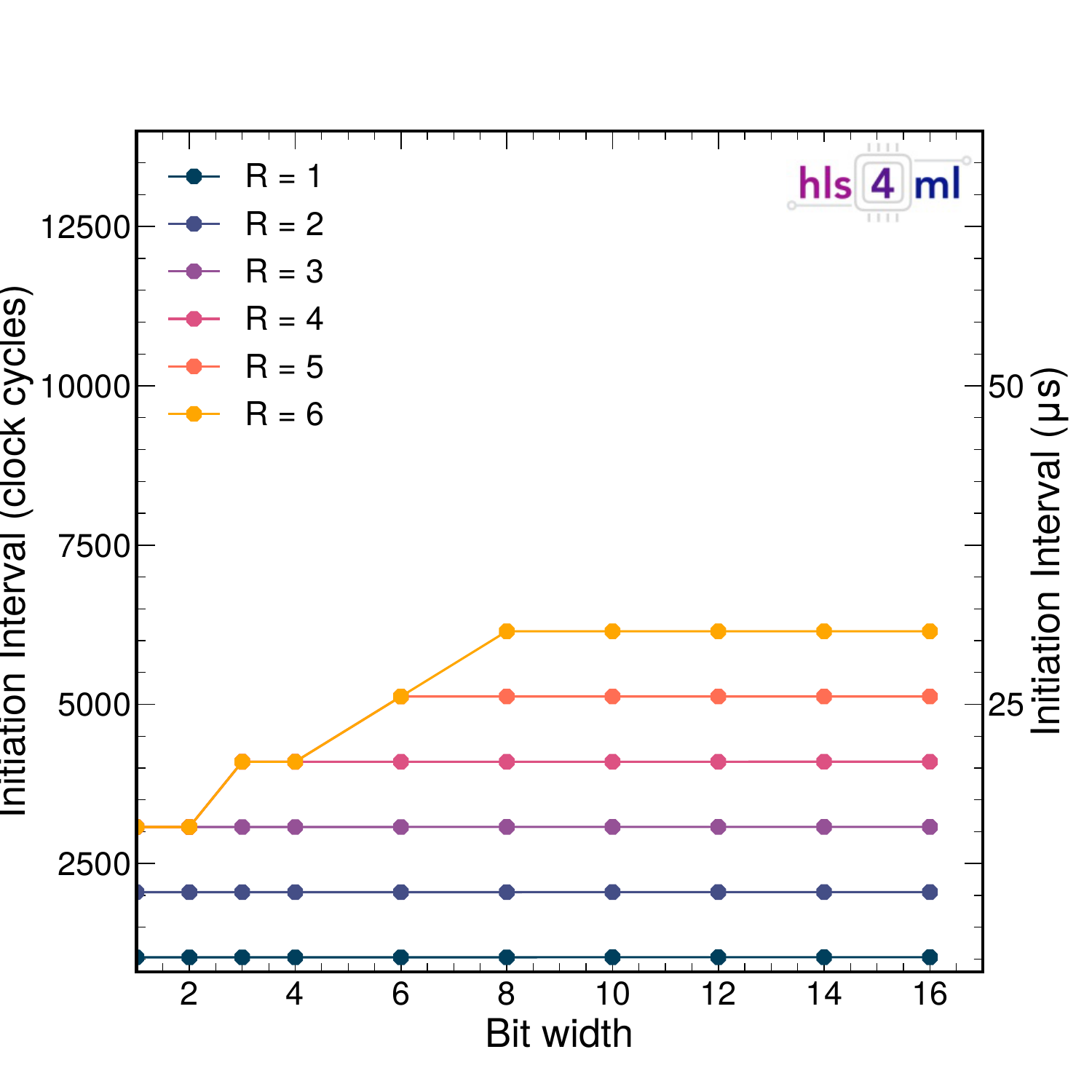}\\
    \includegraphics[width=0.41\textwidth]{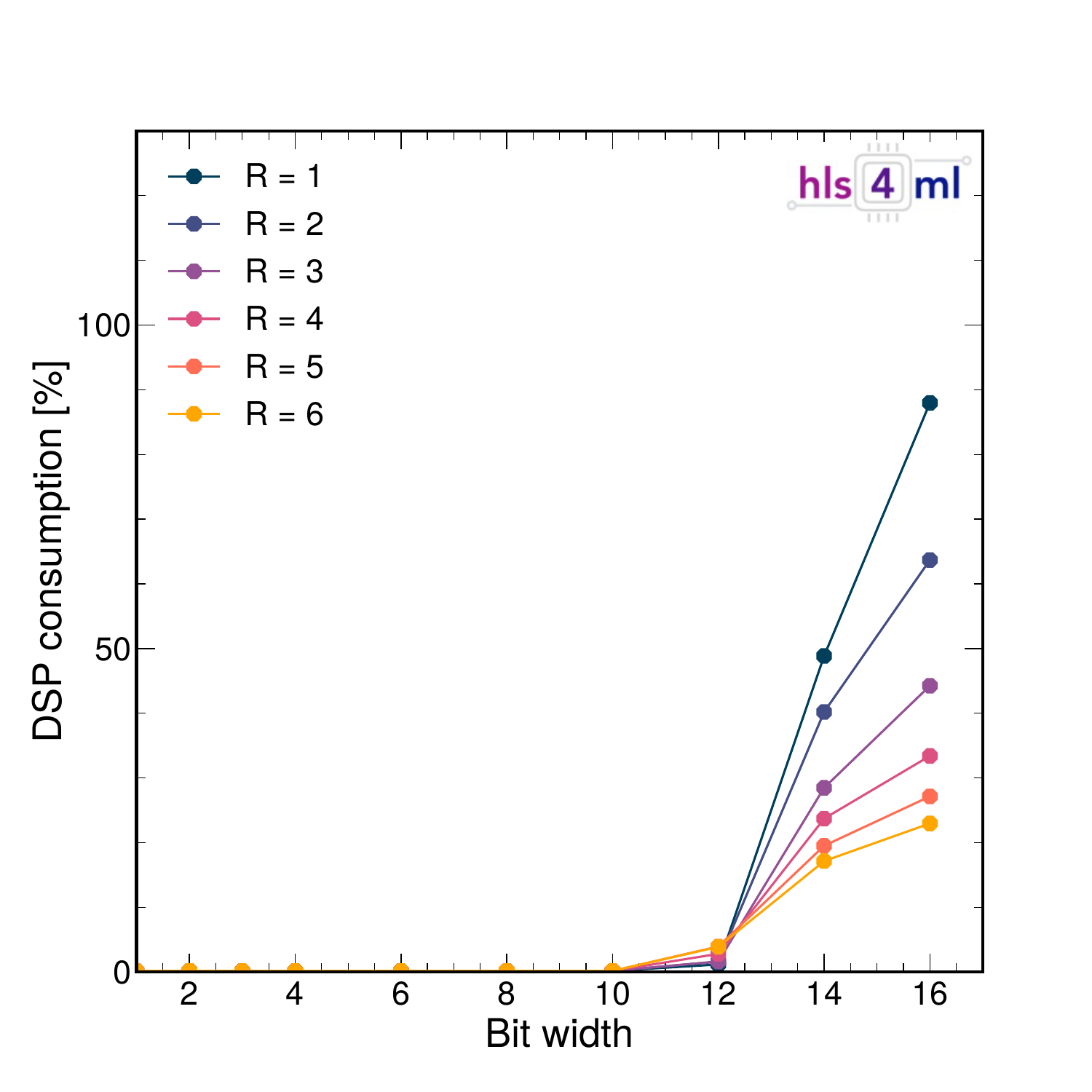}
    \includegraphics[width=0.41\textwidth]{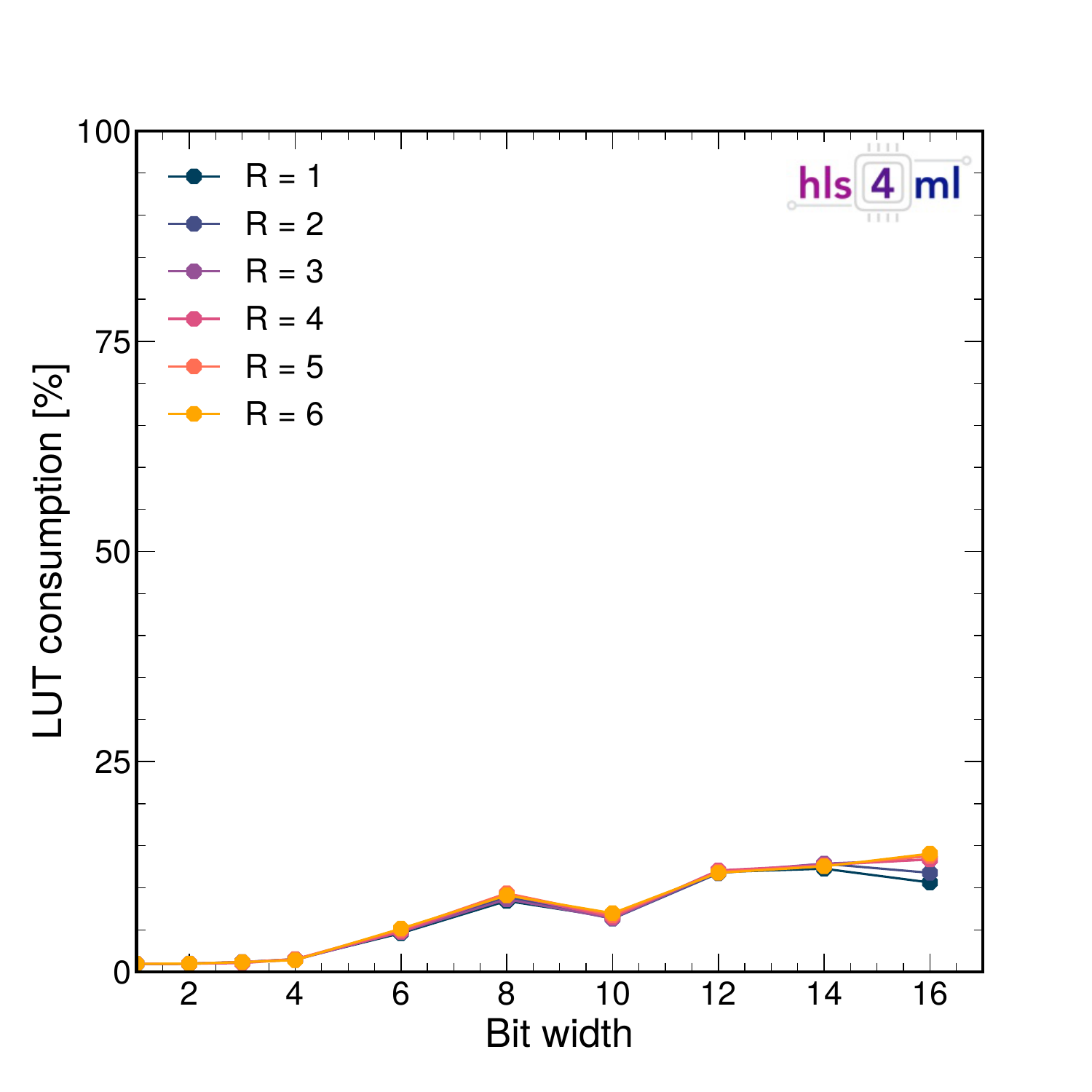}  \\
    \includegraphics[width=0.41\textwidth]{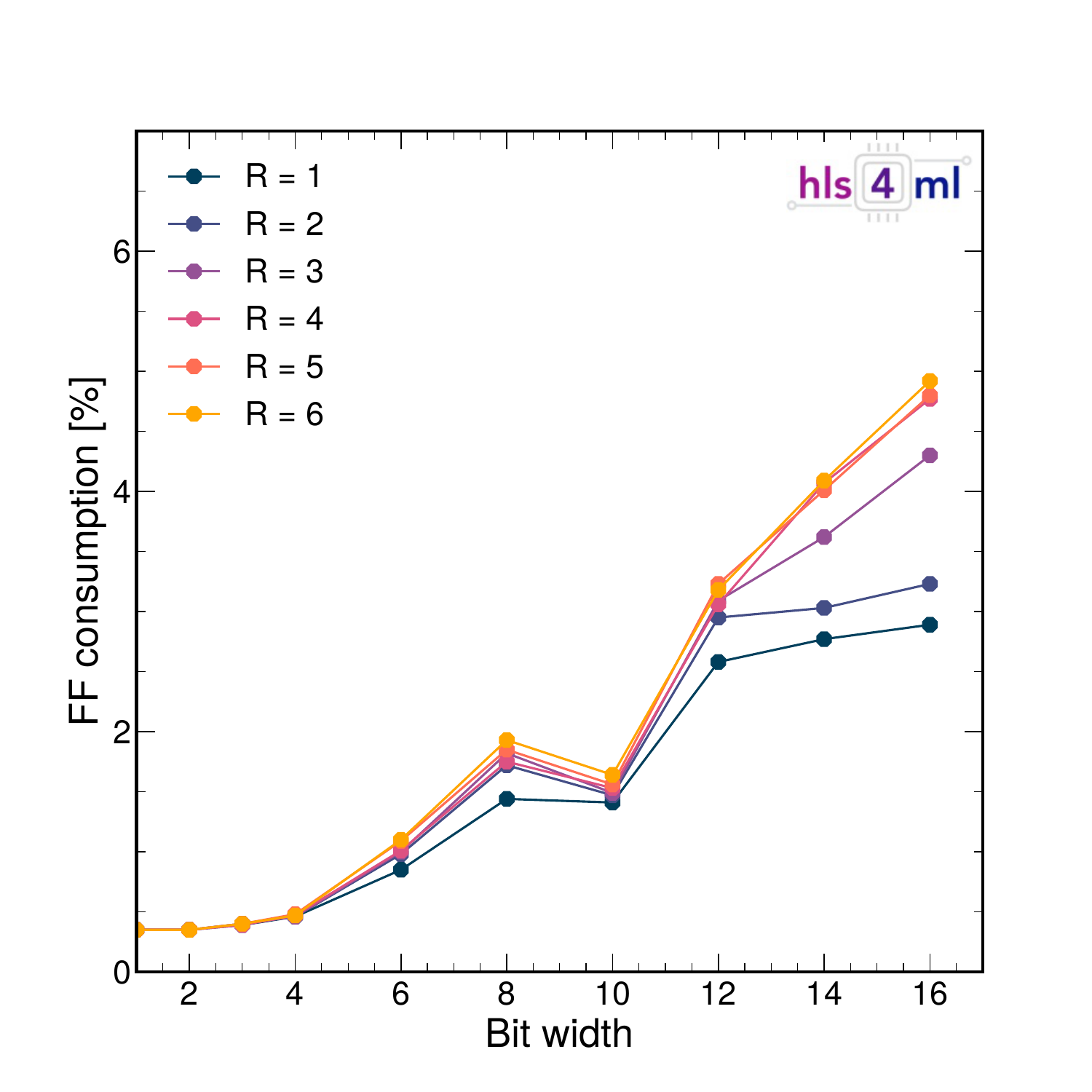}
        \caption{The model latency (top left), initiation interval (top right), DSP (middle left), LUT (middle right) and FF (bottom left)consumption as a function of bit width and for different reuse factors for the Baseline Pruned (BP) model. \label{fig:pruned_full_per_reuse}}
\end{figure}

\begin{figure}[ht!]
    \centering
    \includegraphics[width=0.41\textwidth]{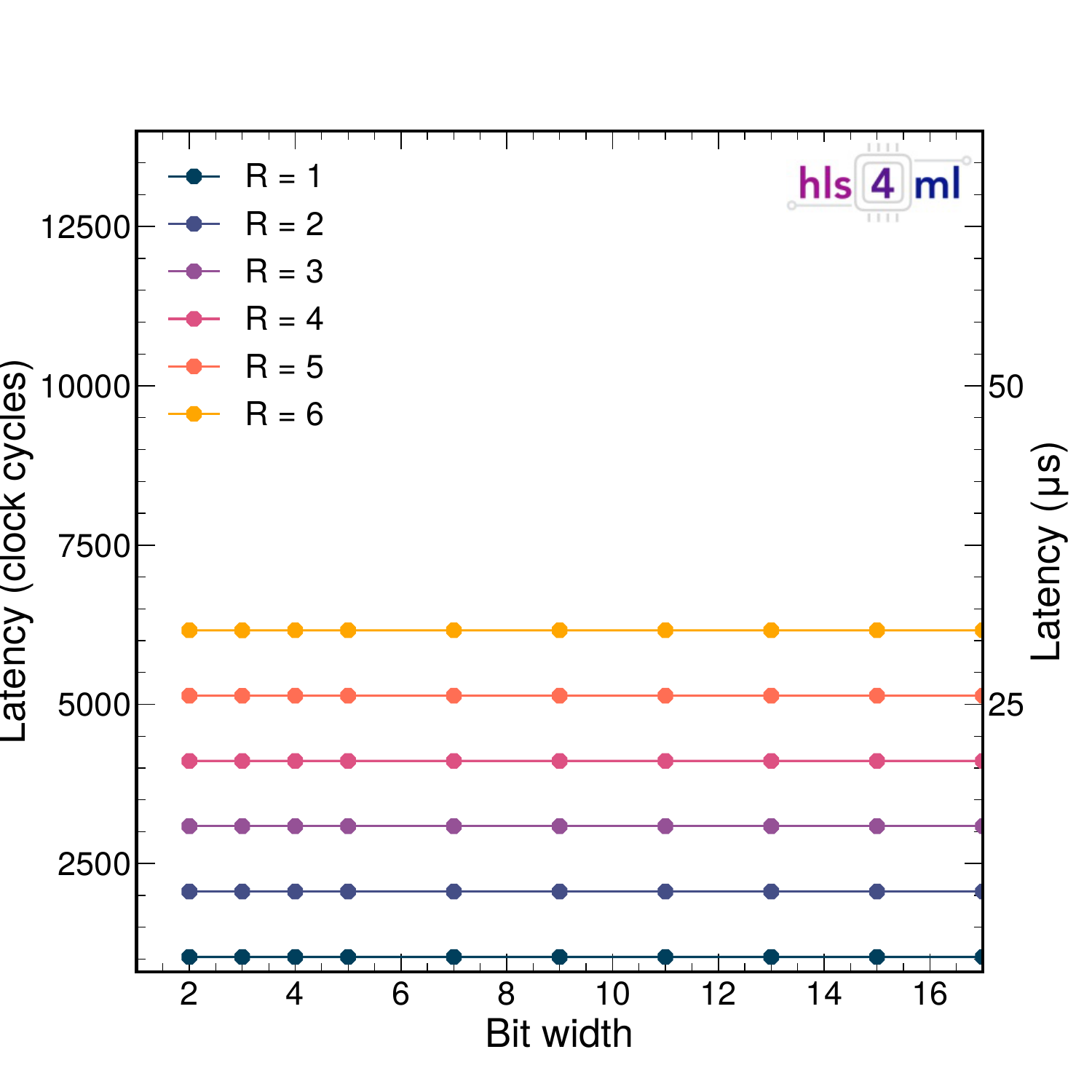}
    \includegraphics[width=0.41\textwidth]{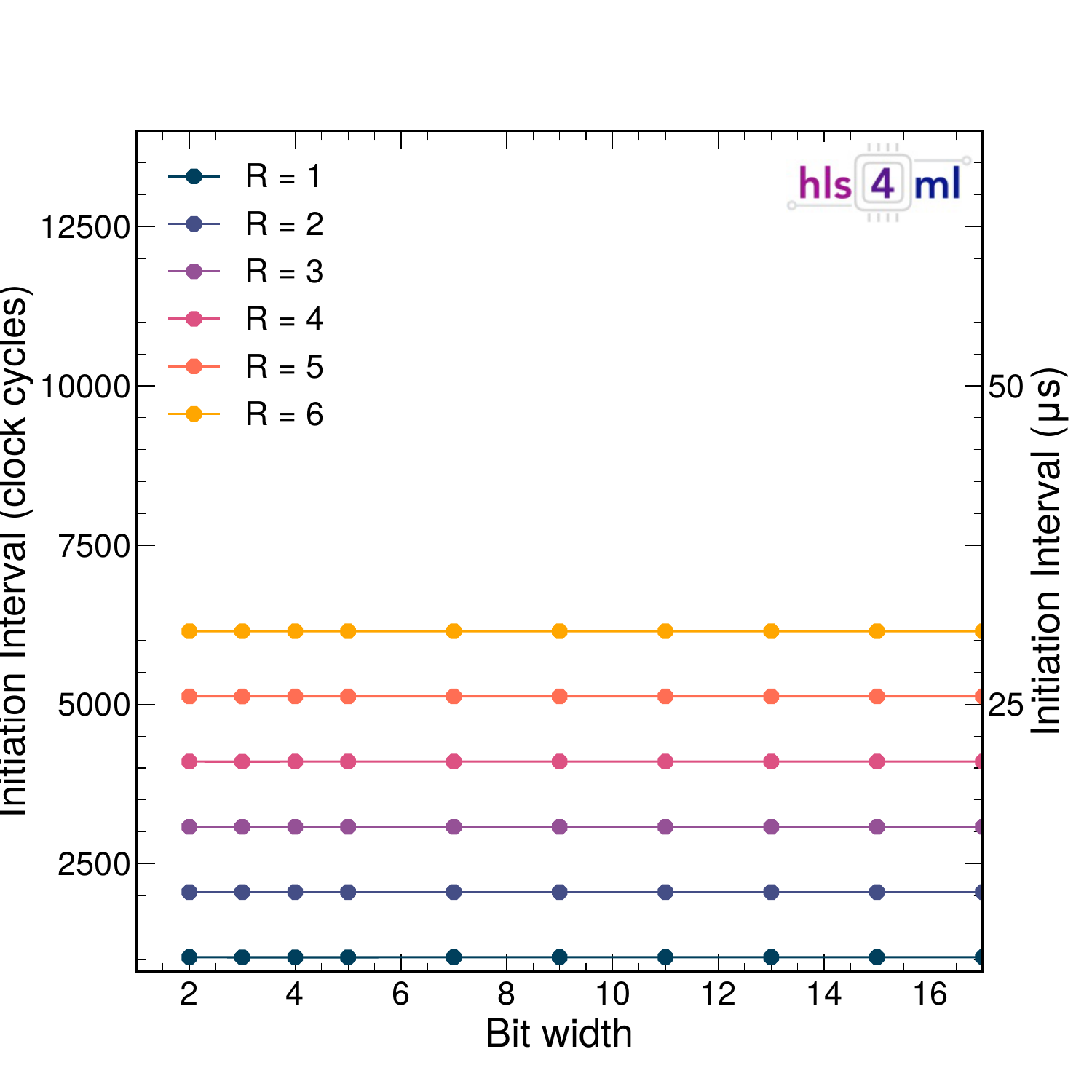}\\
    \includegraphics[width=0.41\textwidth]{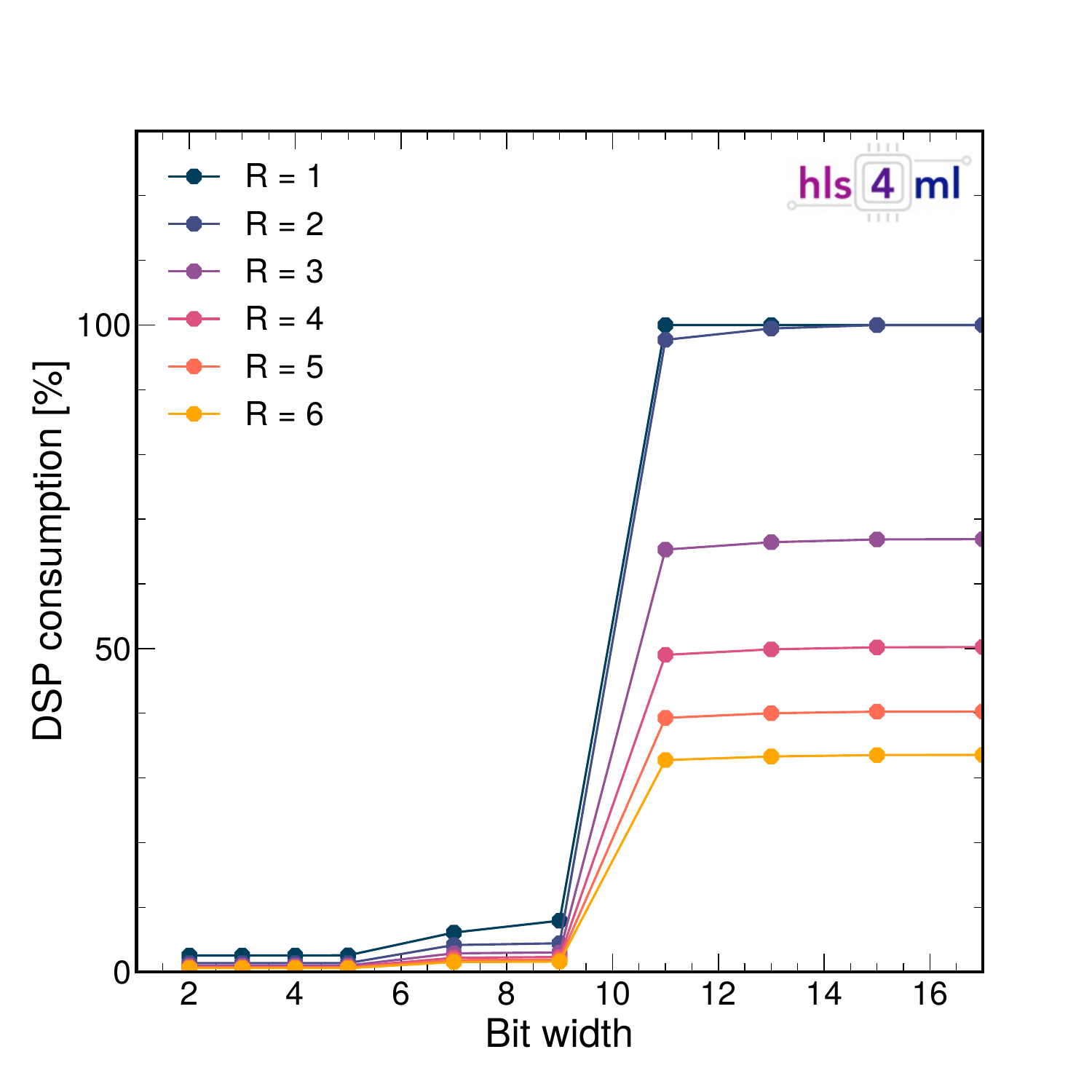}
    \includegraphics[width=0.41\textwidth]{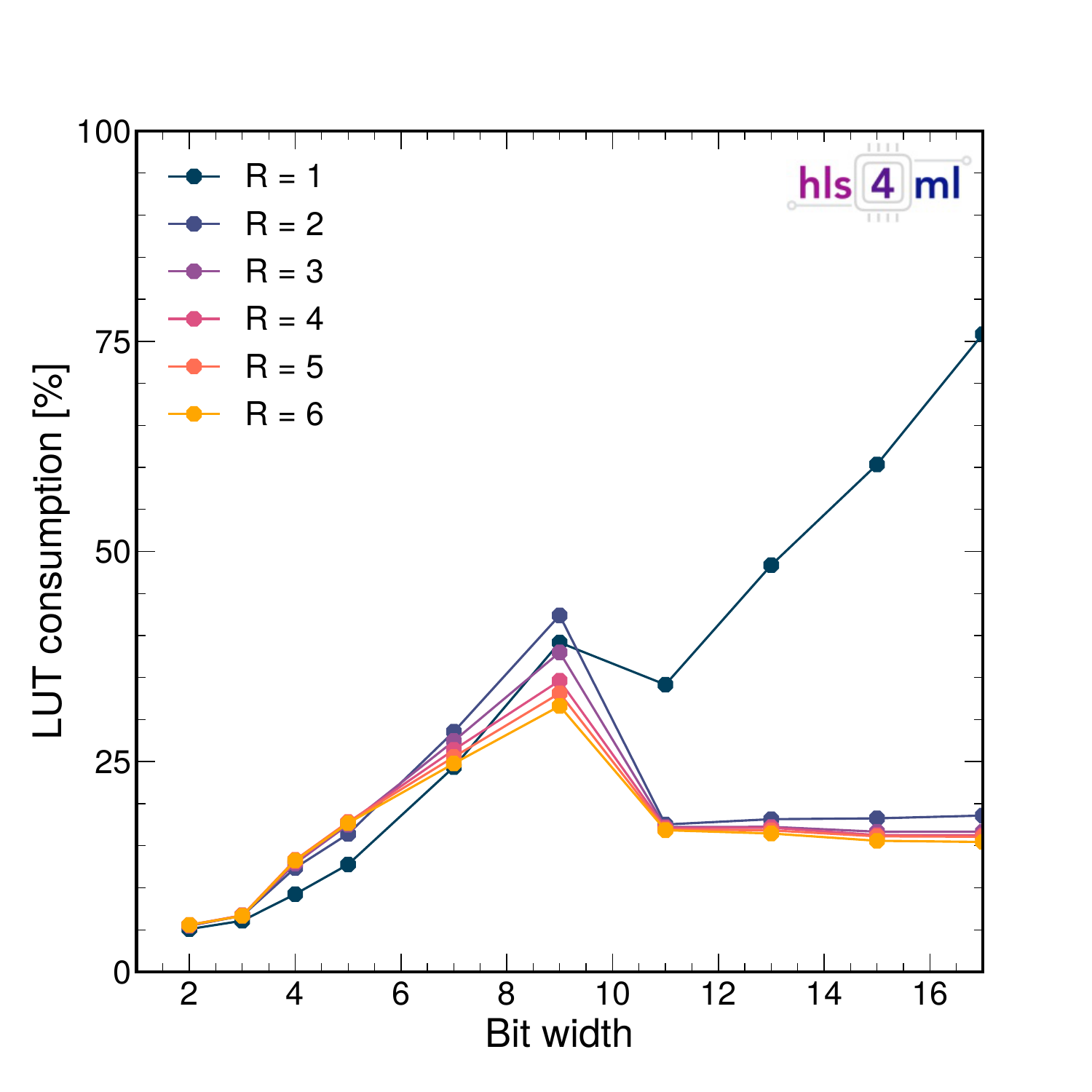}\\
    \includegraphics[width=0.41\textwidth]{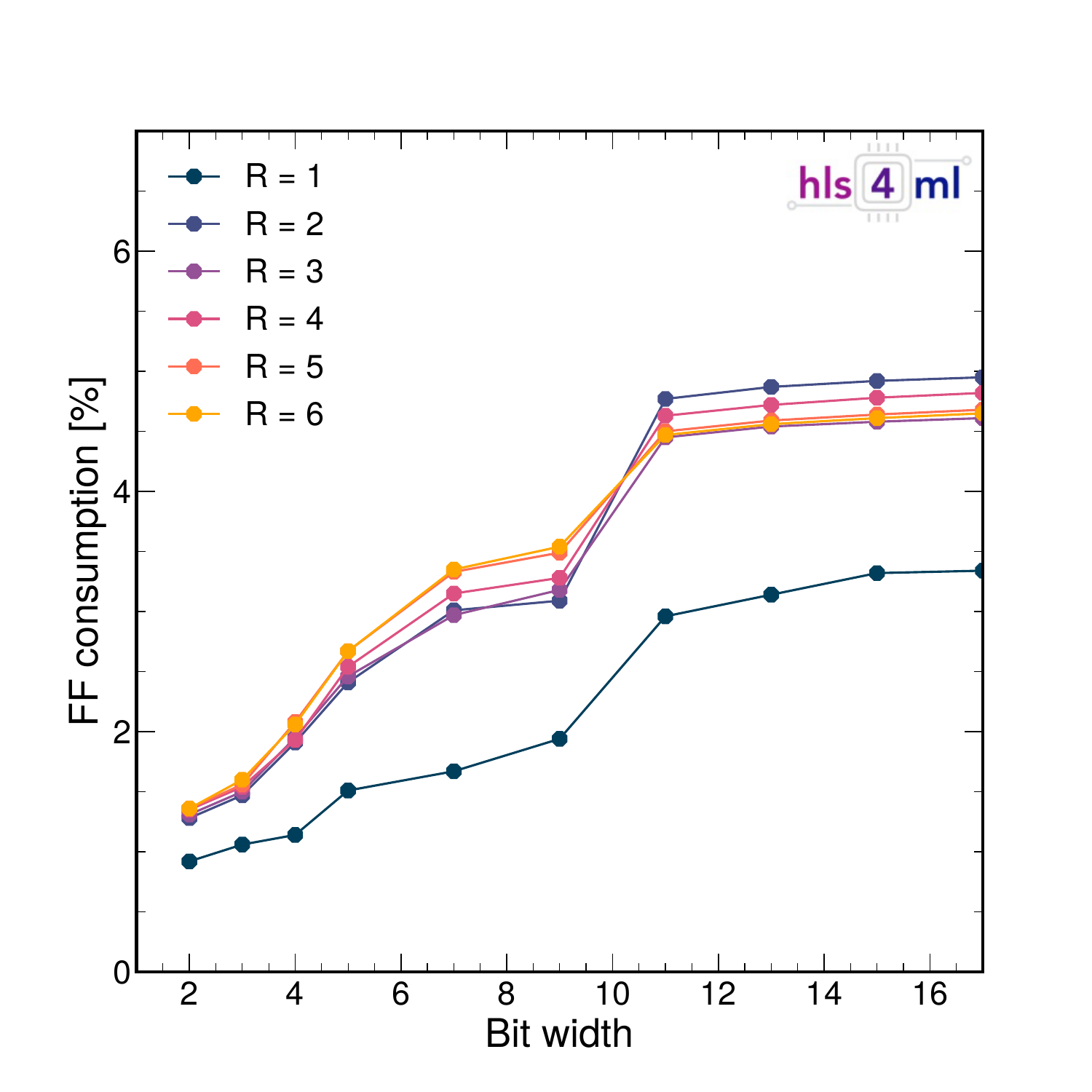}
        \caption{The model latency (top left), initiation interval (top right), DSP (middle left), LUT (middle right) and FF (bottom) consumption as a function of bit width and for different reuse factors for the QKeras (Q) model. \label{fig:quant_per_reuse}}
\end{figure}

Figures~\ref{fig:full_per_reuse}-\ref{fig:quant_per_reuse} show the model latency, initiation interval, DSP, LUT and FF consumption as a function of bit width and for different reuse factors for the BF, BP and Q models, respectively. 
A similar behaviour is observed for all models, where the latency roughly scales with one unit of reuse factor and the DSP consumption scales as the inverse of the reuse factor. 
The BRAM consumption does not depend on the reuse factor.
% , but does have a small dependence on the bit width. This is more pronounced for the BF and BP models as every layer is quantized to a narrower bit width, whereas the Q models have some layers at a higher bit width (like batch normalization layers). Most of the BRAMs are used for layer outputs ({\em channels}) and some for the sliding window used in the convolutional layer implementation. The BRAM consumption naturally depends on the bit width, as Vivado HLS will internally decide whether the FIFO will be placed on BRAM or LUT depending on the length of the stream and the bit width used.

For the BF models in Fig.~\ref{fig:full_per_reuse}, there is an unexpected drop in DSP consumption at a bit width of 12 for models using a reuse factor of one. 
For this model, only 13\% of the DSPs (corresponding to 876 units) are used, whereas the same mode with a reuse factor of six uses 19\% of the available DSPs (corresponding to a total of 1,311). 
We would expect models with higher reuse factors to use fewer resources and not vice versa. 
This unexpected behaviour is only observed for one data point. 
We have investigated things to the extent possible, but can only map the results back to how resources are allocated within HLS.

 For the Q models in Fig.~\ref{fig:quant_per_reuse}, the DSP consumption (middle left) of the models using a reuse factor of one and those using a reuse factor of two overlap above a bit width of ten. 
 The reason for this is that the maximum number of DSPs are reached for both model types, and multiplications are therefore forced to other resources. 
 This effect can be seen in the LUT consumption (middle right), where the model using a reuse factor of 1 uses significantly more LUTs than the other models.

\clearpage
\bibliographystyle{lucas_unsrt}   
\bibliography{references}

\end{document}